\documentclass[10pt,twocolumn,letterpaper]{article}

\usepackage[pagenumbers]{iccv} 

\usepackage{soul}
\usepackage{booktabs}
\usepackage{algpseudocode}
\usepackage{algorithm}
\usepackage{multirow}
\usepackage{multicol}
\usepackage{mathtools}
\usepackage{float}
\usepackage{ragged2e}
\usepackage[export]{adjustbox}
\usepackage{dsfont}
\usepackage{soul}
\let\mathbb\mathds

\newcommand{\myparagraph}[1]{\vspace{3pt}\noindent{\bf #1}}

\definecolor{iccvblue}{rgb}{0.21,0.49,0.74}
\usepackage[pagebackref,breaklinks,colorlinks,allcolors=iccvblue]{hyperref}

\def\paperID{2932} 
\def\confName{ICCV}
\def\confYear{2025}
\def\methodName{{AIM}\xspace}
\def\methodFullName{Amending Inherent Interpretability via Self-Supervised Masking\xspace}

\title{AIM: \underline{A}mending Inherent \underline{I}nterpretability via Self-Supervised \underline{M}asking}

\author{Eyad Alshami$^{1, 2}$
\quad
Shashank Agnihotri$^{3}$
\quad
Bernt Schiele$^{1,2}$
\quad
Margret Keuper$^{1,3}$\\
{\normalsize$^{1}$Max-Planck-Institute for Informatics, Saarland Informatics Campus, Germany} \\
{\normalsize$^{2}$RTG Neuroexplicit Models of Language, Vision, and Action, Saarbrücken, Germany} \\
{\normalsize$^{3}$Data and Web Science Group, University of Mannheim, Germany} \\
{\tt\small \{ealshami,schiele,keuper\}@mpi-inf.mpg.de, shashank.agnihotri@uni-mannheim.de}
}

\begin{document}
\maketitle
\begin{abstract}
It has been observed that deep neural networks (DNNs) often use both genuine as well as spurious features. 
In this work, we propose ``\methodFullName'' (\methodName), a simple yet interestingly effective method that promotes the network’s utilization of genuine features over spurious alternatives without requiring additional annotations. 
In particular, \methodName uses features at multiple encoding stages to guide a self-supervised, sample-specific feature-masking process. As a result, \methodName enables the training of well-performing and inherently interpretable models that faithfully summarize the decision process.
We validate AIM across a diverse range of challenging datasets that test both out-of-distribution generalization and fine-grained visual understanding. 
These include general-purpose classification benchmarks such as ImageNet100, HardImageNet, and ImageWoof, as well as fine-grained classification datasets such as Waterbirds, TravelingBirds, and CUB-200.
\methodName demonstrates significant dual benefits: interpretability improvements, as measured by the Energy Pointing Game (EPG) score, and accuracy gains over strong baselines.
These consistent gains across domains and architectures provide compelling evidence that \methodName promotes the use of genuine and meaningful features that directly contribute to improved generalization and human-aligned interpretability.
\end{abstract}    
\section{Introduction}
\label{sec:intro}
\begin{figure}[ht]
    \setlength{\belowcaptionskip}{0pt}   
    \centering

    \includegraphics[width=0.80\linewidth, valign=b]{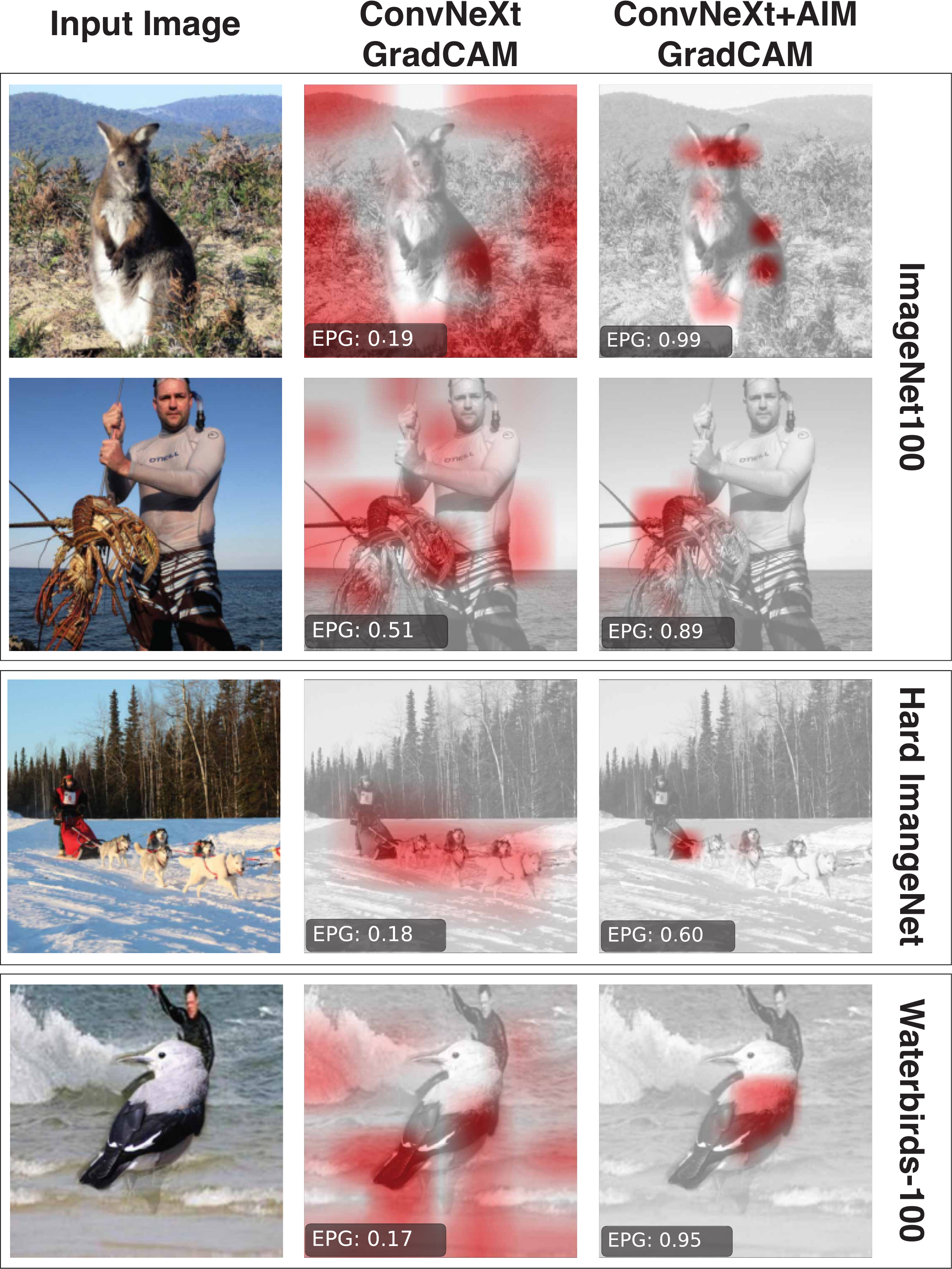}

    \caption{\textbf{\methodName uses self-supervised masking to focus more on the object of interest, relying only on the image label.} As shown, and in terms of attribution localization, it outperforms baseline methods, even in challenging scenarios like the WaterBirds dataset.}
    \label{fig:teaser}
\end{figure}

Modern deep neural networks (DNNs) have achieved remarkable success across domains such as Natural Language Processing and Computer Vision.
Despite their impressive performance metrics, these models often use spurious features that happen to correlate with target labels in training data but lack causal relevance to the task. This phenomenon, sometimes called `Clever Hans' behavior~\cite{lapuschkin2019unmasking, sebeok1981clever}. A classic example is classification models trained to distinguish between `land birds' and `water birds' for the WaterBirds dataset~\cite{sagawa2019distributionally, petryk2022guiding} that often learn to classify the background environment rather than the birds themselves, resulting in poor generalization when birds appear in atypical habitats. It is therefore desired that models leverage \emph{genuine} features, that are distinctive, class specific, and are localized on the object. Recent research has shown that, while DNNs exhibit dependence on spurious features, they simultaneously acquire some genuine features~\cite{kirichenko2022last}. This key insight suggests an opportunity: How can we promote the models’ utilization of genuine features while suppressing spurious ones? 
Prior works proposed using extra annotations in the form of bounding boxes, segmentation masks, or other guiding mechanisms~\cite{fel2022harmonizing, yang2023improving, linsley2018learning, gao2022aligning, gao2022res, petryk2022guiding, rao2023studying, teney2020learning, mitsuhara2019embedding, schramowski2020making, shen2021human, teso2023leveraging}. These mechanisms help focus the model on genuine features while ignoring spurious ones during training. 
However, getting these extra annotations is often nonviable. 

Thus, we propose \methodName (\methodFullName), a method that encourages the model to focus on genuine features and ignore spurious ones without needing annotations beyond image labels. \methodName employs a self-supervised masking mechanism that systematically identifies and prioritizes dependable feature maps of convolutional neural networks by masking out spurious features and retaining only dependable ones. Unlike previous approaches that rely on external attribution methods or require expensive additional annotations, \methodName operates by applying learnable binary masks to feature maps, allowing the model itself to determine which regions to retain or discard based on task performance. Our conjecture is that, when forced to select a subset of spatial features prior to making a classification decision, a model will consider those features most dependable that generalize best, i.e., that are genuine. We confirm this hypothesis using various analyses, including Energy Pointing Game (EPG) scores and evaluations using challenging datasets that provide many spurious cues.

The \methodName mechanism involves both a bottom-up processing of visual information through convolutional layers and a top-down pathway that refines feature selection. This feature refinement progressively identifies and filters out spurious features while preserving dependable ones. Importantly, this masking mechanism makes the model's decision process transparent: what is visible in the final feature representations directly causes the classification outcome, creating inherently interpretable models rather than relying on post-hoc interpretability methods.

We evaluate \methodName on challenging datasets specifically designed to test models' resilience to spurious features, including Waterbirds and TravelingBirds~\cite{koh2020concept}. These datasets present scenarios where background features strongly correlate with class labels during training but not during testing, challenging the models' out-of-distribution (OOD) generalization capabilities. We also validate our approach on standard fine-grained classification benchmarks such as CUB-200~\cite{WahCUB_200_2011}. Across these evaluations, \methodName demonstrates significant improvements in localization accuracy (measured by the Energy Pointing Game score). It also improves classification performance in OOD scenarios, showcasing improved generalization through the use of genuine features. Our results demonstrate that by encouraging the model to narrow its selection of spatial features for classification, it improves its focus on dependable ones. \methodName achieves this through a masking mechanism that produces inherently interpretable models without compromising task performance. The spatial masks learned by \methodName provide clear visual evidence of the features driving the model's decisions, establishing a ``what you see causes what you get'' relationship between the used features and predictions.

The primary contributions of this work are threefold:
\begin{itemize}
    \item We propose a simple yet effective self-supervised masking mechanism that guides DNNs to utilize dependable features over spurious alternatives, yielding inherently interpretable decisions while requiring only image labels.
    \item We demonstrate that our approach significantly improves the models' ability to localize genuine features, as quantified by Energy Pointing Game scores across multiple datasets.
    \item We show through extensive experiments that \methodName yields improvements in challenging out-of-distribution generalization scenarios where spurious features typically cause models to fail.
\end{itemize}

\section{Related Work}
\label{sec:related_work}

The prevalence of spurious correlations in DNNs, coupled with their increasing deployment in critical applications, has prompted extensive research in interpretability.

\myparagraph{Model Guidance and Attribution Methods.} Attribution methods generate attention maps~\cite{selvaraju2017grad, sundararajan2017axiomatic, shrikumar2017learning, bohle2022b, wang2020score, ramaswamy2020ablation, jiang2021layercam, chattopadhay2018grad} that highlight important input regions contributing to the final decision, aiding in the identification of erroneous reasoning by the model. Model guidance builds on these methods to align vision systems with ground truth guidance sources~\cite{ross2017right, shen2021human, gao2022aligning, gao2022res, teso2019explanatory, teso2019toward}, ensuring models are `right for the right reasons'~\cite{ross2017right}. This strategy relies on extra annotations, such as bounding boxes or attention maps~\cite{fel2022harmonizing, yang2023improving, linsley2018learning, gao2022aligning, gao2022res, petryk2022guiding, rao2023studying, teney2020learning}, which can be expensive and imperfect~\cite{gao2022res}.
Several methods aim to reduce the dependency on extra annotations. For instance, cost-effective model guidance can be achieved using only a small fraction of annotated images~\cite{rao2023studying}. When no extra annotations are available, approaches like~\cite{ross2017right} iteratively generate models with different reasoning but still require expert selection. Others improve explanations annotation-free simply by tuning the classification head's loss function (e.g., using binary cross-entropy)~\cite{how_to_prob}.
Alternatively,~\cite{asgari2022masktune} fine-tunes the model by masking discriminative features identified by the trained model.
Recent work~\cite{kirichenko2022last} observed that DNNs, while relying on spurious correlations, still learn genuine features. However, their method assumes prior knowledge of the spurious correlation.

\myparagraph{Content-Based Conditional Operations.} Methods in this domain~\cite{taghanaki2022masktunemitigatingspuriouscorrelations, Verelst_2020, hesse2023content} constrain the model to prioritize relevant spatial regions within the input features without requiring additional annotations. Some apply masking in the feature maps during the forward pass~\cite{Verelst_2020, hesse2023content}, pushing learned features to focus on `regions of interest'. Others apply masking in the input image domain~\cite{taghanaki2022masktunemitigatingspuriouscorrelations}.
For example,~\cite{hesse2023content} uses mask estimators with the Gumbel-softmax trick to predict binary masks that identify crucial areas and preserve them in higher resolutions. Similarly,~\cite{Verelst_2020} employs mask estimators with the Gumbel-softmax trick to select and process only important spatial regions, accelerating inference. Both works~\cite{Verelst_2020, hesse2023content} achieve spatial selection by progressively applying the masking strategy as the input moves through the network, from the initial layers all the way to the final layers, in a bottom-up fashion. A common problem reported by both works is that the generated masks tend to be fully active, requiring additional loss functions to push them to be sparse.
We hypothesize that the issue of fully active masks stems from the bottom-up approach itself. In contrast, \methodName allows the network to reassess the generated feature maps across the entire architecture, utilizing the top-down approach.  
This naturally produces sparse masks and enables the model to use spatially sparse feature maps, enhancing explainability.
\section{Proposed AIM Method}
\label{sec:method}
\begin{figure*}[ht]
    \setlength{\abovecaptionskip}{2pt}   
    \setlength{\belowcaptionskip}{0pt}   
    \centering
    \includegraphics[width=1.0\textwidth]{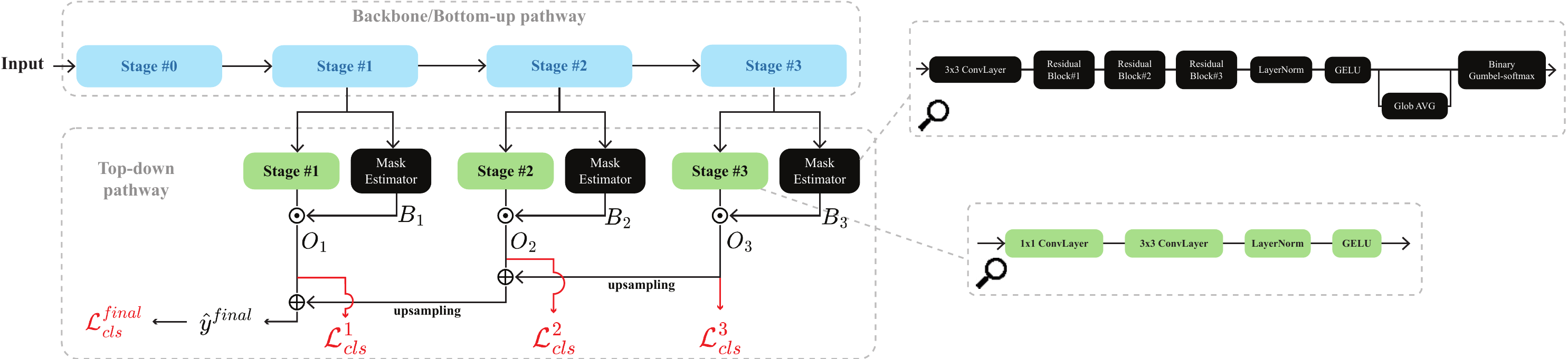}

    \caption{\textbf{Abstract Diagram of the [backbone]+\methodName Architecture}. The architecture consists of a \textbf{bottom-up} backbone and a \textbf{top-down} masking pathway. The bottom-up pathway has four encoding stages ($L=0$ to $L=3$). The top-down pathway mirrors this structure, with each stage $T$ corresponding to a bottom-up stage $L$. Each top-down stage has two parallel branches: one estimates a binary mask via a convolutional network with Gumbel-softmax, and the other processes features using a structure inspired by \cite{FPN}. The estimated binary mask is element-wise multiplied with the processed features to create a spatially sparse feature map. These sparse maps are then iteratively combined with the output of the subsequent top-down stage through element-wise summation.}
    \label{fig:Architecture}
\end{figure*}
Our method comprises two main pathways: a convolutional neural network (CNN) for the bottom-up pathway with multiple encoding stages, and a top-down pathway with corresponding self-supervised masking modules. 
Our work proposes a novel top-down pathway that helps the CNN focus on genuine features and ignore the spurious ones learned by the encoding stages of the bottom-up pathway. 
It achieves this using two main components: first, a mask estimator that sparsifies the feature maps from the encoding stages,
and second, a pathway that combines the sparse feature maps from different encoding stages.
The following describes these main components of our method in detail.
Please note, we address the task of image classification involving $C$ classes, given a dataset $\{(x_i, y_i)\}_{i=1}^n$ of size $n$, where $x_i \in \mathbb{R}^{h \times w \times 3}$ represents input images and $ y_i $ their corresponding labels.
Our approach does not require any additional annotations.

\myparagraph{Overall Architecture.}\label{subsec:architecture}  
Our architecture builds on the Feature Pyramid Network (FPN) framework~\cite{FPN}, adopting its top-down pathway structure. As illustrated in \Cref{fig:Architecture}, the model operates through two distinct pathways: a bottom-up pathway for hierarchical feature extraction and a top-down pathway for multi-scale feature integration.  
The bottom-up pathway employs a backbone network to generate hierarchical feature representations from input images. The top-down pathway iteratively combines these multi-scale features, propagating semantic information from the final high-level feature maps backward to earlier stages. Unlike the original FPN, which extends the top-down pathway to the highest-resolution initial feature map, we introduce a hyperparameter to control the termination depth of this pathway. This modification enables systematic analysis of how varying degrees of semantic detail from intermediate layers affect both the guidance mechanism and overall network performance. 
For example in our baseline ConvNeXt-Tiny backbone \cite{liu2022convnet2020s}, comprising four convolutional stages $\{S_0, S_1, S_2, S_3\}$ (where $S_3$ marks the final stage) with spatial resolutions $\{56^2, 28^2, 14^2, 7^2\}$, the top-down pathway stages $\{T_0, T_1, T_2, T_3\}$ mirror these resolutions. We study how integrating intermediate top-down features (\eg $T_1$, $T_2$) to the final stage $T_3$ enhances the model’s self-guiding capability.  
In the top-down pathway, shown in \Cref{fig:Architecture}, each stage receives the output of the corresponding stage in the bottom-up pathway, processes it, and prepares it for integration with outputs from the subsequent stage in the top-down sequence. This involves passing the feature maps through: 1) a $1 \times 1$ convolutional layer to align channel dimensions across stages, 2) a $3 \times 3$ convolutional layer for spatial refinement and 4) Layer Normalization and GELU activation.
The computational overhead and the increase in the number of parameters are moderate, as summarized in \Cref{sec:appendix:computational_overhead} in the appendix.

\myparagraph{Mask Estimation.} To enable spatial selection of dependable feature regions, we incorporate a learnable mask estimator at every stage of the top-down pathway. 
Each estimator consists of a lightweight convolutional neural network (CNN) that predicts a binary mask using the Gumbel-Softmax trick \cite{Verelst_2020}. 
While prior work by \citet{Verelst_2020} and \citet{hesse2023content} employs a bottom-up masking strategy (where binary masks iteratively select spatial regions at each stage) we adapt this concept to our top-down framework. 
Our empirical results demonstrate that this adaptation inherently produces spatially sparse and focused masks, enabling the network to prioritize salient regions without an additional supervision signal. 
As shown in \Cref{fig:Architecture}, the architecture of the mask estimators used in our method begins with a $3 \times 3$ convolutional layer, followed by three residual blocks that each utilize $3 \times 3$ convolutional layers. 
The output is then split into two branches: an identity branch and a global average pooling operation to capture global context information. 
The global context vector is expanded and concatenated with the output of the identity branch. Finally, a $1\times 1$ convolutional layer is applied to generate the final single-channel feature maps. 
This single layer is passed through the Gumbel-softmax module adapted from~\cite{Verelst_2020} to generate the final binary mask.
Each mask estimator uses the output from the corresponding stage of the backbone network as its input and generates a binary mask $ B $ with the same spatial resolution as the input feature maps. These masks highlight model-regarded dependable features in the feature maps generated by the corresponding convolutional stage.  
Formally, at each stage $ \ell $ in the architecture, the bottom-up stage $ S_\ell $ processes its input $ x_\ell $, producing feature maps $ S_\ell(x_\ell) $. These feature maps are then passed through the two branches at the corresponding top-down stage. 
The first branch, denoted as $ T_\ell $, is responsible for unifying the number of channels and post-processing the feature maps to prepare them for merging. It takes $ S_\ell(x_\ell) $ as input and produces the transformed feature maps $ T_\ell(S_\ell(x_\ell)) $.
The second branch is responsible for generating the binary mask, which highlights dependable features identified by the model. This process involves two steps. First, a soft attention decision map $ A_\ell \in \mathbb{R}^{w_\ell \times h_\ell} $ is computed by a simple mask estimating module $M$:
\begin{equation}A_\ell = M(S_\ell(x_\ell))\end{equation}
Next, and following~\cite{hesse2023content}, to obtain the binary mask, a binary Gumbel-softmax module $G$ is applied element-wise to $ A_\ell $, resulting in $ B_\ell \in \{0,1\}^{w_\ell \times h_\ell} $:
\begin{equation}B_\ell = G(A_\ell)\end{equation}
Finally, the spatially sparse output of the top-down stage $ \ell $ is then computed by element-wise multiplying the processed feature maps from the feature processing branch with the binary mask from the mask estimator:
\begin{equation}O_\ell = T_\ell(S_\ell(x_\ell)) \odot B_\ell\end{equation}
where $ \odot $ denotes element-wise multiplication. This operation results in $ O_\ell $, which retains only the dependable features as determined by the binary mask, effectively filtering out spatial regions with spurious features.

\myparagraph{Top-Down Sparse Feature Fusion}
Along the top-down pathway, the output of each stage is up-sampled through nearest-neighbor interpolation to match the resolution of the next lower stage and then merged with its feature maps through element-wise summation. This continues down the pathway, progressively integrating multi-scale information. The final aggregated feature maps, containing reliable multi-scale features, are used for classification.

\myparagraph{Supervising The Mask Estimators.} Since our method does not rely on additional annotations, we supervise the mask estimators indirectly using the classification loss computed on the masked feature maps $O_\ell$. At each stage of the top-down pathway, this loss is applied before merging with higher-stage feature maps. This strategy ensures that each stage independently identifies and learns important regions based solely on the feature maps available up to that stage.

We pass these feature maps through a classifier $ f_\ell $ to obtain the predicted class probabilities $ \hat{y}^{(\ell)} $:
\begin{equation}
\hat{y}^{(\ell)} = f_\ell(O_\ell)
\end{equation}
And then compute the classification loss at each stage, $\mathcal{L}_{\text{cls}}^{(\ell)}$. 
At the final stage, the merged sparse feature maps, combining outputs from all previous stages, are passed through the final classifier $ f_{\text{final}} $ to obtain the final predicted class probabilities $ \hat{y}^{\text{final}}$:
\begin{equation}
\hat{y}^{\text{final}} = f_{\text{final}}(O_{\text{final}})
\end{equation}
During training, our model self-guides to highlight spatial regions with dependable features within each stage's output. 
By enabling the network to select these regions across all layers, we empirically show that this improves classification performance and enables the reliance on spatially sparse maps, leading to transparent and inherently interpretable decision-making.

\myparagraph{Optional Mask Annealing. } 
Our approach naturally generates sparse masks, but enforcing additional sparsity during training on challenging out-of-distribution datasets, such as Waterbirds and TravelingBirds, improved performance and produced more focused masks. This is done by applying a mean-squared loss on the number of active elements in the generated masks using a threshold $\tau_i$ as follows:
\begin{equation}
\resizebox{0.85\columnwidth}{!}{$
\mathcal{L}_{\text{masks}_i} = (r_i - \tau_i)^2\;\;\; 
\textrm{where} \;\;\; r_i = \frac{\sum_{j=0,k=0}^{B^i_h, B^i_w} \mathbb{1}(B^i_{j,k} = 1)}{\sum_{j,k}^{B^i_h, B^i_w}1}
$}
\end{equation}
Where $r_i$ is the ratio of active elements in the generated binary mask $B^i$, and $\tau_i$ is a threshold hyperparameter that can be selected for each stage's mask estimator. We use a masking annealing technique to help the network gradually adapt to sparsity constraints without disrupting learning. Training begins with fully active masks (i.e.~$\tau_i=1.0$) and progressively lowers the active-area loss threshold each epoch until it reaches a target value (e.g. $\tau_i = 0.35$), which is then held for the remainder of training. The annealing duration is treated as a hyperparameter. This strategy improves mask quality and stabilizes learning. (For more details, see \Cref{sec:appendix:Ablation_results}).
Based on this setup, the final loss is defined as:
\begin{equation}
 \mathcal{L}_{\textit{Total}} = \lambda \sum_{i=L}^{\ell}\mathcal{L}_{\text{masks}_i} + \sum_{i=L}^{\ell}\mathcal{L}_{\text{cls}}^{(i)}  
 \end{equation}

Here, $L$ is the index of the highest stage, and $\ell$ denotes the final stage we aim to reach in the top-down pathway. The parameter $\lambda$, set to 6 in all experiments, weights the mask’s active-area loss to be on the same order of magnitude as the classification loss. For a full list of hyperparameters, see \Cref{sec:appendix:implementation_details}.

\section{Experiments}
\label{sec:experiments}
\begin{figure*}[t]
    \setlength{\abovecaptionskip}{0pt}   
    \setlength{\belowcaptionskip}{0pt}   
    \centering
    \begin{tabular}{@{\hskip 2.9mm}c@{\hskip 2.9mm}c@{\hskip 2.9mm}c}
        \includegraphics[width=0.3\textwidth]{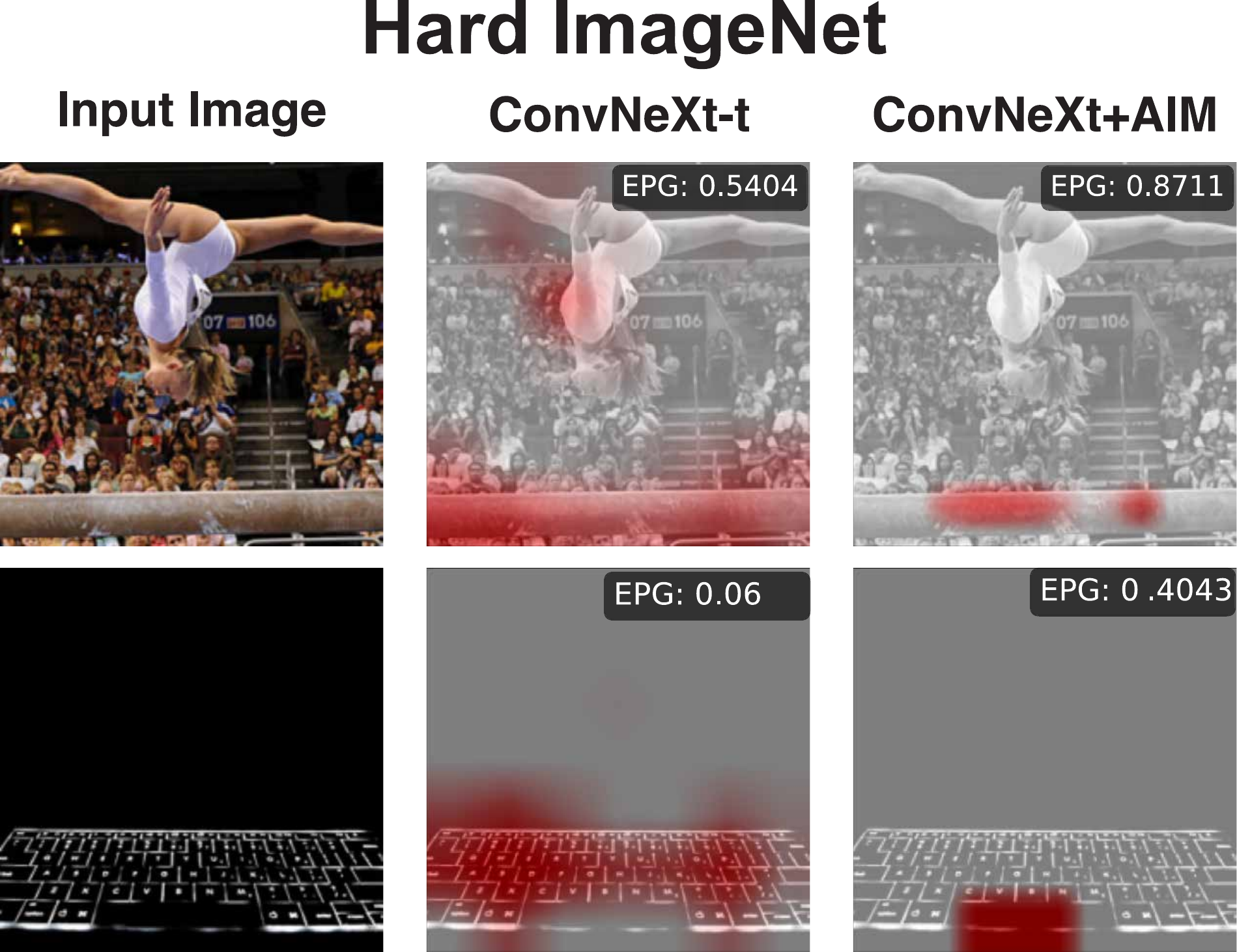} &
        \includegraphics[width=0.3\textwidth]{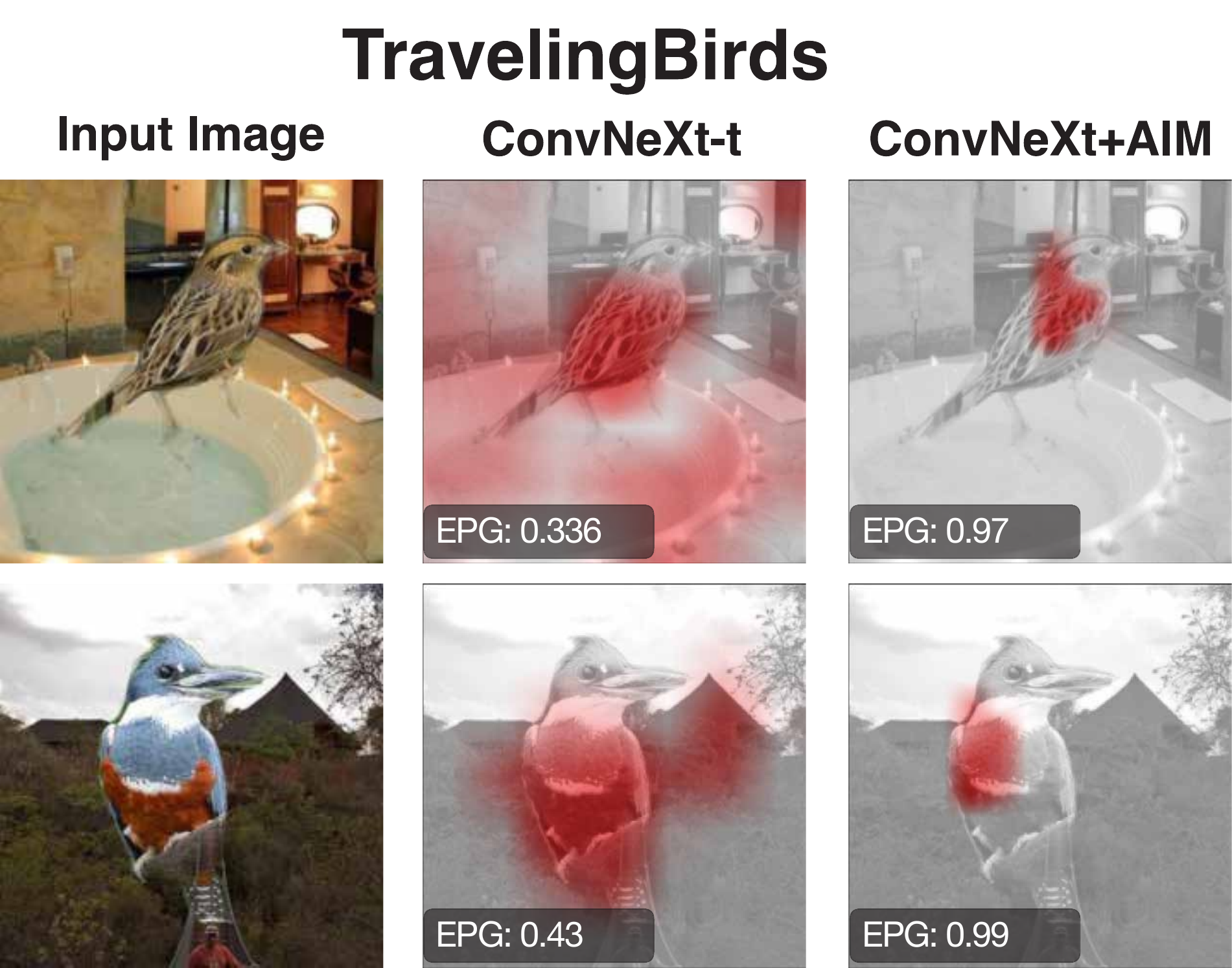} &
        \includegraphics[width=0.3\textwidth]{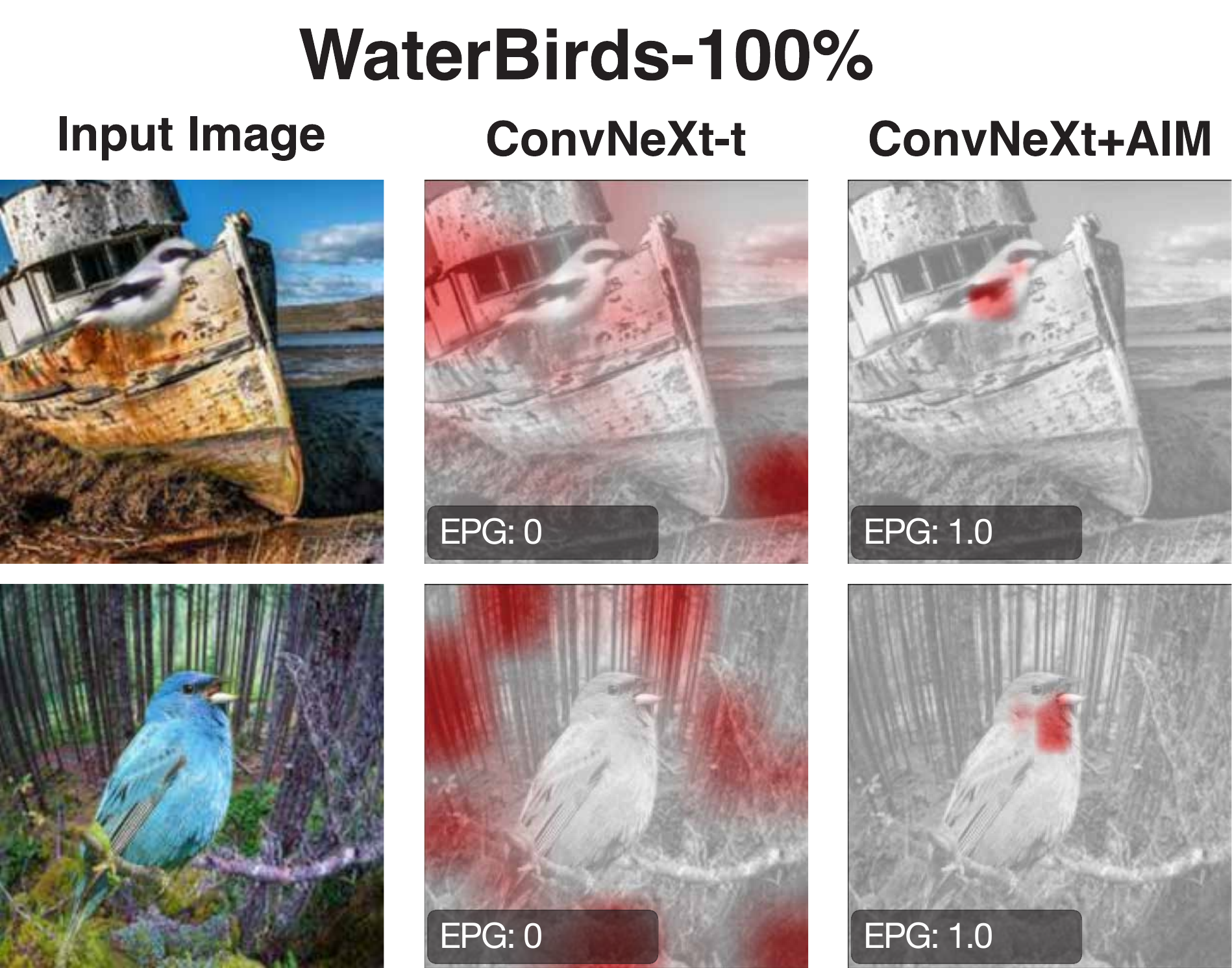} \\
        \hline 
        \noalign{\vskip 5pt}
        \multicolumn{3}{c}{\includegraphics[width=0.945\textwidth]{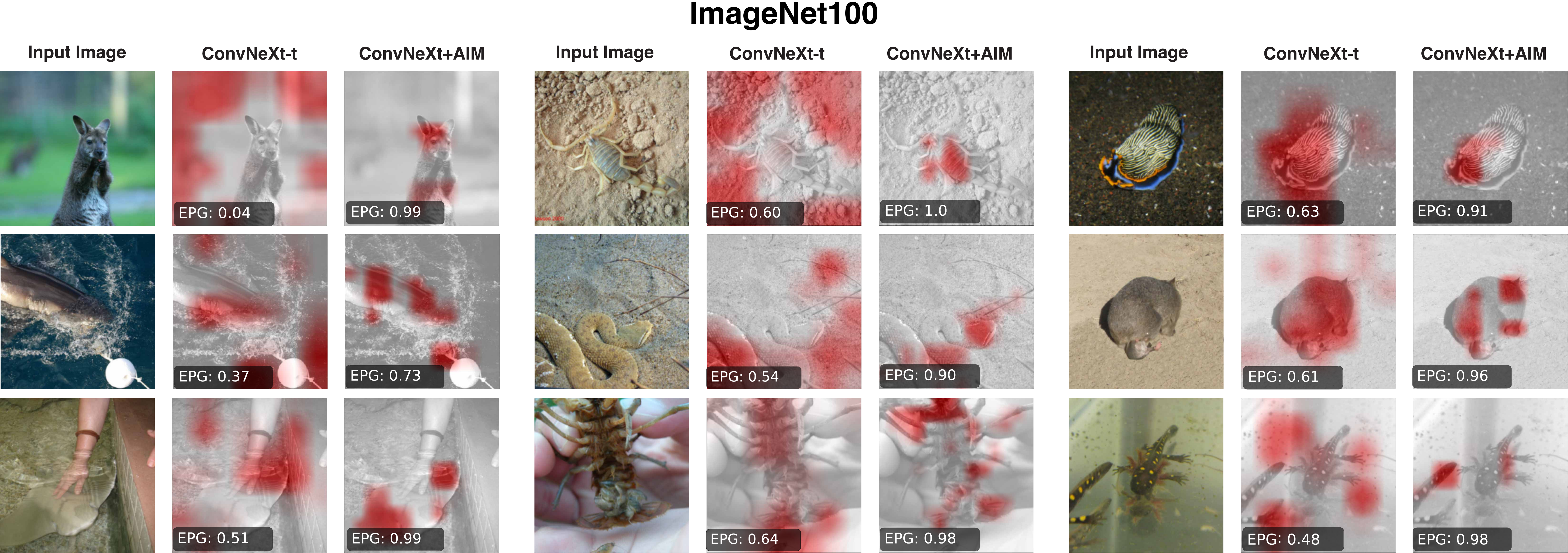}}
    \end{tabular}
    \caption{\footnotesize \textbf{Models amended with AIM consistently exhibit enhanced localization of genuine features, effectively suppressing spurious cues in both in-domain and out-of-domain scenarios.} A qualitative visualization of Grad-CAM heatmaps comparing baseline ConvNeXt-tiny models and ConvNeXt-tiny+AIM models across HardImagenet (classes shown: Balance Beam, Space Bar), TravelingBirds, WaterBirds-100\%, and ImageNet100 datasets. The EPG scores, with a range of 0.0 to 1.0, are indicated on each heatmap. For more qualitative results, see Appendix~\ref{sec:appendix:Qualitative_results}.}
    \label{fig:GradCAM-vis}
\end{figure*}
In this section we show that \methodName helps to retain in-domain performance while significantly boosting out-of-domain performance.
First, however, we describe some important implementation details (more details in 
\Cref{sec:appendix:implementation_details}).

\myparagraph{\methodName Architectural Variants.} As detailed in \Cref{subsec:architecture}, we parameterize the top-down pathway's depth by the number of stages traversed, denoting these variants as ``Backbone+\methodName [\textit{index}]'', where \textit{index} signifies the stage T where propagation ceases. For instance, with ResNet50, which has five convolutional stages, including the stem cell, we implement two main variants: ResNet50+\methodName(2) incorporates feature maps from stages 4, 3, and 2, while ResNet50+\methodName(3) includes only feature maps from stages 4 and 3.

\myparagraph{Baselines.} To evaluate our approach, we tested the effect of applying \methodName across various backbone architectures by comparing the performance of each backbone with and without \methodName integration. We utilized ConvNeXt-tiny~\cite{liu2022convnet2020s}, ResNet-50~\cite{he2015deepresiduallearningimage}, and ResNet-101~\cite{he2015deepresiduallearningimage}. All of these models are pre-trained on ImageNet-1k~\cite{5206848}.

\myparagraph{Evaluation Metrics.} 
To assess how effectively \methodName promotes dependable feature learning, we evaluate spatial localization using the Energy Pointing Game (EPG) score~\cite{wang2020score}. This metric, based on attribution maps (e.g., GradCAM~\cite{selvaraju2017grad}, Guided GradCAM~\cite{Guided_grad_cam}, or other attribution methods) and ground-truth binary masks, calculates the ratio of attribution within the mask’s active region to the total attribution. For implementation details, see ~\Cref{sec:appendix:implementation_details-EPG}. In addition, we show that our method not only preserves but also improves classification accuracy.

\myparagraph{Mask Annealing via Active-Area Loss.} 
As detailed in \Cref{subsec:architecture}, in addition to the stage $T$ parameterization, we employ progressive mask sparsification via threshold annealing during training, reducing the initial 100\% active area to either 35\% or 25\%. This introduces a new parameter, $\tau$, representing the final mask retention. Consequently, our models are denoted as ``backbone+\methodName [stage $T$, threshold $\tau$]''. For example, ResNet50+\methodName [2, 25\%] signifies operation through stage $T=2$ with 25\% mask retention.

\myparagraph{Datasets.} We evaluate our method on two categories of datasets. First, to test its robustness to spurious correlations, we use the synthetic Waterbirds (95\% and 100\% versions)~\cite{sagawa2019distributionally, petryk2022guiding} and Travelingbirds~\cite{koh2020concept} datasets. Second, to assess the broader adaptability and effectiveness of \methodName, we use standard benchmarks including ImageNet100~\cite{imagenet100}, Hard-ImageNet~\cite{hardimagenet}, and the fine-grained Caltech-UCSD Birds-200-2011 (CUB-200-2011)~\cite{WahCUB_200_2011}. Further details are provided in \Cref{sec:appendix:implementation_details-Datasets}.

\subsection{Results on Out-Of-Domain Datasets}
\begin{figure}[t]
    \setlength{\abovecaptionskip}{0pt}   
    \setlength{\belowcaptionskip}{0pt}   
    \centering

        \includegraphics[width=0.8\columnwidth]{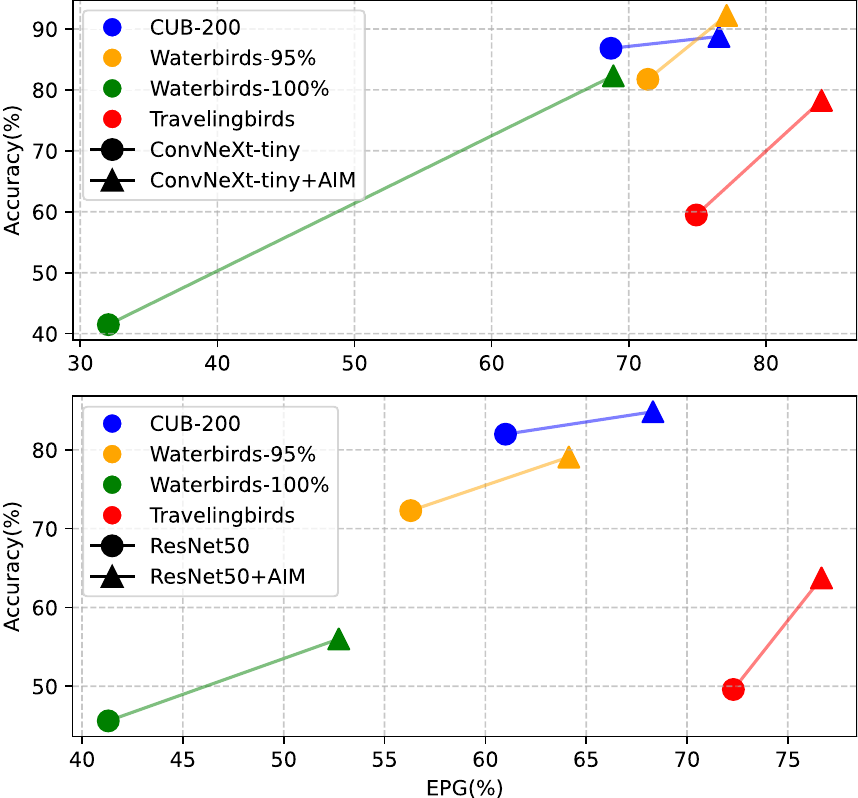} \\

    \caption{\small \textbf{Main results:} Across several datasets (CUB-200, Waterbirds-95\%, Waterbirds-100\%, Travelingbirds) and architectures (ConvNeXt-tiny, ResNet50) our \methodName approach consistency outperforms the respective baseline in both accuracy as well as interpretability as measured by the Energy Pointing Game score. On WaterBirds, following~\cite{sagawa2019distributionally}, we report the worst-group accuracy.}
    \label{fig:models_comparisons_epg_acc}
\end{figure}

\begin{figure}[b]
    \setlength{\abovecaptionskip}{0pt}   
    \setlength{\belowcaptionskip}{0pt}   
    \centering
    \begin{tabular}{c}
        \includegraphics[width=0.85\columnwidth]{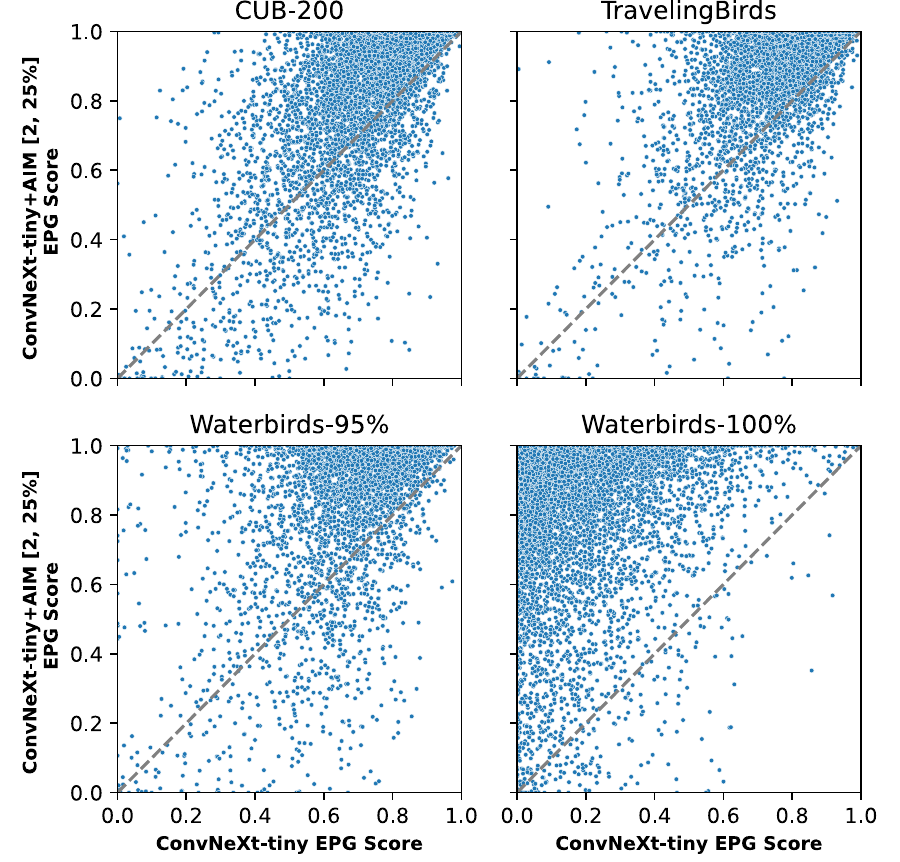} \\
    \end{tabular}
    \caption{\footnotesize EPG scores per sample are plotted for baseline model (x-axis) v/s model amended with \methodName (y-axis). We observe at a per-sample level for each of the four datasets that majority for the samples the EPG scores are improved by amending the model with our proposed \methodName.}
    \label{fig:epg_scatter_plot}
\end{figure}

The Waterbirds and TravelingBirds datasets contain synthetic spurious correlations, causing models to incorrectly rely on background cues rather than foreground objects. As \Cref{fig:models_comparisons_epg_acc} illustrates, our proposed \methodName mechanism consistently surpasses baseline backbone models. Vanilla backbone performance significantly degrades due to these biases; however, models integrated with \methodName show notable improvements in both EPG and accuracy across all tested out-of-domain datasets.
The primary motivation of \methodName, detailed in \Cref{sec:intro}, is to enhance the localization of genuine image features, thereby improving interpretability without compromising accuracy. The improved accuracy across various backbones emerges as an additional beneficial outcome.
The self-supervised masking strategy employed by \methodName enables models to consistently identify and rely on dependable features. \Cref{fig:GradCAM-vis} visually demonstrates this, contrasting baseline models, which are often distracted by misleading background cues, with \methodName-equipped models that reliably emphasize dependable regions. To confirm this visual improvement is perceived by humans, we conducted a user study that showed participants preferred our model's attribution maps over the baseline in 70.7\% of cases ($p < 0.00001$), providing strong evidence of more human-aligned interpretability (see \Cref{sec:appendix:user_study} for details). Quantitatively, higher EPG scores, computed using dataset-provided binary masks, confirm this improved localization. We also evaluate other attribution methods and observe similar EPG improvements on the Waterbirds-95\% dataset using Guided GradCAM and Guided Backprop, as shown in \Cref{sec:appendix:attribution_agnostic}, further confirming the robustness of our localization gains. These scores validate our hypothesis from Sec.\ref{sec:intro} regarding the dependability of features identified by \methodName.
As summarized visually in \Cref{fig:models_comparisons_epg_acc} and detailed with precise metrics in \Cref{tab:models_comparison}, we see substantial EPG score improvements: approximately 6\% for Waterbirds-95\%, 30\% for Waterbirds-100\%, and 10\% for TravelingBirds. Accuracy gains are equally notable, reaching around 10\% for Waterbirds-95\%, 40\% for Waterbirds-100\%, and 18\% for TravelingBirds.
Furthermore, \Cref{fig:epg_scatter_plot} provides a detailed per-sample analysis of EPG scores for baseline models versus models with \methodName. It demonstrates that across all datasets, the majority of individual samples exhibit improved EPG scores. While overall improvements are substantial, a subset of examples maintained comparable performance, particularly when baseline EPG was already high, and a minimal number of instances showed slight EPG decreases. Additional comparisons against other relevant methods are provided in the \Cref{sec:appendix:other_methods_comparison}.

\begin{table*}[htbp]
    \centering
    \renewcommand{\arraystretch}{1.3}
    \caption{
        \textbf{Average Test Accuracies for ConvNeXt+AIM Configurations.}
        Comparison of the ConvNeXt-tiny baseline against our AIM-enhanced models on multiple benchmarks. The method shows significant gains, especially in worst-group accuracy on Waterbirds, highlighting its effectiveness in mitigating spurious correlations. All values are mean accuracy (\%) $\pm$ standard deviation.
    }
    \label{tab:models_comparison}
\scalebox{0.7}{
    \begin{tabular}{@{}l c c c c c c c c c c@{}}
        \toprule
        \multirow{3}{*}{\textbf{Model}} & \multicolumn{2}{c}{\textbf{ImageNet100}} & \multicolumn{2}{c}{\textbf{Hard-ImageNet}} & \multicolumn{4}{c}{\textbf{Waterbirds}} & \multicolumn{2}{c}{\textbf{TravelingBirds}} \\
        \cline{2-3} \cline{4-5} \cline{6-9} \cline{10-11}
        & & & & & \multicolumn{2}{c}{\textbf{100\%}} & \multicolumn{2}{c}{\textbf{95\%}} & & \\
        \cline{6-7} \cline{8-9}
        & \textbf{Acc} & \textbf{EPG} & \textbf{Acc} & \textbf{EPG} & \textbf{WG-Acc} & \textbf{EPG} & \textbf{WG-Acc} & \textbf{EPG} & \textbf{Acc} & \textbf{EPG} \\
        \hline
        ConvNeXt-t & 89.2 ($\pm$0.1) & 91.4 ($\pm$0.3) & 96.2 ($\pm$0.2) & 36.6 ($\pm$0.5) & 39.6 ($\pm$5.4) & 57.2 ($\pm$6.0) & 81.6 ($\pm$3.2) & 68.3 ($\pm$3.2) & 59.5 ($\pm$0.8) & 74.4 ($\pm$0.6) \\
        
        \textbf{ConvNeXt-t+AIM[1, 25\%]} & 90.5 ($\pm$2.1) & 91.5 ($\pm$0.1) & \underline{97.1 ($\pm$0.3)} & \underline{38.8 ($\pm$0.9)} & 73.6 ($\pm$4.5) & \underline{60.1 ($\pm$1.3)} & 91.2 ($\pm$0.8) & \textbf{77.1 ($\pm$5.2)} & \underline{77.1 ($\pm$0.3)} & \underline{79.0 ($\pm$0.7)} \\
        
        \textbf{ConvNeXt-t+AIM[1, 35\%]} & \underline{90.5 ($\pm$0.1)} & 89.1 ($\pm$0.2) & \textbf{97.3 ($\pm$0.2)} & 33.2 ($\pm$0.5) & \underline{77.1 ($\pm$4.4)} & 57.2 ($\pm$1.3) & 90.7 ($\pm$0.7) & 63.0 ($\pm$1.2) & 71.5 ($\pm$1.3) & 72.6 ($\pm$1.5) \\

        \textbf{ConvNeXt-t+AIM[2, 25\%]} & 90.1 ($\pm$2.3) & \textbf{92.8 ($\pm$0.8)} & 96.8 ($\pm$0.5) & \textbf{40.1 ($\pm$1.5)} & 74.0 ($\pm$5.0) & 58.0 ($\pm$1.3) & \textbf{92.7 ($\pm$1.2)} & \underline{75.0 ($\pm$6.0)} & \textbf{77.4 ($\pm$0.2)} & \textbf{85.0 ($\pm$2.0)} \\
        
        \textbf{ConvNeXt-t+AIM[2, 35\%]} & \textbf{90.7 ($\pm$0.0)} & \underline{91.8 ($\pm$0.1)} & 97.1 ($\pm$0.1) & 33.6 ($\pm$1.1) & \textbf{78.1 ($\pm$2.3)} & \textbf{68.5 ($\pm$3.6)} & \underline{92.3 ($\pm$0.6)} & 71.7 ($\pm$6.4) & 71.0 ($\pm$0.4) & 77.7 ($\pm$0.4) \\
        \bottomrule
    \end{tabular}
    }
\end{table*}

\subsection{Comprehensive Evaluation on Diverse Classification Tasks}

To evaluate the adaptability and effectiveness of our proposed \methodName{} mechanism, we conducted experiments on a range of classification benchmarks, from fine-grained tasks to broader, general-purpose datasets.

First, we tested our method on the CUB-200 dataset, which poses a challenging fine-grained classification task requiring models to identify subtle and localized visual features. As shown in~\Cref{fig:models_comparisons_epg_acc}, incorporating our \methodName{} mechanism improves localization performance significantly: ConvNeXt-tiny+AIM achieves an approximate 6\% increase in EPG score over the baseline ConvNeXt-tiny model, while ResNet+AIM improves localization by around 9\% compared to the baseline ResNet model. These localization improvements are accompanied by a slight accuracy increase of about 2–3\%. For full details see \Cref{sec:appendix:extended_results_of_convnext_aim}.

Next, to validate the broader applicability beyond the domain of bird datasets, we evaluated \methodName on general-purpose image classification benchmarks. We conducted experiments using ConvNeXt-tiny on ImageNet100~\cite{imagenet100} and the challenging HardImageNet~\cite{hardimagenet}. As summarized in \Cref{tab:models_comparison}, our method shows consistent benefits across diverse domains. On ImageNet 100, it improves the EPG score by nearly 3 points while maintaining baseline accuracy, reinforcing our core claim of enhanced localization. On the more difficult HardImageNet, it boosts both accuracy and EPG, confirming its robustness as a domain-agnostic mechanism.

\section{Inherent Interpretability With Self-Supervised Masking}
\label{sec:interpretability_and_masking}
\begin{figure}[t]
    \setlength{\abovecaptionskip}{0pt}   
    \setlength{\belowcaptionskip}{0pt}   
    \centering
    \begin{tabular}{c}
        \includegraphics[width=0.8\columnwidth]{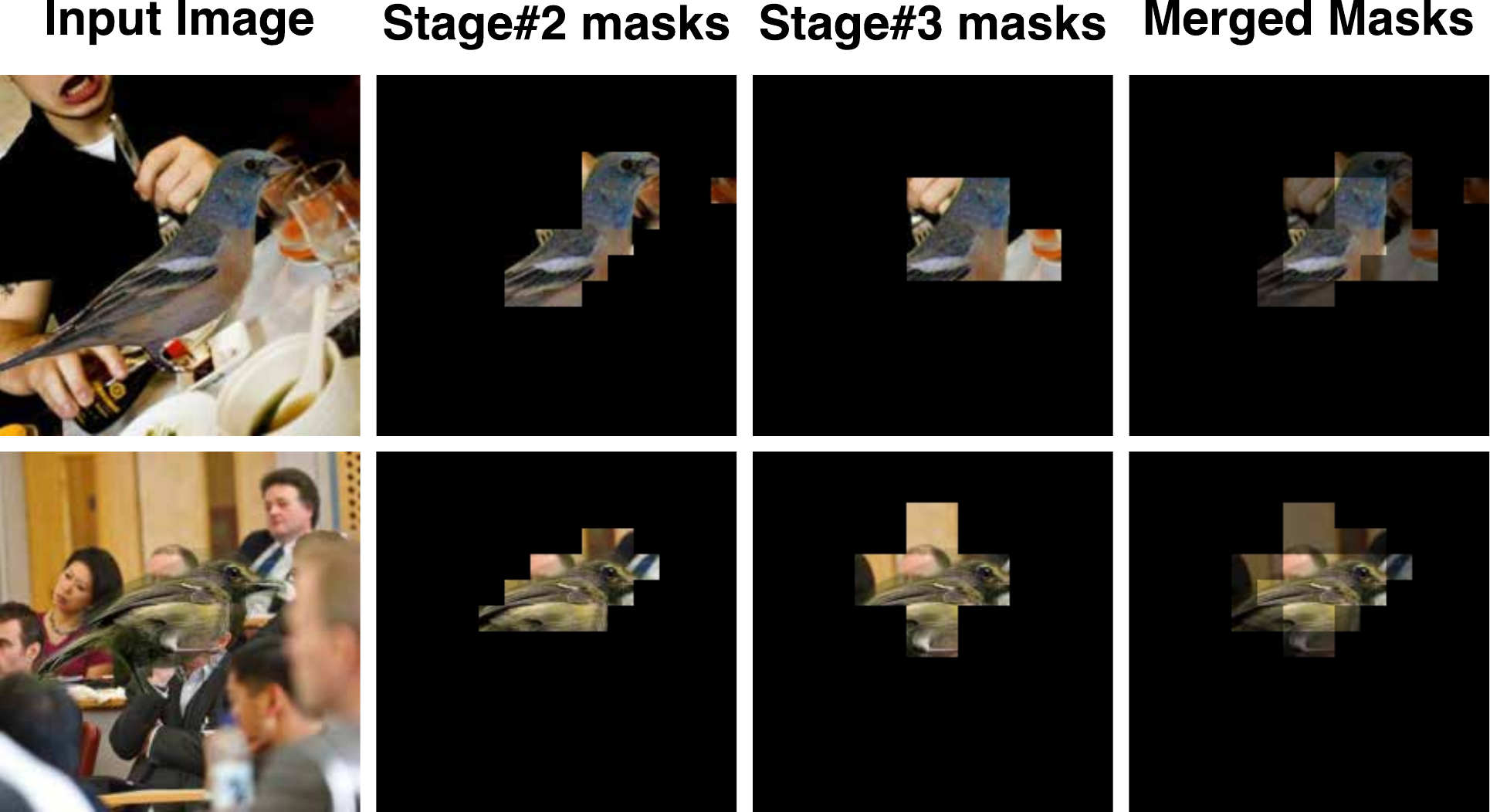}
    \end{tabular}
    \caption{Illustration of masks learned at the two stages within the network (using ConvNeXt-tiny+AIM [2, 25\%]), along with the final merged mask for each image. These merged masks highlight the sparse regions within the corresponding feature maps.}
    \label{fig:block-wise-mask-vis}
\end{figure}

Our \methodName mechanism offers inherent interpretability, which we visualize by depicting input images alongside their corresponding masks from the top-down pathway in \Cref{fig:block-wise-mask-vis}. This interpretability arises from the self-supervised masking performed by the mask estimator within \methodName. Furthermore, combined masks clearly illustrate how sparse feature maps from different stages of the top-down pathway are merged.

\begin{figure}[t]
    \setlength{\abovecaptionskip}{0pt}   
    \setlength{\belowcaptionskip}{0pt}   
    \centering
    \begin{tabular}{@{}c}
        \includegraphics[width=\columnwidth]{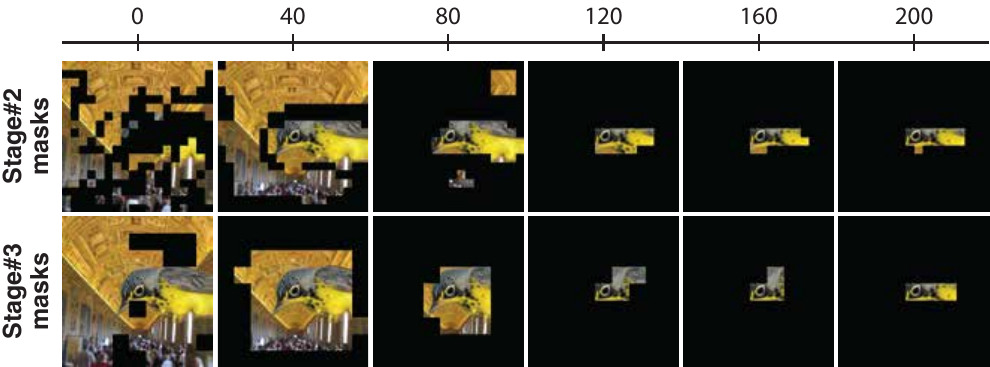}
    \end{tabular}
    \caption{\footnotesize Visualization of the evolution of learned masks at two different stages (Stage \#2 and Stage \#3) of a ConvNeXt-tiny+AIM [2, 25\%] model throughout the training epochs. As training progresses, the masks gradually become more sparse and accurately localized, highlighting the model's improved ability to identify and focus on regions containing genuine features in a self-supervised manner.}
    \label{fig:mask-evolution}
\end{figure}
\Cref{fig:mask-evolution} visualizes the evolution of these masks over epochs. The mask estimator initially starts with random values, but as training progresses, relying solely on the classification loss from image labels, it learns to focus on dependable features within the feature maps. As discussed in \Cref{sec:experiments}, these dependable features correspond to genuine features, indicated by EPG scores. Conversely, if the model learns incorrect masking, low EPG scores reflect non-genuine features and result in lower accuracy scores. Since \methodName's self-supervised masking mechanism is part of the model's forward pass, visualizing these masks directly reveals the basis of the model's decisions. This establishes a clear ``what you see causes what you get'' relationship between features and predictions.

\section{Analysis and Ablation}
\label{sec:analysis}
The following further explores the effectiveness of \methodName.
\begin{table}[t]
    \setlength{\abovecaptionskip}{2pt}   
    \setlength{\belowcaptionskip}{0pt}   
    \centering
    \caption{\footnotesize \textbf{Lower performance with Bottom-up Guiding Approach.} Result of the bottom-up ConvNeXt-t+\methodName applying the masking mechanism in the second and third stages in the vanilla ConvNeXt-tiny, compared to the top-down Refocs+ConvNeXt-t model.}
    \scriptsize
    \scalebox{1.2}{
    \begin{tabular}{@{}lc@{}}
        \toprule
        \textbf{Model}                              &  \textbf{CUB-200 (\%)} \\ \midrule
        \textit{bottom-up masking [1, 25\%] } ConvNeXt-t    & 72.79  ($\pm$8.51) \\
        ConvNeXt-t+\methodName [1, 25\%] & \textbf{88.82 ($\pm$0.213)} \\
        \midrule
        \textit{bottom-up masking [2, 25\%]} ConvNeXt-t     & 84.00 ($\pm$1.38) \\ 
        ConvNeXt-t+\methodName [2, 25\%]     & \textbf{88.677 ($\pm$0.25)} \\ \bottomrule
    \end{tabular}
    }
    \label{table:bottom-up_guidance}    
\end{table}

\subsection{\textbf{\textit{Top-down}} approach v/s \textbf{\textit{Bottom-up}} approach}
Inspired by~\cite{Verelst_2020}, we initially tested a bottom-up masking approach, utilizing the same mask estimators described in \Cref{subsec:architecture} but without a top-down pathway. In this setup, each convolutional stage of the backbone model had two branches: the original convolutional path and a mask estimator. The latter predicted a binary mask, applied to the convolutional output to create spatially sparse feature maps that proceeded to the next stage. Unlike~\cite{Verelst_2020}, we did not employ a skip-connection to convert sparse feature maps back into dense ones, aiming to preserve inherent explainability. However, this bottom-up guiding method performed poorly on the CUB-200 dataset, as shown in \Cref{table:bottom-up_guidance}, where bracketed numbers indicate the used stage and annealing. Furthermore, the generated masks tended to remain fully active despite applying the mask active-area loss (see \Cref{sec:appendix:Ablation_results-Bottom_up_bottlenecking} for further analysis), unlike the naturally focused masks produced by the top-down approach.

\subsection{Does \methodName have a center bias?}
To investigate potential center bias \cite{fatima2025corner} in our experiments, we tested models on images with birds positioned at the edges rather than center-frame. \Cref{table:center_bias} shows that while both vanilla ConvNeXt-tiny and ConvNeXt-tiny+\methodName experienced performance decreases compared to center-cropped images, \methodName still outperformed the baseline by approximately 2.5\%. Furthermore, as depicted in Figure~\ref{fig:center_bias}, \methodName continued to generate masks focused on birds despite partial visibility, confirming our approach does not depend on center bias for its effectiveness.
\begin{table}[t]
    \setlength{\abovecaptionskip}{1pt}   
    \setlength{\belowcaptionskip}{0pt}
    \centering
    \caption{\textbf{\footnotesize \methodName does not exploit the center-bias} \methodName manages to detect and focus on the bird achieving higher results compared to the vanilla ConvNeXt-tiny model}
    \scriptsize
    \scalebox{1.2}{
    \begin{tabular}{@{}lc@{}}
        \toprule
        \textbf{Model}                              &  \textbf{CUB-200 (\%)} \\ \midrule
        ConvNeXt-tiny    & 76.98 ($\pm$0.18) \\ 
        ConvNeXt-t+\methodName [1, 25\%]    & \textbf{79.33 ($\pm$0.45)} \\ \bottomrule
    \end{tabular}
    }
    \label{table:center_bias}    
\end{table} 
\begin{figure}[h!]
        \setlength{\abovecaptionskip}{1pt}   
        \setlength{\belowcaptionskip}{0pt}
        \centering
        \includegraphics[width=0.45\textwidth]{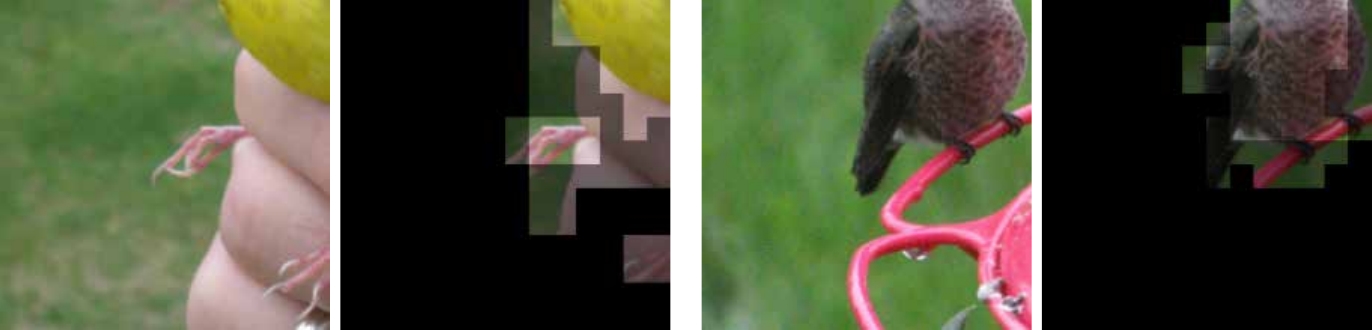}
        \caption{\textbf{\methodName models do not have a center-bias.} This illustration shows the merged masks generated by ConvNeXt-t+\methodName (2, 25\%) on two images from CUB-200.}
        \label{fig:center_bias}
\end{figure}
\vspace{-1em}

\section{Conclusion}
\label{sec:limits_future_work}
In this work, we propose \methodFullName (\methodName), a simple yet effective method that encourages networks to focus on dependable rather than spurious features via a self-supervised feature-masking process. Evaluated using the Energy Pointing Game (EPG) score on out-of-distribution and fine-grained classification tasks, \methodName improves localization on dependable features without sacrificing accuracy or requiring annotations beyond class labels.
\methodName produces inherently interpretable models by integrating sparse, top-down feature selection directly into the forward pass. It consistently improves both accuracy and localization across diverse datasets and architectures, with minimal overhead. These results highlight AIM’s potential as a lightweight and scalable approach to training models that are robust, generalizable, and aligned with meaningful visual cues.

\myparagraph{Future Work.} We plan to extend \methodName to Vision Transformers~\cite{VIT} by either reshaping patch embeddings into spatial feature maps or leveraging the hierarchical structure of Swin-Transformers~\cite{swinVIT} for more seamless integration.

\section{Acknowledgment.} Funded in part by the DFG (German Research Foundation, RTG 2853/1). S.A. and M.K. acknowledge support by DFG Research Unit 5336 Learning2Sense.

{
    \small
    \bibliographystyle{ieeenat_fullname}
    \bibliography{main}
}
\newpage
\appendix
{
\onecolumn
    \centering
    \Large
    \textbf{AIM: \underline{A}mending Inherent \underline{I}nterpretability via Self-Supervised \underline{M}asking} \\
    \vspace{0.5em}Paper \#2932 Supplementary Material \\
    \vspace{1.0em}
}

\section*{Table of Contents}
\addcontentsline{toc}{section}{Table of Contents}
The supplementary material covers the following information:
\begin{itemize}[label={}]
    \item \hyperref[sec:appendix:implementation_details]{\textbf{A. Implementation Details}}
    \begin{itemize}[label={}]
        \item \hyperref[sec:appendix:implementation_details-Hyperparameters]{A.1 Hyperparameters}
        \item \hyperref[sec:appendix:implementation_details-Active-area loss threshold annealing]{A.2 Active-area Loss Threshold Annealing}
        \item \hyperref[sec:appendix:implementation_details-Datasets]{A.3 Datasets}
        \begin{itemize}[label={}]
            \item \hyperref[sec:appendix:implementation_details-Datasets-CUB200]{A.3.1 CUB-200}
            \item \hyperref[sec:appendix:implementation_details-Datasets-WaterBirds]{A.3.2 WaterBirds}
            \item \hyperref[sec:appendix:implementation_details-Datasets-TraverlingBirds]{A.3.3 TraverlingBirds}
            \item \hyperref[sec:appendix:implementation_details-Datasets-ImageNet100]{A.3.4 ImageNet100}
            \item \hyperref[sec:appendix:implementation_details-Datasets-ImageWoof]{A.3.5 ImageWoof}
            \item \hyperref[sec:appendix:implementation_details-Datasets-ImageNetHard]{A.3.6 ImageNetHard}
        \end{itemize}
        \item \hyperref[sec:appendix:implementation_details-Testing_the_center_bias_of_the_masks]{A.4 Testing the Center Bias of the Masks}
    \end{itemize}
    \item \hyperref[sec:appendix:Quantitative_results]{\textbf{B. Quantitative Results}}
    \begin{itemize}[label={}]
        \item \hyperref[sec:appendix:Quantitative_results-Center_Bias]{B.1 Center Bias}
        \item \hyperref[sec:appendix:attribution_agnostic]{B.2 Attribution-Agnostic Localization Performance}
        \item \hyperref[sec:appendix:extended_results_of_convnext_aim]{B.3 Extended Results of ConvNeXt+AIM}
        \item \hyperref[sec:appendix:other_methods_comparison]{B.4 Comparison to Other Methods}
    \end{itemize}
    \item \hyperref[sec:appendix:Qualitative_results]{\textbf{C. Qualitative Results}}
    \begin{itemize}[label={}]
        \item \hyperref[sec:appendix:birds_data_qualitative_results]{C.1 Additional Quantitative Results}
        \item \hyperref[sec:appendix:Qualitative_results:center_bias]{C.2 Center Bias}
        \item \hyperref[sec:appendix:Qualitative_results:masks_attribution_maps]{C.3 Qualitative Comparison of Masks and Attribution Maps: AIM vs. Vanilla Backbone Models}
    \end{itemize}
    \item \hyperref[sec:appendix:Ablation_results]{\textbf{D. Ablation Results}}
    \begin{itemize}[label={}]
        \item \hyperref[sec:appendix:Ablation_results-Active_Area_Loss_Annealing]{D.1 Active-Area Loss Annealing}
        \item \hyperref[sec:appendix:Ablation_results-Without_The_Auxiliary_Losses]{D.2 Without the Auxiliary Losses}
        \item \hyperref[sec:appendix:Ablation_results-single_level]{D.3 Do We Need Multiple Mask Estimators at Each Level?}
        \item \hyperref[sec:appendix:Ablation_results-Bottom_up_bottlenecking]{D.4 The Bottom-Up Bottlenecking Approach}
        \item \hyperref[sec:appendix:Ablation_results-initialization]{D.5 Emphasizing peripheral regions in mask estimator initialization}
    \end{itemize}
\end{itemize}

\section{Implementation Details}
\label{sec:appendix:implementation_details}
In this section, we present a comprehensive overview of the implementation details of our proposed model to ensure reproducibility and facilitate future research. We begin by outlining the specific hyperparameters used during training, the datasets utilized in our experiments. We then detail the annealing procedures employed to adapt the threshold of the active-area loss used for the mask estimators in the AIM architecture. Additionally, we explain the shifted-center cropping technique implemented to assess the model's susceptibility to center bias.

\subsection{Hyperparameters}
\label{sec:appendix:implementation_details-Hyperparameters}
The primary hyperparameters used for our AIM models are the shown in Table~\ref{table:appendix:hyperparameters}:

\begin{table}[H]
    \centering
    \renewcommand{\arraystretch}{1.3}
    \small
    \caption{The Hyperparameters values used for training AIM models.}
    \label{table:appendix:hyperparameters}
    \begin{tabular}{@{}lc@{}}
        \toprule
        \textbf{Hyperparameter} & \textbf{Value} \\
        \midrule
        Validation dataset  ratio & 20\% of training dataset \\
        Batch Size & 512 \\
        top-down pathway learning rate & 0.01 \\
        Weight Decay & 0.001 \\
        RandAugment & (ops=3, magnitude=9) \\
        Label Smoothing & 0.05 \\
        Learning Rate Schedule & cosine \\
        Optimizer & AdamW \cite{loshchilov2019decoupledweightdecayregularization} \\
        drop out rate & 0.3 \\
        drop out path rate & 0.3 \\
        \bottomrule
    \end{tabular}
\end{table}

We used different learning rates for each of the backbones, as listed in Table~\ref{table:appendix:learning_rates}. Specifically, for the ConvNeXt-tiny+AIM model a much smaller learning rate, compared to those used for the other models, is needed for the training to converge.

\begin{table}[H]
    \centering
    \renewcommand{\arraystretch}{1.3}
    \small
    \caption{General training setting used for training AIM.}
    \label{table:appendix:learning_rates}
    \begin{tabular}{@{}lc@{}}
        \toprule
        \textbf{Model} & \textbf{backbone Learning rate} \\
        \midrule
       AIM+ConvNeXt & 7e-6 \\
        ResNet50+AIM      & 0.001 \\
        ResNet101+AIM     & 0.001 \\
        \bottomrule
    \end{tabular}
\end{table}

\subsection{Computational Overhead}
\label{sec:appendix:computational_overhead}

Tab.~\ref{tab:computational_overhead} quantifies the computational overhead of our proposed \methodName module. The integration results in a marginal increase in both GFLOPs and model parameters, confirming the method's efficiency.

\begin{table}[H]
    \centering
    \caption{\footnotesize Comparison of GFLOPs and the Number of Parameters with Increment Over Baseline Models 
    }
    \scriptsize
    \scalebox{1.2}{
    \begin{tabular}{l c c}
        \toprule
        \textbf{Model Name} & \textbf{GFLOPs} & \textbf{Parameters (M)} \\
        \midrule
        ConvNeXt-tiny & 4.5 & 28.0 \\
        ConvNeXt-tiny+\methodName [1] & 5.2 (+0.7) & 30.7 (+2.7) \\
        ConvNeXt-tiny+\methodName [2] & 4.6 (+0.1) & 29.9 (+1.9) \\
        \midrule
        ResNet50 & 4.1 & 23.9 \\
        ResNet50+\methodName [2] & 5.1 (+1.0) & 27.6 (+3.7) \\
        ResNet50+\methodName [3] & 4.4 (+0.3) & 26.6 (+2.7) \\
        \bottomrule
    \end{tabular}
    }
    
    \label{tab:computational_overhead}
\end{table}

\subsection{Active-area loss threshold annealing}
\label{sec:appendix:implementation_details-Active-area loss threshold annealing}

\begin{figure}[H]
        \centering
        \includegraphics[width=0.4\linewidth]{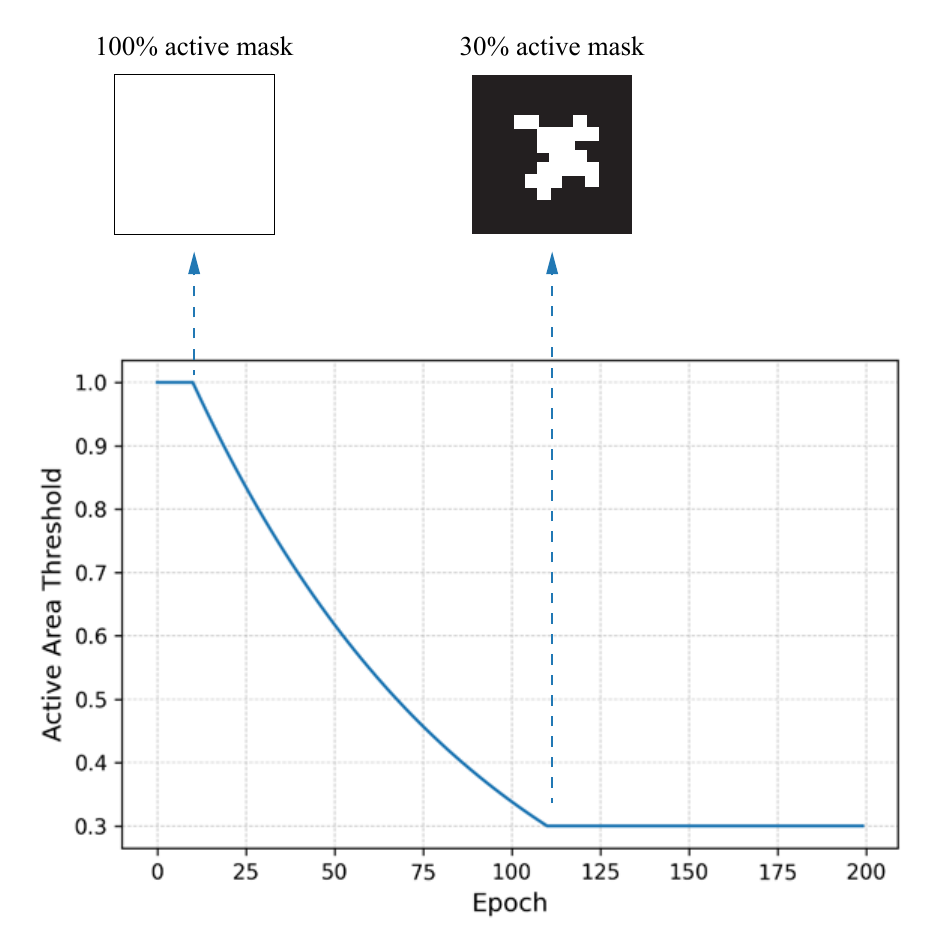}
        \caption{The annealing process of the active-area loss threshold begins either at the start of training or at a specific epoch (e.g., epoch 10 in the figure) and continues for a set number of epochs (e.g., 100 epochs, which is half of the total 200 training epochs). After this period, a final threshold of 0.3 is maintained for the remainder of the training.}
        \label{fig:Thresholding_annealing}
    \end{figure}

We employ a masking annealing technique to ensure a seamless adaptation of the network to the masking process. The main idea of this technique is to increase the masking or decrease the number of active elements (elements with value 1 in the binary mask) as the training evolves (see Figure \ref{fig:Thresholding_annealing}). This is done by controlling the active-area loss threshold throughout the training process, where we start with fully active masks (a threshold of 1.0), enabling the entire network to operate without constraints, and as training advances, we decrease that threshold with each epoch until reaching the final wanted value (for example 0.35 which means 35\% of the mask is only active). The network then uses this final value for the rest of the training. This annealing technique aids the network in adjusting more effectively to the masking constraints, resulting in improved mask quality and a more stable learning process.
The number of epochs or steps for which the annealing of the active-area threshold is carried out is treated as a hyperparameter.

\subsection{Datasets}
\label{sec:appendix:implementation_details-Datasets}
In the following sections, we provide details on the datasets used in our experiments: CUB, Waterbirds-100\%, Waterbirds-95\%, and TravelingBird.

\subsubsection{CUB-200}
\label{sec:appendix:implementation_details-Datasets-CUB200}
The Caltech-UCSD Birds-200-2011 \cite{WahCUB_200_2011}, known as CUB-200-2011, is one of the most well-known datasets in the fine-grained visual classification domain. The dataset consists of 11,788 images of 200 bird species, roughly divided in half for training and half for evaluation. 
There are an average of 30 images per class in the training dataset and 29 images per class in the test dataset. 
Throughout this work, we divided the primary training dataset into 80\% and 20\% training and validation datasets to search over hyperparameters. In our experiments, we utilized an input image size of (224 $\times$ 224) pixels, and for computing the EPG scores, we relied on the binary segmentation \cite{farrell_2022} masks.
\vspace{1em}

 \subsubsection{WaterBirds} 
\label{sec:appendix:implementation_details-Datasets-WaterBirds}
 Waterbirds dataset is a synthetic dataset \cite{sagawa2019distributionally} designed to test classification models by introducing a controllable distribution shift, it is specifically engineered to assess how models respond to shifts in group distributions. The authors took the different bird species from the CUB-200-2011 dataset and simplified the task to be a two-class classification: Waterbirds and Landbirds.
 They manipulate the backgrounds using the binary segmentation masks \cite{farrell_2022}, where they replaced the original background with either water or land scenes taken from the Places dataset \cite{7968387}. This creates four different groups: Landbirds on land-background, Landbirds on water-background, Waterbirds on land-background, and Waterbirds on water-background.

The ability to control the construction of the dataset allows researchers to explore how models handle spurious correlations. In the training dataset, the majority of images fall into main groups - Landbirds on land and Waterbirds on water. However, in the validation and test datasets there is an equal distribution across the four groups, creating a deliberate distribution shift between training and test datasets.

The two mainly used versions of this dataset are, the Waterbirds100\% version \cite{petryk2022guiding}, which presents the most extreme challenge, with training dataset that have a perfect correlation between the type of the bird and the background-Water bird on Water background and Land bird on Land background. The other version is the Waterbirds95\% \cite{sagawa2019distributionally}  where 5\%  of the training images come from the out-of-distribution groups: Landbird on Water-background and Waterbird on land-background. In our experiments, we utilized an input image size of (224 $\times$ 224) pixels. For computing the EPG scores, we also relied on the binary segmentation masks \cite{farrell_2022} of the CUB-200 dataset.
\vspace{1em}
    
\subsubsection{TravelingBirds}
\label{sec:appendix:implementation_details-Datasets-TraverlingBirds}
TravelingBirds dataset \cite{koh2020concept} is a variant of the CUB dataset and it is constructed in a similar way to that of Waterbirds dataset. While it preserves the original 200 classes of the CUB-200-2011 dataset, it changes the background of the birds to spuriously correlate the target label y and the image background within the training set only. The Authors used the binary segmentation masks to isolate the bird's pixels from its original background and put them onto a different background scene taken from the Places dataset \cite{7968387}. Each bird species is laid out on a unique but randomly selected type of scene. During test time, the association between bird label and their background scene type is randomized, Completely disrupting the training set's correlation between background and class labels, resulting in a challenging adversarial setting.
Following \cite{koh2020concept}, we utilized an input image size of (299 × 299) pixels, and for computing the EPG scores, we also relied on the binary segmentation masks \cite{farrell_2022} of the CUB-200 dataset.

\subsubsection{ImageNet-100}
\label{sec:appendix:implementation_details-Datasets-ImageNet100}
ImageNet-100~\cite{imagenet100} is a widely used~\cite{hoffmann2021towards,agnihotriroll} subset of the full ImageNet Large Scale Visual Recognition Challenge (ILSVRC) 2012 dataset \cite{imagenet}. It is composed of 100 distinct object classes selected from the original 1000, containing approximately 128,000 training images and 5,000 validation images. The primary motivation for using this subset is to enable faster model training, hyperparameter tuning, and experimentation compared to the resource-intensive full dataset, while still offering a diverse and challenging multi-class classification benchmark. In our experiments, we follow standard practice and use an input image size of $(224 \times 224)$ pixels.
\vspace{1em}

\subsubsection{ImageWoof}
\label{sec:appendix:implementation_details-Datasets-ImageWoof}
ImageWoof \cite{imagewoof} is a challenging 10-class subset of ImageNet \cite{imagenet} designed for rapid yet difficult experimentation. The dataset is intentionally curated to be a hard, fine-grained classification problem, as it contains 10 breeds of dogs that are visually very similar. By focusing on these hard-to-distinguish classes, ImageWoof provides a computationally inexpensive benchmark that is more demanding than a random subset of ImageNet of a similar size. This makes it particularly useful for quickly iterating on and evaluating new model architectures and training methodologies. In our experiments, we utilize an input image size of $(224 \times 224)$ pixels.

\subsubsection{Hard-ImageNet}
\label{sec:appendix:implementation_details-Datasets-ImageNetHard}
ImageNet-Hard \cite{hardimagenet} is a benchmark designed to evaluate if models classify images "for the right reasons" rather than relying on spurious correlations. It consists of 19k training images and 750 quintuply-validated test images across 15 classes: Dog sled, Howler Monkey, Seat Belt, Ski, Sunglasses, Swimming Cap, Balance Beam, Horizontal bar, Patio, Hockey Puck, Miniskirt, Space Bar, Volleyball, Baseball Player, and Snorkel.

The benchmark challenges models with images where the object of interest is captured under suboptimal conditions (e.g., not large or centered). Leveraging segmentation masks helps diagnose whether a model's prediction is based on the object itself or on misleading background cues, thus providing a deeper understanding of model behavior beyond simple accuracy metrics. For evaluation, we use an input size of $(224 \times 224)$ pixels.

\subsection{Energy Pointing Game (EPG) Score}
\label{sec:appendix:implementation_details-EPG}

The Energy Pointing Game (EPG) score~\cite{wang2020score} is a metric designed to quantify the spatial localization capabilities of a model. Specifically, it evaluates the extent to which the model's attribution aligns with the ground-truth object regions.

The computation of the EPG score requires the following components:
\begin{itemize}
\item \textbf{Attribution Map:} An importance map generated by an attribution method such as Saliency Maps~\cite{saliency_map}, GradCAM~\cite{selvaraju2017grad}, Guided GradCAM~\cite{Guided_grad_cam}, or any comparable technique.
\item \textbf{Ground-Truth Binary Mask:} A binary segmentation mask that delineates the region corresponding to the object of interest.
\end{itemize}

The EPG score is calculated as the ratio between the total attribution energy within the active (foreground) region of the binary mask and the total attribution energy across the entire attribution map. A higher EPG score indicates that the model concentrates its decision-making evidence within the relevant object regions, thereby demonstrating better spatial grounding of its predictions.

\subsection{Testing the center bias of the masks}
\label{sec:appendix:implementation_details-Testing_the_center_bias_of_the_masks}
\begin{figure}[H]
        \centering
        \includegraphics[width=0.5\linewidth]{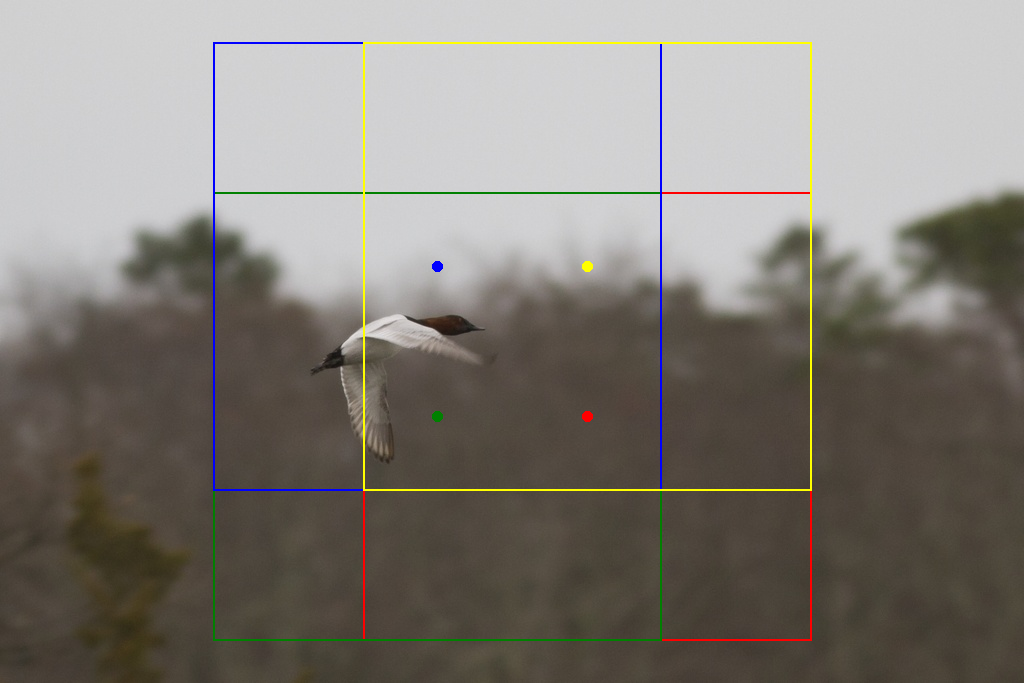}
        \caption{The image illustrates the shifted cropping technique used to assess the center bias of the generated masks. Each colored dot represents the center of its corresponding square crop, which is outlined in the same color. For each image, one of the four centers is randomly selected, and the image is cropped accordingly.}
        \label{fig:appending:masks_center_bias}
    \end{figure}

One of the concerns we had after seeing the masks that our method generated on the CUB-200 dataset was the susceptibility of our models' masks to center bias. Thus, rather than center-corp the images, we designed a cropping mechanism that, given a specific deviation from the center, randomly chose one new center for cropping in a way that guarantees to violate the center bias of the object in the newly cropped image. We denote this modified dataset as the shifted-center CUB-200. Figure~\ref{fig:appending:masks_center_bias} illustrates the shifted cropping technique used to assess the center bias of the generated masks. Each colored dot represents the center of its corresponding square crop, which is outlined in the same color. For each image, one of the four centers is randomly selected, and the image is cropped accordingly.

It is important to note that this cropping technique might also crop important parts from the object of interest, making the classification much harder due to the lack of task-related features. For example, the body might be cropped out while only the legs of the bird are kept. \vspace{1em}

\section{Quantitative results}
\label{sec:appendix:Quantitative_results}
This section provides a comprehensive overview of the performance metrics obtained from our trained models under different configurations.

\subsection{Center bias}
\label{sec:appendix:Quantitative_results-Center_Bias}

Table~\ref{table:appendix_cub_shifted_results} presents the performance metrics for the baseline vanilla ConvNeXt-tiny model as well as the ConvNeXt-tiny+AIM models on the Shifted-center CUB-200 dataset. Notably, both the vanilla ConvNeXt-tiny and all variants of the AIM architecture experienced a performance drop, with the ConvNeXt-tiny showing a decrease of approximately 10\% and the AIM variants showing a decline of around 9\% compared to the center-cropped experiments (see Figure~\ref{tab:models_comparison}). Despite this reduction in accuracy, our proposed model variants still outperform the baseline ConvNeXt-tiny model by nearly 2\%. For qualitative comparison of the GradCAM and the generated masks see Figure~\ref{fig:appendix_center_bias}

\begin{table}[H]
\centering
\caption{presents the performance metric scores for the baseline vanilla ConvNeXt-tiny model as well as the ConvNeXt-tiny+AIM models on the Shifted-center CUB-200 dataset. Notably, both the vanilla ConvNeXt-tiny and all variants of the AIM architecture experienced a performance drop, with the ConvNeXt-tiny showing a decrease of approximately 10\% and the AIM variants showing a decline of around 9\% compared to the center-cropped experiments (see~\Cref{tab:appendix_models_comparison}). Despite this reduction in accuracy, our proposed model variants still outperform the baseline ConvNeXt-tiny model by nearly 2\%. For qualitative comparison of the GradCAM and the generated masks see Figure~\ref{fig:appendix_center_bias}.}
    \begin{tabular}{@{}lc@{}}
        \toprule
        \textbf{Model} &  \textbf{Test Accuracy ($\pm$ Std)} \\ \midrule

        ConvNeXt-tiny & 76.98 ($\pm$0.18) \\ \midrule

       AIM+ConvNeXt (1, 25\%)           &  79.08 ($\pm$0.44) \\ 
       AIM+ConvNeXt (1, 35\%)           & 79.33 ($\pm$0.45) \\ 
       AIM+ConvNeXt (1, No Annealing)   &  79.13 ($\pm$0.35) \\ \midrule

       AIM+ConvNeXt (2, 25\%)           &  79.11 ($\pm$0.12) \\ 
       AIM+ConvNeXt (2, 35\%)           &  79.11 ($\pm$0.12) \\ 
       AIM+ConvNeXt (2, No Annealing)   &  79.12 ($\pm$0.13) \\ 

        \bottomrule
    \end{tabular}
    
    \label{table:appendix_cub_shifted_results}
\end{table}

Upon analyzing the generated masks on the cropped test images (see Figures \ref{fig:appendix_center_bias}), we can see that large bird regions are missing, which can attribute the performance reduction to the loss of essential parts during the cropping process. The masks clearly focus on the task-related regions or what's left of them after cropping. This is particularly evident in the masks generated by stage\#2 in the second example, where the model focuses on the bird's legs and belly.

\subsection{Attribution-Agnostic Localization Performance:}
\label{sec:appendix:attribution_agnostic}
In addition to GradCAM, we evaluate other attribution methods to assess the generality of our localization improvements. Table~\ref{tab:epg_other_methods} reports EPG scores on the Waterbirds-95\% dataset using Guided GradCAM and Guided Backprop, both of which show consistent gains with \methodName.

    \begin{table}[H]
        \centering
        \caption{Attribution-agnostic EPG improvements on Waterbirds-95\%. \methodName consistently boosts localization scores regardless of the attribution method used.}
        \begin{tabular}{lcc}
        \toprule
            \textbf{Model} & \multicolumn{2}{c}{\textbf{EPG score} $\uparrow$} \\ 
            \cmidrule(lr){2-3}
             & \textbf{Guided GradCAM} & \textbf{Guided Backpropagation} \\ \toprule
            Vanilla ConvNext-tiny       & 46.17($\pm$ 2.5)  & 31.26 ($\pm$ 5.4) \\ 
            ConvNext+AIM (2, 25\%)  &  \textbf{76.77 ($\pm$ 2.7)}  & \textbf{51.89 ($\pm$ 0.3)}  \\
            \bottomrule
        \end{tabular}
        
        \label{tab:epg_other_methods}
    \end{table}

\subsection{Extended Results of ConvNeXt+AIM:}
\label{sec:appendix:extended_results_of_convnext_aim}
In this section, we provide extended results for our AIM-enhanced ConvNeXt model, as detailed in Table~\ref{tab:appendix_models_comparison}. We supplement our findings on the Waterbirds dataset by including the overall accuracy metric, which further highlights the benefits of our approach. For instance, the best-performing AIM model achieves an overall accuracy of $89.1\% \pm 0.8\%$ on the Waterbirds (100\%) benchmark, a significant increase from the baseline's $71.2\% \pm 2.2\%$. Furthermore, we present results on the CUB-200 dataset, where AIM-based models also outperform the baseline model.

\begin{table}[H]
    \centering
    \setlength{\abovecaptionskip}{1pt}   
    \setlength{\belowcaptionskip}{0pt}
     \caption{
                {\footnotesize \textbf{Average Test Accuracies for different configurations of ConvNeXt+AIM.} Comparison of the \textbf{ConvNeXt-tiny} baseline against our AIM-enhanced models on three benchmarks. Our method achieves substantial gains, most notably improving the \textbf{worst-group accuracy} on the Waterbirds dataset, which highlights its effectiveness in mitigating spurious correlations. All values are mean accuracy (\%) $\pm$ standard deviation.}}
    \scalebox{0.65}{
    \begin{tabular}{l c c c c c c c c c c}
        \toprule
        \multirow{3}{*}{\textbf{Model}} & \multicolumn{2}{c}{\textbf{CUB}} & \multicolumn{6}{c}{\textbf{Waterbirds}} & \multicolumn{2}{c}{\textbf{Travelingbirds}}
        \\
        \cline{4-9}
        & & & \multicolumn{3}{c}{\textbf{100\%}} & \multicolumn{3}{c}{\textbf{95\%}} & & \\
        & \textbf{Acc} & \textbf{EPG} & \textbf{WG-Acc} & \textbf{Overall Acc} & \textbf{EPG} & \textbf{WG-Acc} & \textbf{Overall Acc} & \textbf{EPG} & \textbf{Acc} & \textbf{EPG}\\
        \midrule
        ConvNeXt-t                       & 87.9 ($\pm$0.02)     &      68.3 ($\pm$0.1)   & 39.6 ($\pm$5.4)                           &  71.2	($\pm$2.2)    & 57.19 ($\pm$6) & 81.6 ($\pm$3.2)                            & 94.8	($\pm$ 0.3) & 68.3	($\pm$ 3.2) & 59.5 ($\pm$0.8) & 74.4 ($\pm$0.6)\\

        \textbf{ConvNeXt-t+AIM[1, 25\%]}     & \textbf{88.6 ($\pm$0.1)}  & \textbf{76.804 ($\pm$3.3)}  &  
        73.6 ($\pm$4.5)  & 86.2 ($\pm$2.6)  & 60.11 ($\pm$1.3) & 
        91.2 ($\pm$ 0.8)  & 95.7 ($\pm$ 0.7) & \textbf{77.1 ($\pm$5.15)} & 
        77.1 ($\pm$0.3) & 79 ($\pm$0.7)\\
        
        \textbf{ConvNeXt-t+AIM[1, 35\%]}     & 88.18 ($\pm$0.1)  &55 ($\pm$3.8)  & 
        77.1 ($\pm$4.4)  &  88.4	($\pm$1.8)  & 57.2 ($\pm$1.3) & 
         90.7 ($\pm$0.7)  & 96 ($\pm$ 0.3)  & 63 ($\pm$1.2) & 
        71.5 ($\pm$1.3) &  72.6 ($\pm$1.5)\\

        \textbf{ConvNeXt-t+AIM[2, 25\%]}     & 87.78 ($\pm$0.2)  & 75.17($\pm$1.2)  
        & 74 ($\pm$5)  &  84($\pm$6)  & 58 ($\pm$1.3)
        & \textbf{92.7 ($\pm$1.2)}  & \textbf{96.6 ($\pm$ 0.2)} & 75 ($\pm$6)
        & \textbf{77.4	($\pm$0.2)} & \textbf{85 ($\pm$2)}\\ 
        
        \textbf{ConvNeXt-t+AIM[2, 35\%]}     & 88.5 ($\pm$0.2)   &  62.5 ($\pm$0.3) 
        & \textbf{78.1 ($\pm$2.3)}  &  \textbf{89.1	($\pm$0.8)}   &   \textbf{68.5 ($\pm$3.6)}
        & 92.3 ($\pm$0.6)  & 96.31 ($\pm$ 0.1) & 71.7 ($\pm$ 6.4)
        & 71 ($\pm$0.4) & 77.7 ($\pm$0.4) \\
        \bottomrule
    \end{tabular}
        }
    
    \label{tab:appendix_models_comparison}

\end{table}

\subsection{Comparison To Other Methods:}
\label{sec:appendix:other_methods_comparison}

This section compares AIM to other methods. Unlike our approach, which does not use any form of guidance, other methods often rely on guidance mechanisms to direct the trained model. Examples of such guidance include using binary masks~\cite{rao2023studying} or leveraging the output of a CLIP-based model controlled by a text prompt GALS~\cite{petryk2022guiding} or using specific loss to account for imbalanced data~\cite{van2023pdisconet}.
Our proposed \methodName solely relies on image labels used by traditional image classification methods.

\begin{table}[H]
    \centering
    \caption{
                {\footnotesize
                \textbf{Average Test Accuracies for Various Models with and without \methodName Modification.} This table presents the average test accuracies (with standard deviations) for different models on the CUB-200, Waterbirds, and Travelingbirds datasets. The models include ConvNeXt-tiny, ResNet50, and ResNet101 with and without the \methodName modification. The AIM-enhanced models are denoted as \textit{[backbone]+AIM (stage index, mask active-area loss)}, such as ConvNeXt-tiny+AIM(1, 25\%). The results are organized by dataset and further divided into categories: overall accuracy and worst-group accuracy for both 100\% and 95\% subsets where applicable. Improvements or declines in performance due to the \methodName modification are highlighted in green and red, respectively. This comprehensive comparison provides insights into the effectiveness of the \methodName approach in enhancing model performance across different datasets and scenarios.}}
    \setlength{\abovecaptionskip}{1pt}   
    \setlength{\belowcaptionskip}{0pt}
     
    \scalebox{0.8}{
    \begin{tabular}{l c c c c c c }
        \toprule
        \multirow{3}{*}{\textbf{Model}} & \multirow{3}{*}{\textbf{CUB-200 (\%)}} & \multicolumn{4}{c}{\textbf{Waterbirds (\%)}} & \multirow{3}{*}{\textbf{Travelingbirds (\%)}}
        \\
        \cline{3-6}
        & & \multicolumn{2}{c}{\textbf{100\%}} & \multicolumn{2}{c}{\textbf{95\%}} & \\
        & & \textbf{Worst-Group} & \textbf{Overall} & \textbf{Worst-Group} & \textbf{Overall} & \\
        \midrule
        ConvNeXt-tiny                       & 87.9 ($\pm$0.02)                           & 39.6 ($\pm$5.4)                           &  71.2	($\pm$2.2)                           & 81.6 ($\pm$3.2)                            & 94.8	($\pm$ 0.3) &  59.5 ($\pm$0.8)\\

        \textbf{ConvNeXt-t+AIM[1, 25\%]}     & \textbf{88.6 ($\pm$0.1)}    &  
        73.6 ($\pm$4.5)  & 86.2 ($\pm$2.6)   & 
        91.2 ($\pm$ 0.8)  & 95.7 ($\pm$ 0.7)  & 
        77.1 ($\pm$0.3) \\
        
        \textbf{ConvNeXt-t+AIM[1, 35\%]}     & 88.18 ($\pm$0.1)  & 
        77.1 ($\pm$4.4)  &  88.4	($\pm$1.8)   & 
         90.7 ($\pm$0.7)  & 96 ($\pm$ 0.3)  & 
        71.5 ($\pm$1.3) \\

        \textbf{ConvNeXt-t+AIM[2, 25\%]}     & 87.78 ($\pm$0.2)    
        & 74 ($\pm$5)  &  84($\pm$6)  
        & \textbf{92.7 ($\pm$1.2)}  & \textbf{96.6 ($\pm$ 0.2)} 
        & \textbf{77.4	($\pm$0.2)} \\ 
        
        \textbf{ConvNeXt-t+AIM[2, 35\%]}     & 88.5 ($\pm$0.2)    
        & \textbf{78.1 ($\pm$2.3)}  &  \textbf{89.1	($\pm$0.8)}   
        & 92.3 ($\pm$0.6)  & 96.31 ($\pm$ 0.1) 
        & 71 ($\pm$0.4)  \\ \midrule

        DRO~\cite{van2023pdisconet} + ResNet50  & - & - & - & 91.4 ($\pm$1.1) & - & - \\
        GALS~\cite{petryk2022guiding} + ResNet50   & - & \textbf{56.71} (-) & - & 76.54 (-) & - & - \\
        BCos-ResNet50~\cite{rao2023studying}  & - & 56.1 ($\pm$4) & - & - & - & - \\
        ResNet50 & 81.32 & 41.29 (± 1.99)   & 69.48 (± 2) & 71.09 (± 5.98) 	 & 91.21 (± 0.27) & 50.21 (± 0.5) \\
        \textbf{ResNet50+AIM (2, 25\%) }         & 83.90 & 52.19 (± 9.11) & 73.84 (± 1.14) & 75.97 (± 2.344) & 92.1 (± 0.31) & 65.2 (± 0.24) \\
        \textbf{ResNet50+AIM (3, 25\%)}          & 83.16 & 40.12 (± 2.1) & 71.36 (± 1.7) & 77.2875 (± 0.404) & 92.38 (± 0.2) & 64 (± 0.51) \\
        \midrule

        ECBM~\cite{xu2024energy} + ResNet101 & - & - & - & - & - & \textbf{58.4} (-) \\
        ResNet101 & 76.41 (± 0.619)         & 36.48 (± 4.22) & 71.69 (± 2.2) & 77.565 (± 3.2) & 92.82 (± 0.31) & 19.54 (± 0.34) \\
        \textbf{ResNet101+AIM (2, 25\%) }        & 82.55 (± 0.22) & 38.11 (± 1.4) & 71.46 (± 0.53) & 82.39 (± 0.88)	 & 93.12 (± 0.74) & 44.03 (± 0.96) \\
        \textbf{ResNet101+AIM (3, 25\%) }        & 82.635 (± 0.28) & 38.81 (± 1.4) & 71.8 (± 0.69) & 82.39 (± 1.328) 	 & 93.34 (± 0.7) & 45.13 (± 0.7) \\
        \bottomrule
    \end{tabular}
        }
    
    \label{tab:append_models_comparison}

\end{table}

\section{Qualitative results}
\label{sec:appendix:Qualitative_results}

In this section, we explore the qualitative aspects of our study by showcasing masks and Grad-CAM attributions across different settings. These visualizations provide insights into the models' decision-making processes, highlighting areas of focus and variation in response to different configurations.

\subsection{Additional qualitative results:}
\label{sec:appendix:birds_data_qualitative_results}

\begin{figure}[ht]
    \setlength{\abovecaptionskip}{0pt}   
    \setlength{\belowcaptionskip}{0pt}   
    \centering
    \begin{tabular}{@{\hskip 2.9mm}c@{\hskip 2.9mm}c}
        \includegraphics[width=0.3\textwidth]{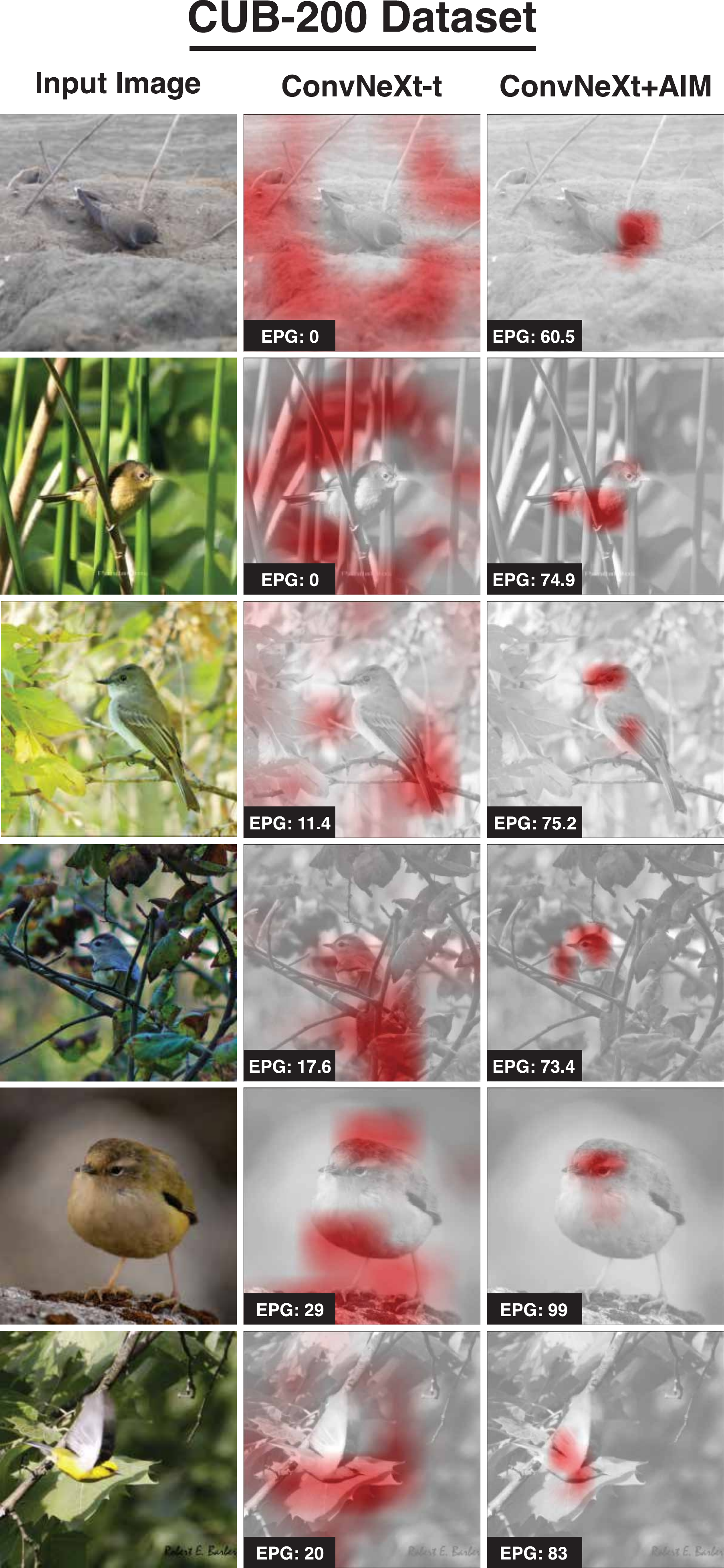} &
        \includegraphics[width=0.3\textwidth]{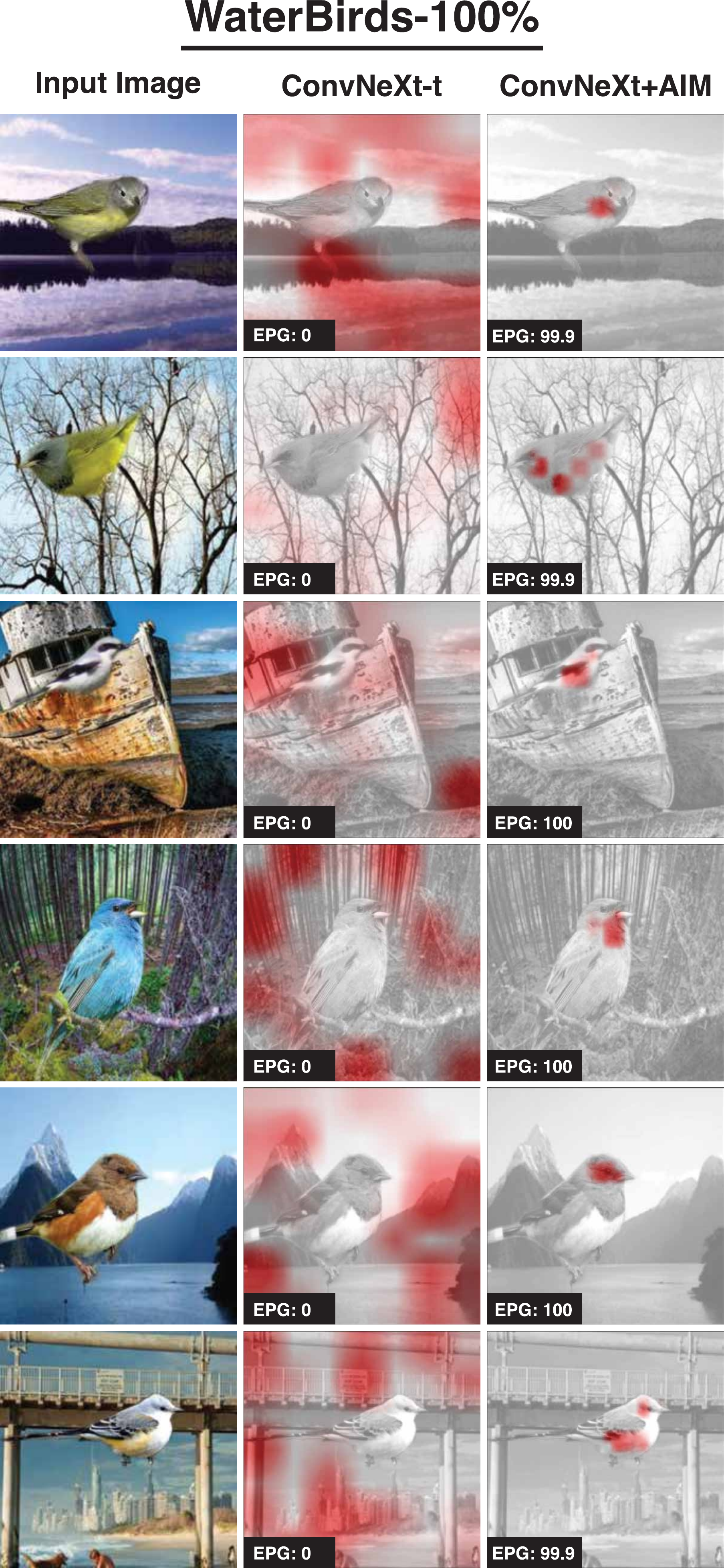} \\
        \includegraphics[width=0.3\textwidth]{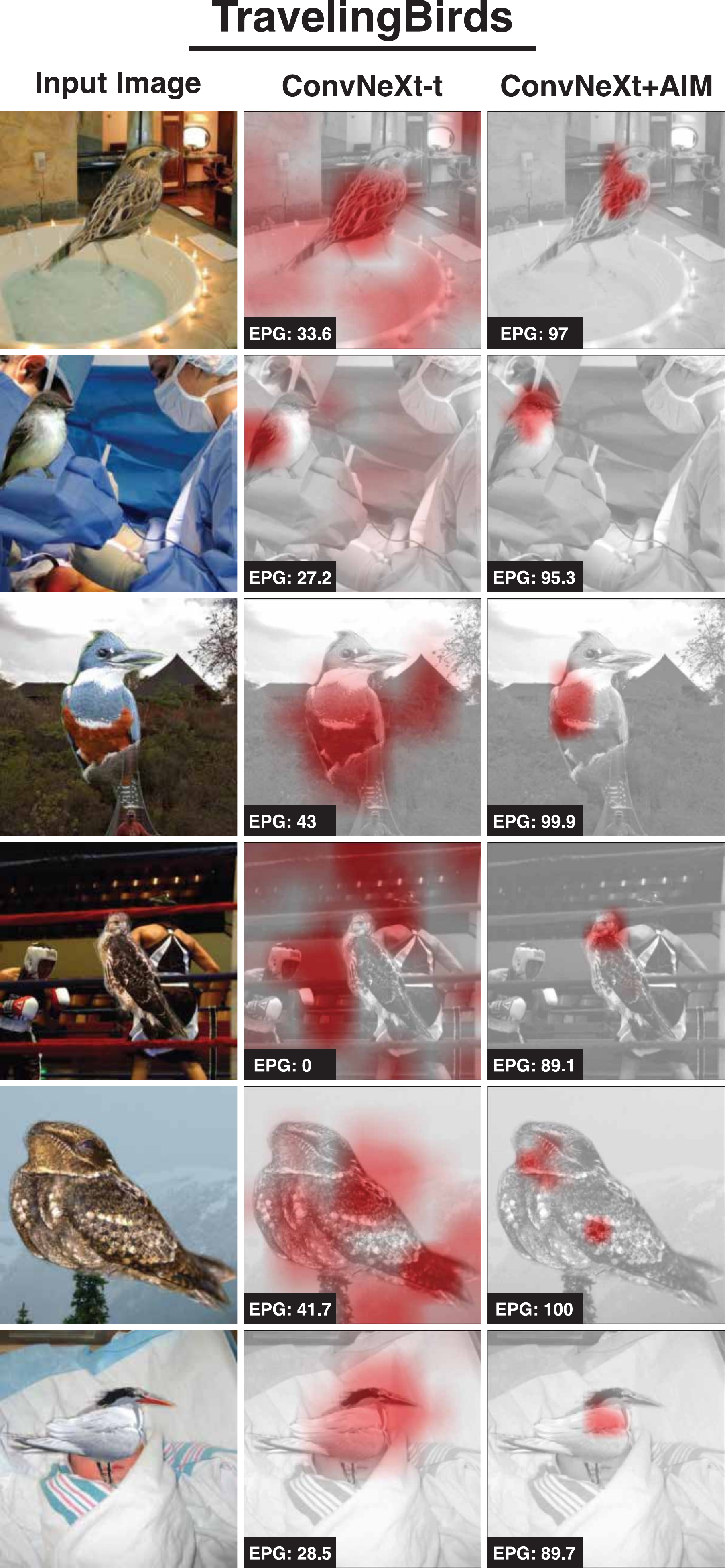} &
        \includegraphics[width=0.3\textwidth]{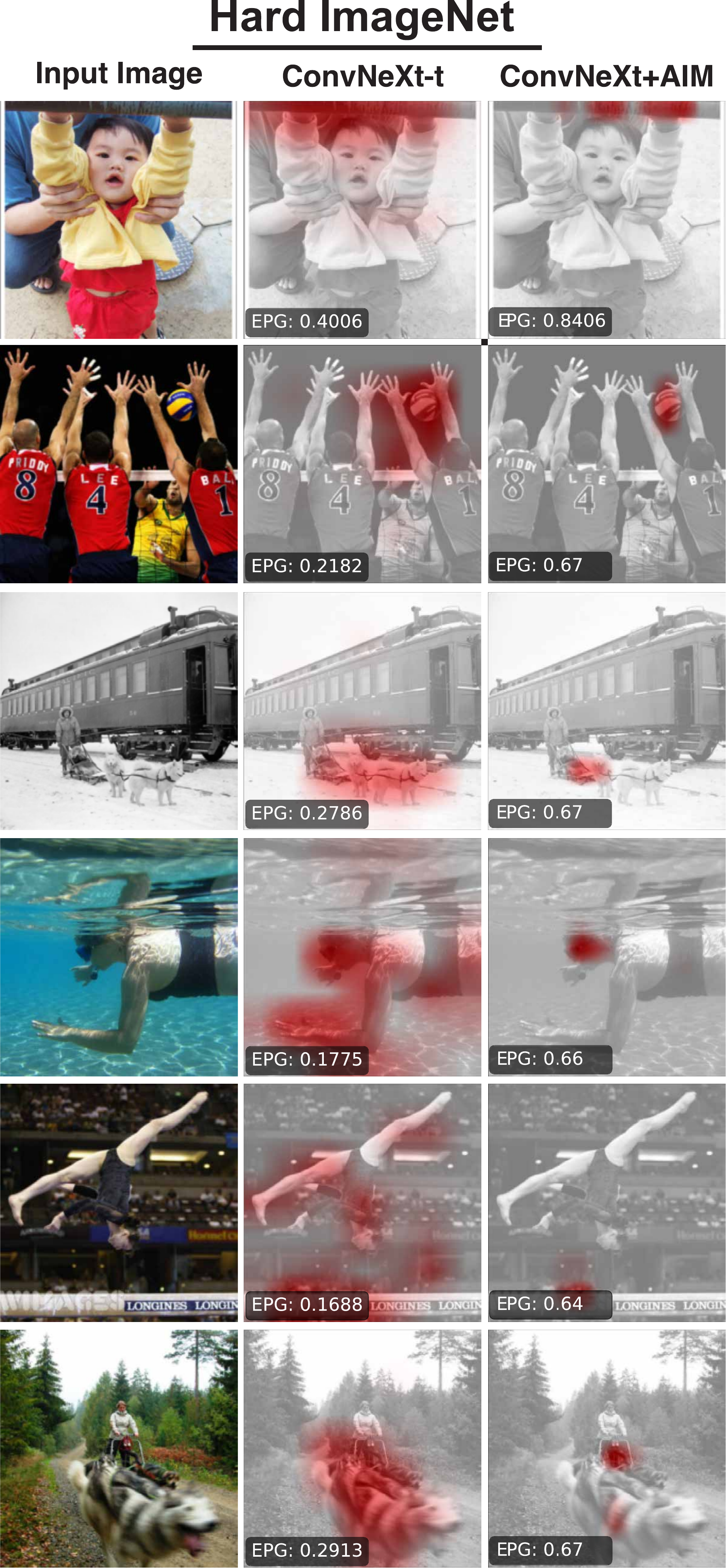}
    \end{tabular}
    \caption{\footnotesize \textbf{Models amended with AIM consistently exhibit enhanced localization of genuine features, effectively suppressing spurious cues in both in-domain and out-of-domain scenarios.} A qualitative visualization of Grad-CAM heatmaps comparing baseline ConvNeXt-tiny models and ConvNeXt-tiny+AIM models across CUB-200, TravelingBirds, WaterBirds-100\%, and Hard-ImageNet. The EPG scores are indicated on each heatmap.}
    \label{fig:appendix_GradCAM-vis}
\end{figure}

In the figure~\ref{fig:appendix_GradCAM-vis}, we illustrate additional examples from CUB, waterbirds-100\%, Travelingbirds and Hard-ImageNet datasets, where it is clear from the GradCAM attributes that AIM-based models focus on the object and ignore the spurious features.

\subsection{Center bias}
\label{sec:appendix:Qualitative_results:center_bias}
\begin{figure}[ht]
        \centering
        \includegraphics[width=1.0\textwidth]{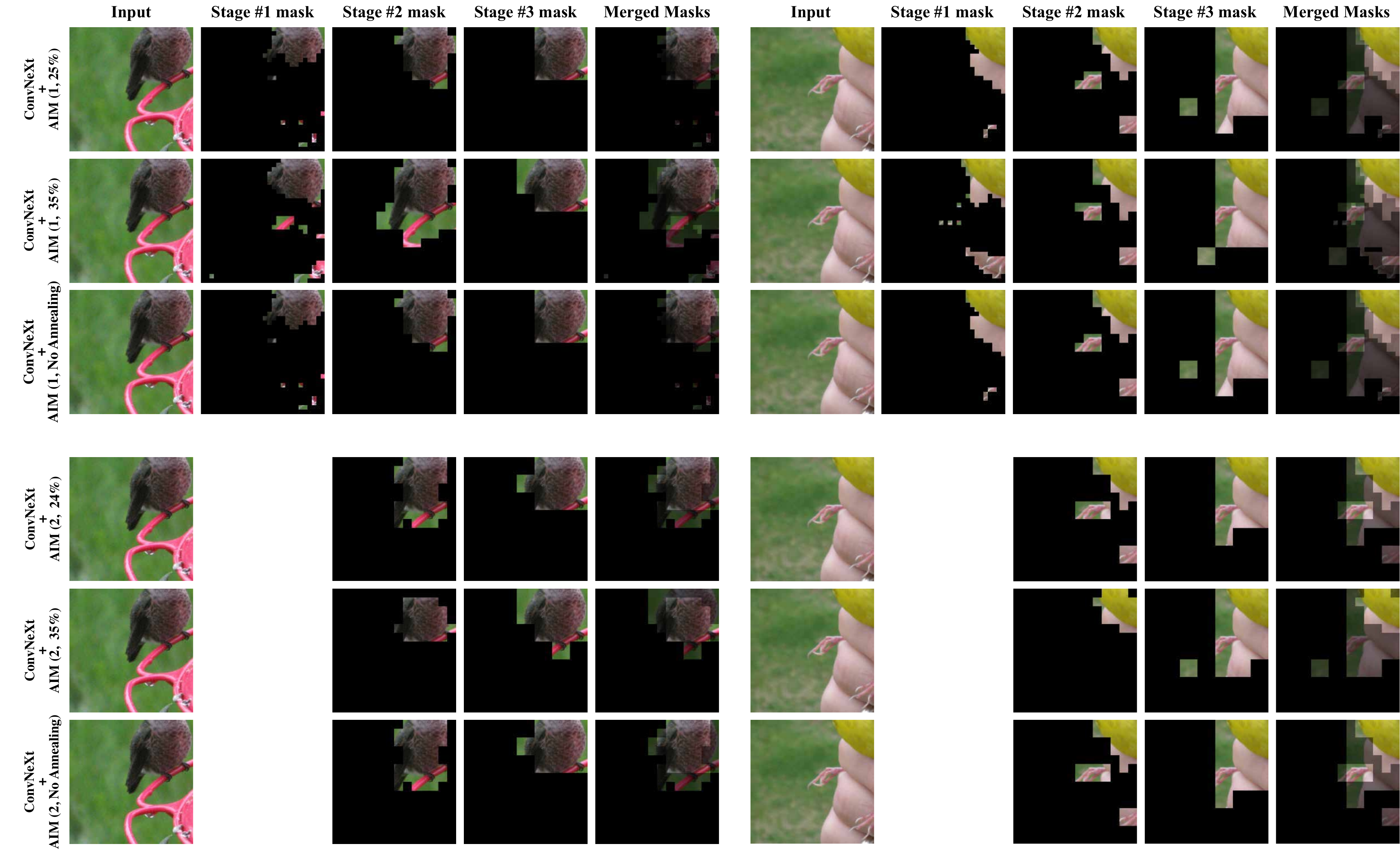}
        \caption{\textbf{Qualitative Comparison between the different architectural configurations of AIM on the Shifted-center CUB-200 setting.} This figure illustrates the masks generated at each stage for two primary architectural variants: ConvNeXt+AIM ($T$=1) (The group of images at the top) and ConvNeXt+AIM ($T$=2) (the group of images at the bottom), each utilizing different mask active-area thresholds (each row represent different active-area loss setting). The masks produced by AIM (columns denoted stage\#1 masks, stage\#2 masks, stage\#3 masks) explicitly delineate the regions utilized by the model at each stage, thereby demonstrating that AIM effectively mitigates susceptibility to center bias. The "Merged Masks" column demonstrates where the final feature maps will be zero, highlighting the discarded regions.}
        \label{fig:appendix_center_bias}
\end{figure}

Figure~\ref{fig:appendix_center_bias} presents a qualitative comparison of different architectural configurations of AIM on the Shifted-center CUB-200 setting. The figure illustrates the masks generated at each stage for two primary architectural variants: ConvNeXt-tiny+AIM (1), shown in the top group of images, and ConvNeXt-tiny+AIM (2), shown in the bottom group. Each variant employs different mask active-area thresholds, with each row representing a distinct threshold setting. The masks produced by AIM (columns denoted stage\#1 masks, stage\#2 masks, stage\#3 masks) explicitly delineate the regions utilized by the model at each stage, thereby demonstrating that AIM effectively mitigates susceptibility to center bias. The ``Merged Masks'' column demonstrates where the final feature maps will be zero, highlighting the discarded regions.

\subsection{Qualitative Comparison of Masks and Attribution Maps: AIM vs. Vanilla Backbone Models}
\label{sec:appendix:Qualitative_results:masks_attribution_maps}
In this section, we present qualitative results to illustrate the effectiveness of our proposed approach. We selected five different input images from each of the following datasets: CUB-200, Waterbirds-95\%, Waterbirds-100\%, and Traveling Birds. For these images, we compare the GradCAM attribution maps generated by the AIM+[backbone] models with different mask active-area thresholds to those produced by the vanilla [backbone] models. Additionally, we display the generated merged masks from the AIM+[backbone] models.

\begin{figure}[ht]
    \centering
    \begin{tabular}{ccc}
        \includegraphics[width=0.32\textwidth]{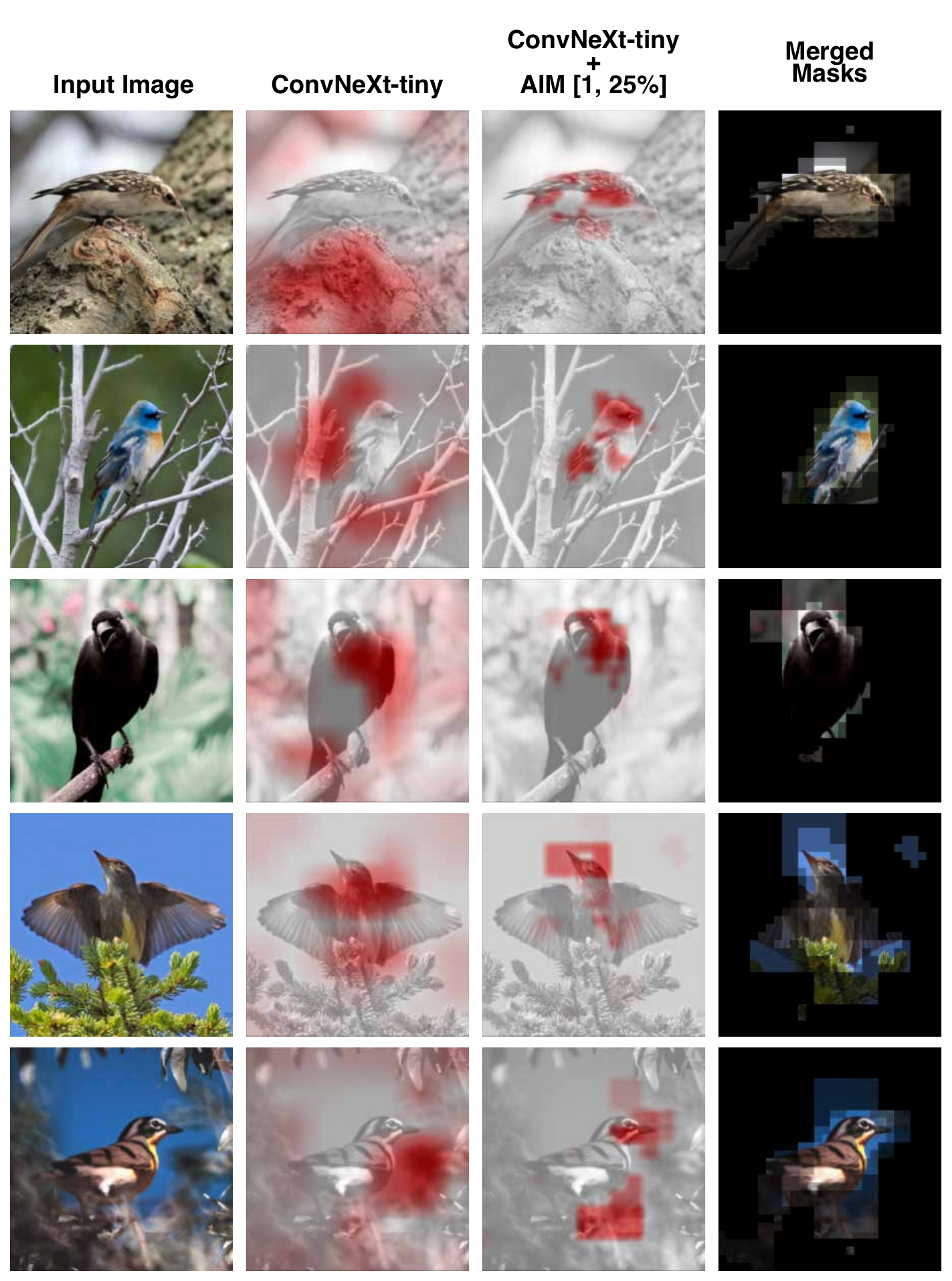} &
        \includegraphics[width=0.32\textwidth]{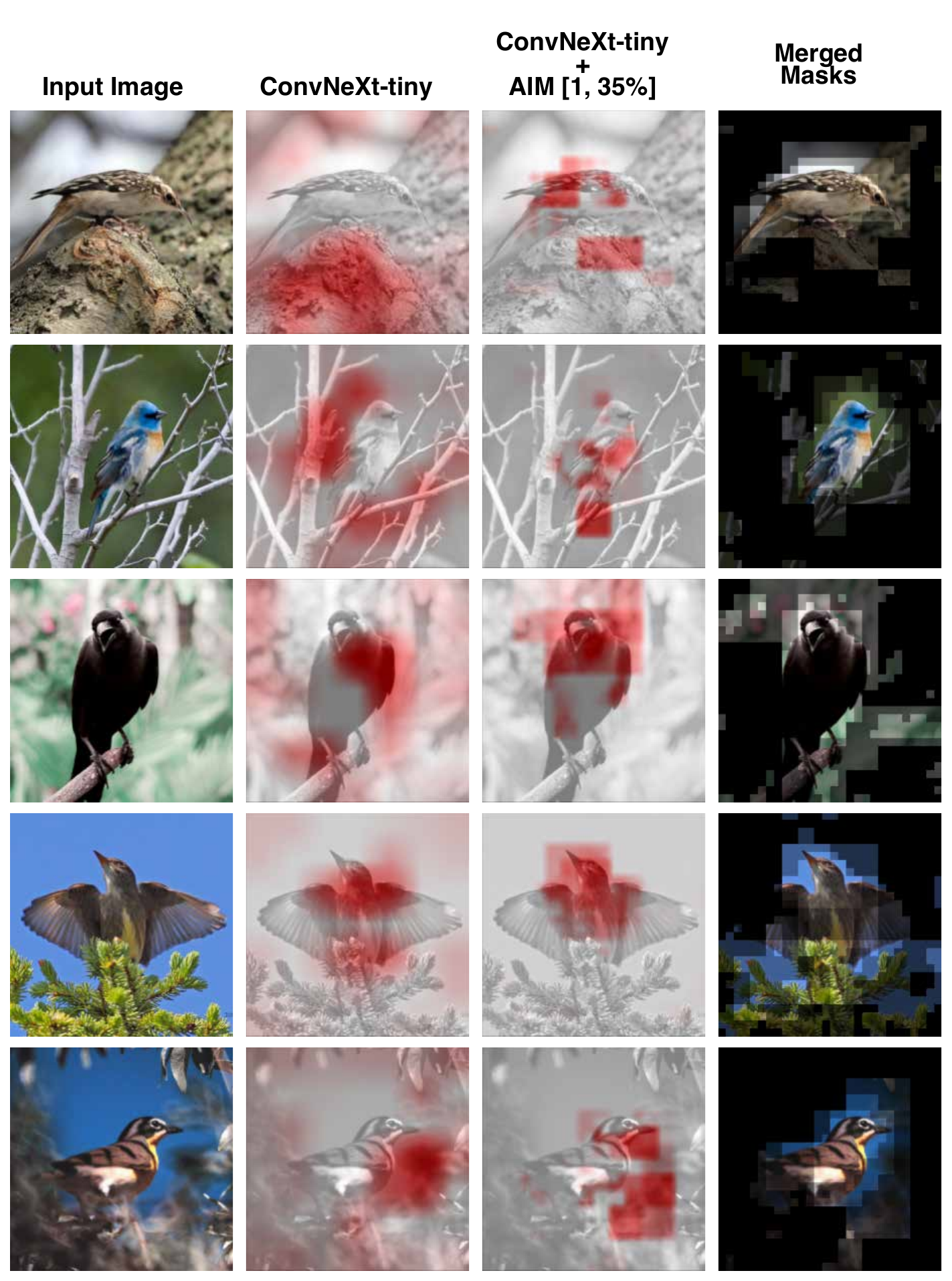} &
        \includegraphics[width=0.32\textwidth]{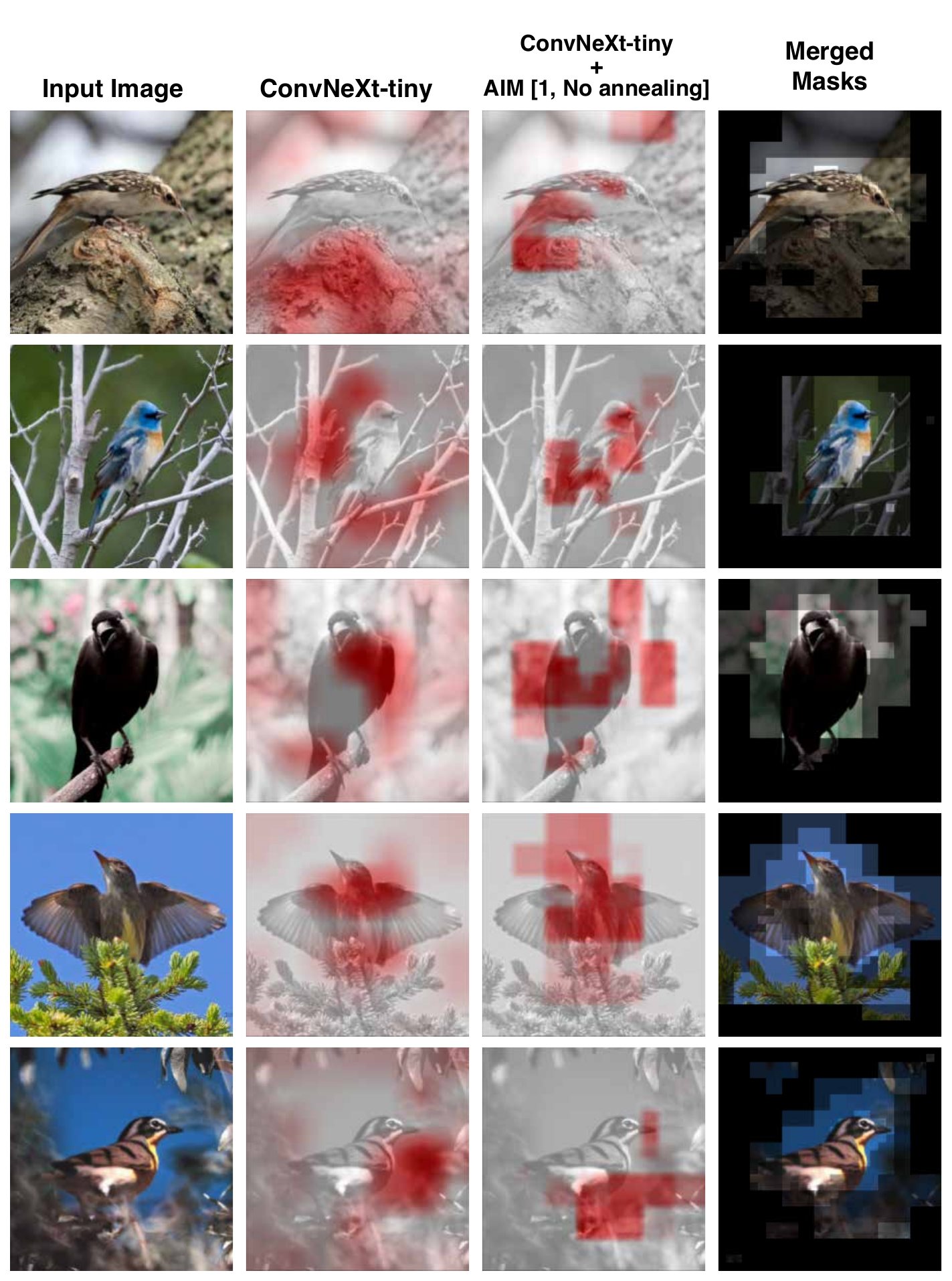}
    \end{tabular}
    \caption{\textbf{Qualitative results on the CUB-200 dataset.} Comparison of GradCAM attribution maps between ConvNeXt-tiny+AIM models with different mask active-area thresholds, along with the generated merged masks.}
    \label{fig:appendix-attribution_cub-ConvNeXt-b1}
\end{figure}

\begin{figure}[ht]
    \centering
    \begin{tabular}{ccc}
        \includegraphics[width=0.32\textwidth]{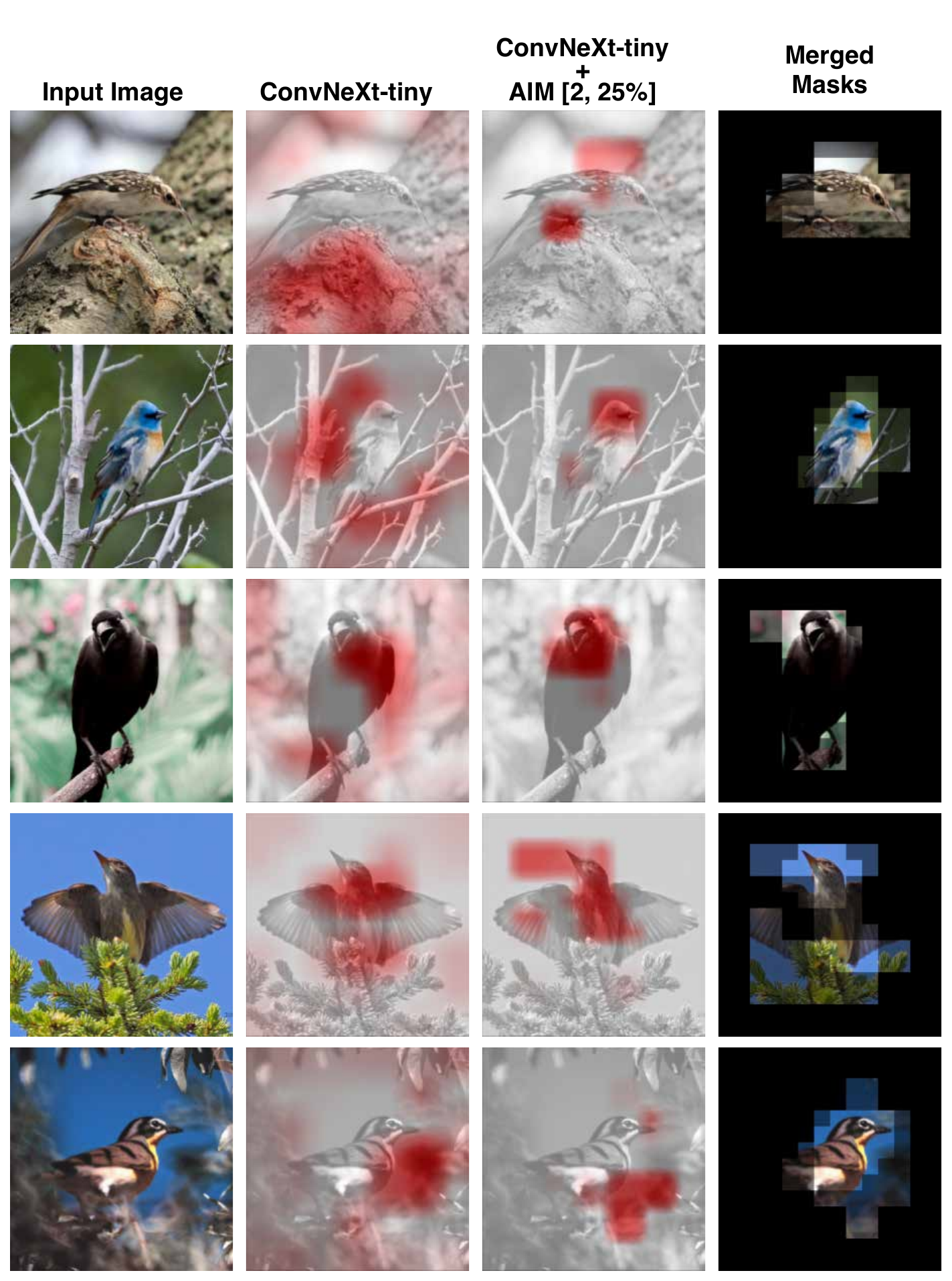} &
        \includegraphics[width=0.32\textwidth]{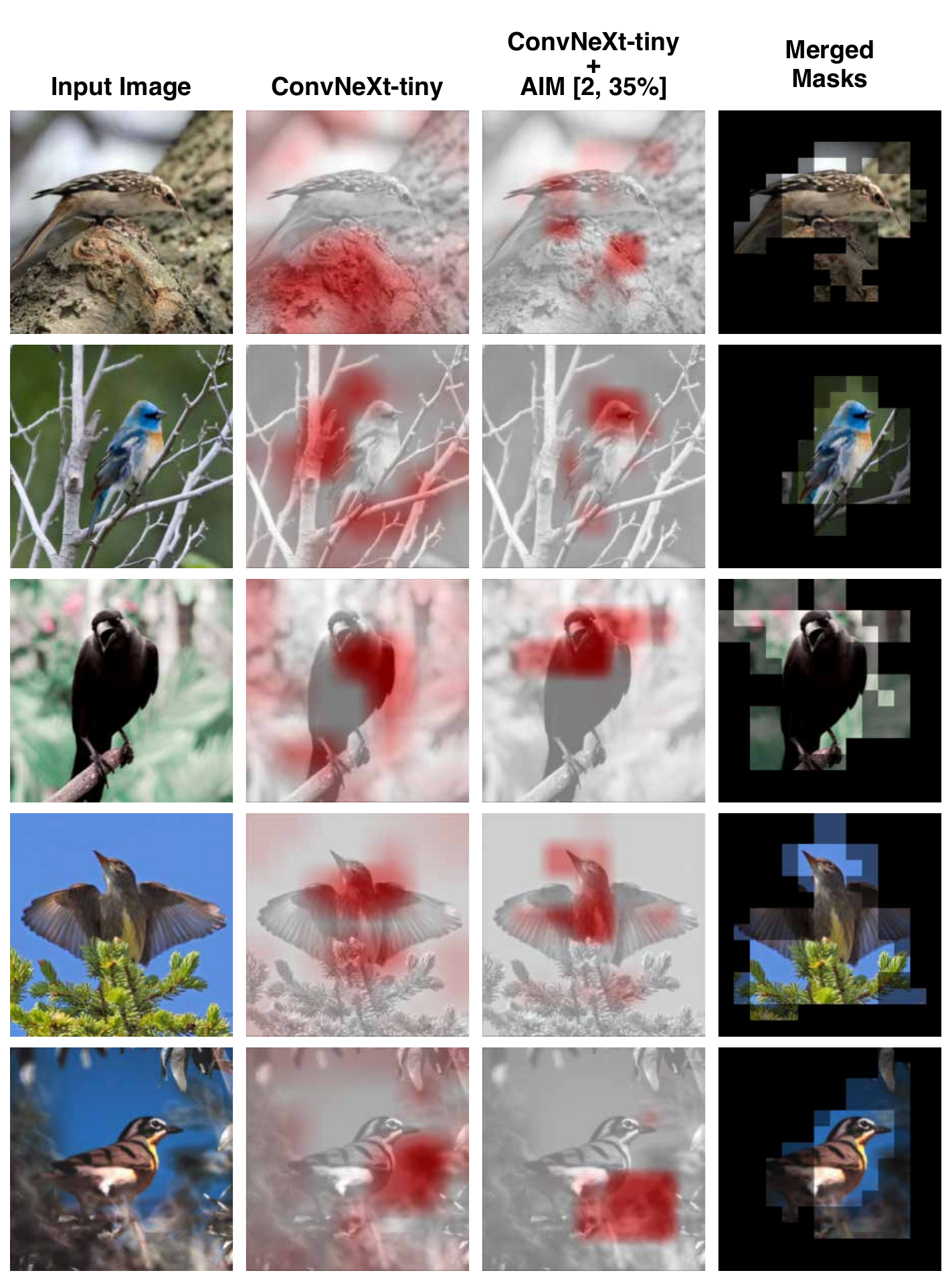} &
        \includegraphics[width=0.32\textwidth]{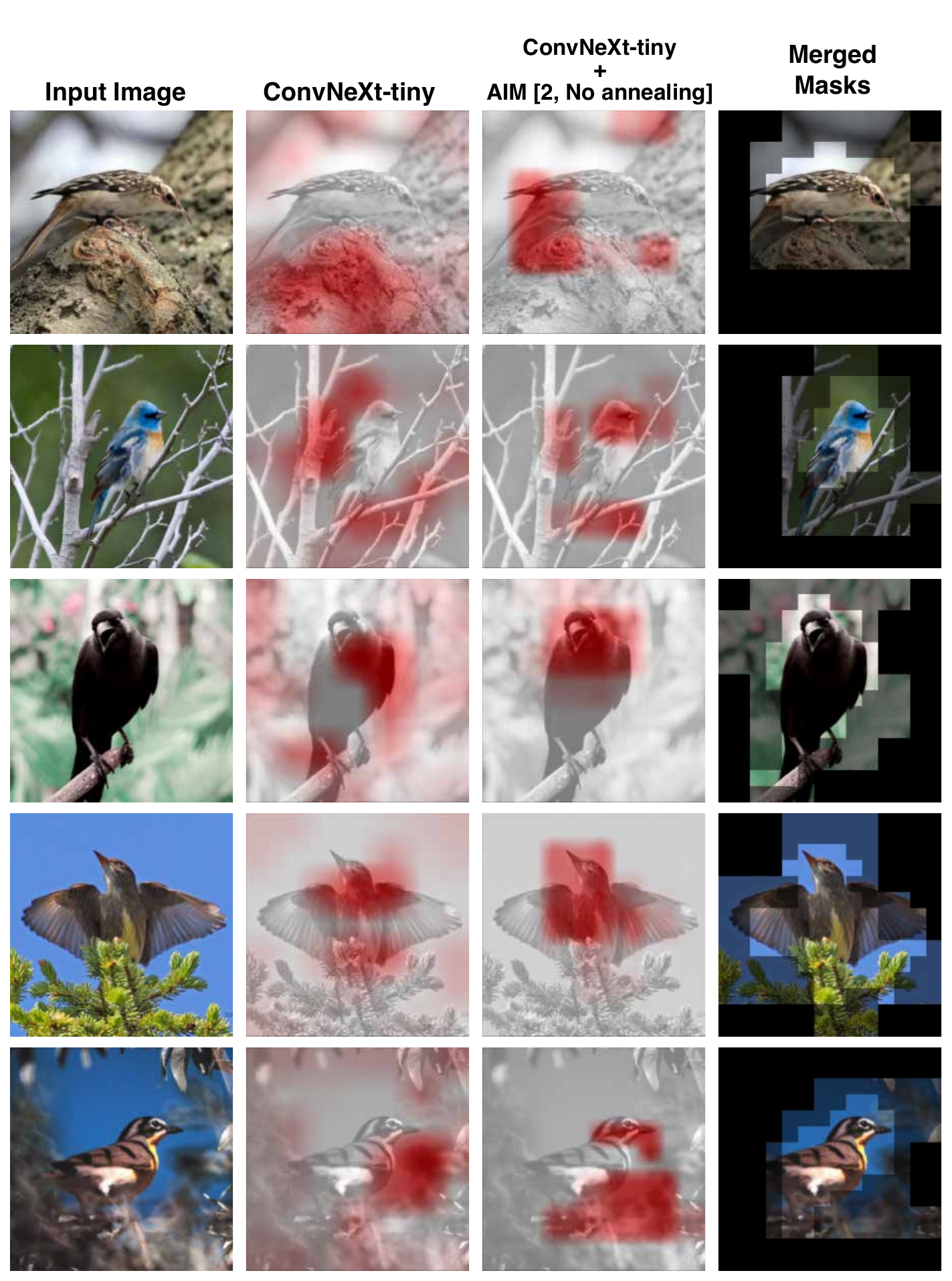}
    \end{tabular}
    \caption{\textbf{Qualitative results on the CUB-200 dataset.} Comparison of GradCAM attribution maps between ConvNeXt-tiny+AIM models with different mask active-area thresholds, along with the generated merged masks.}
    \label{fig:appendix-attribution_cub-ConvNeXt-b2}
\end{figure}

\begin{figure}[ht]
    \centering
    \begin{tabular}{ccc}
        \includegraphics[width=0.32\textwidth]{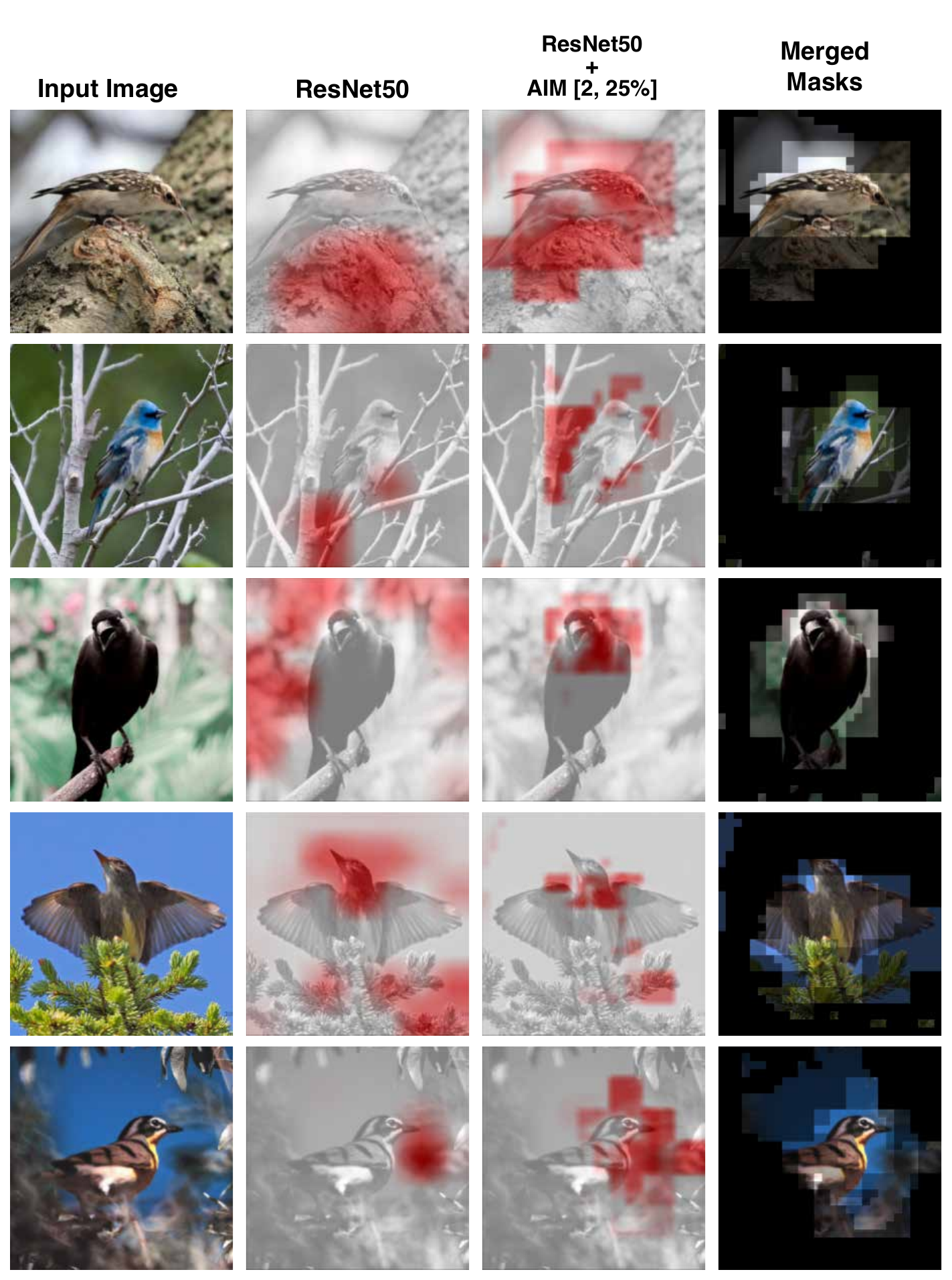} &
        \includegraphics[width=0.32\textwidth]{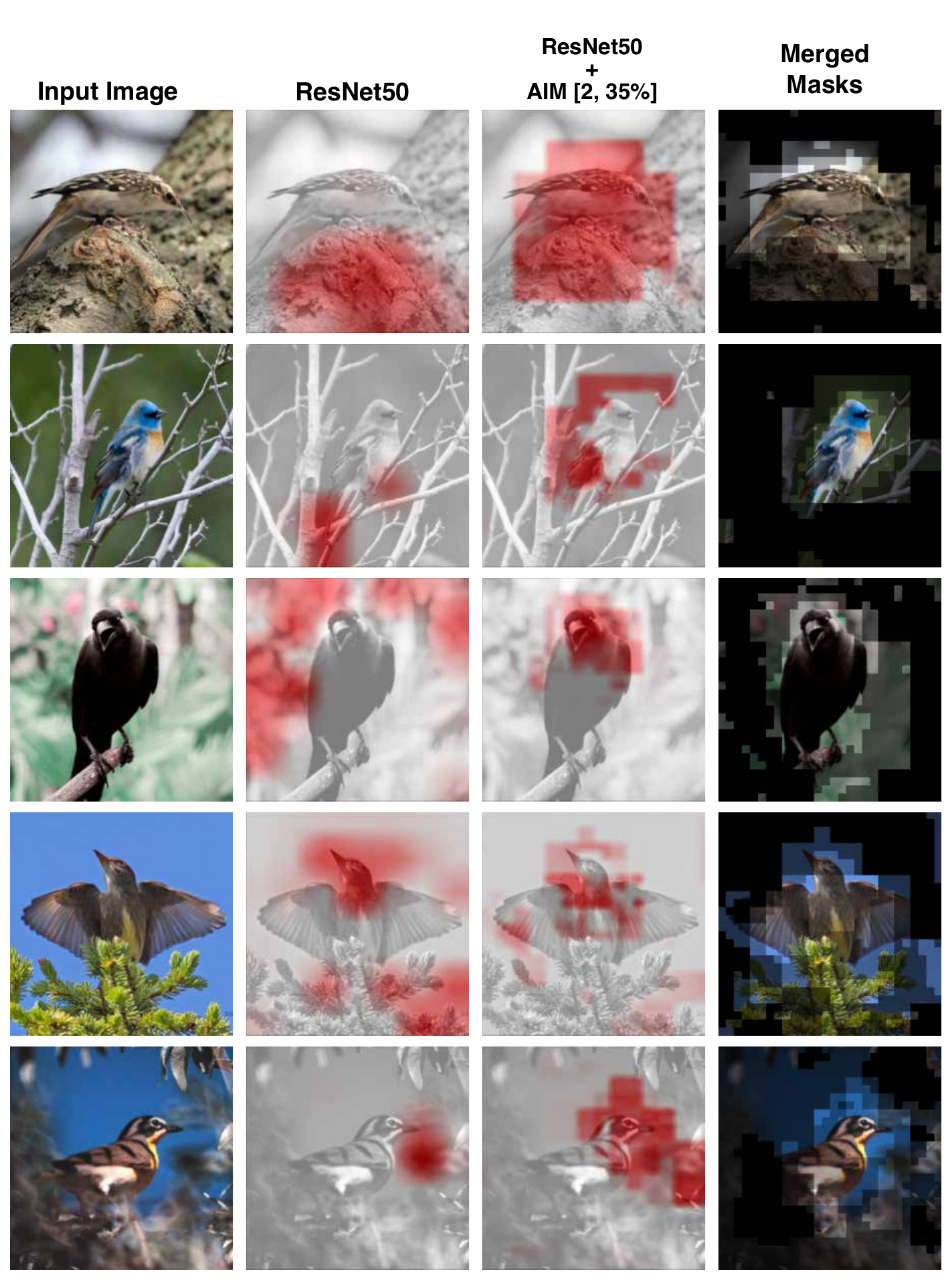} &
        \includegraphics[width=0.32\textwidth]{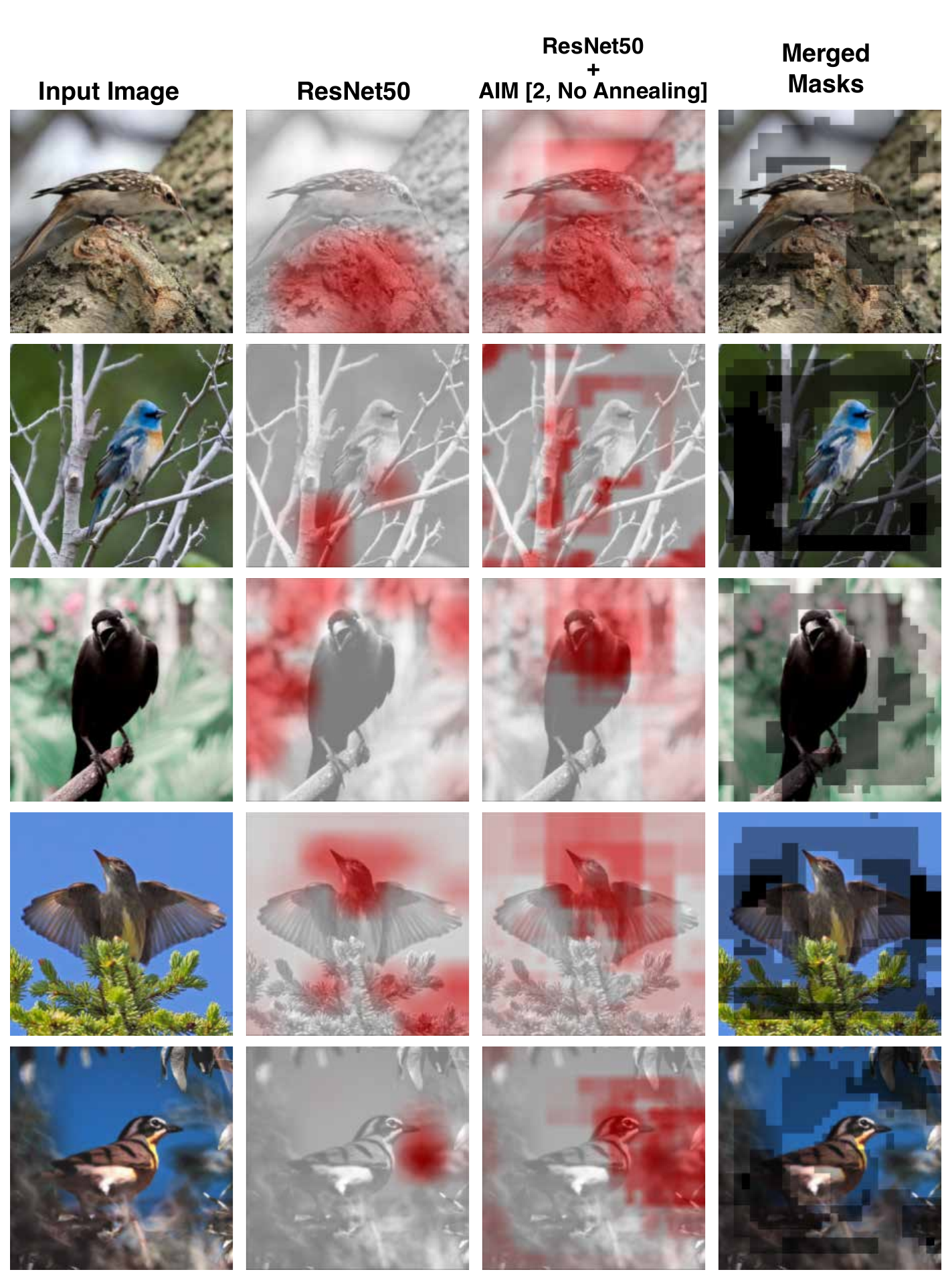}
    \end{tabular}
    \caption{\textbf{Qualitative results on the CUB-200 dataset.} Comparison of GradCAM attribution maps between ResNet50+AIM models with different mask active-area thresholds, along with the generated merged masks.}
    \label{fig:appendix-attribution_cub-ResNet50-b2}
\end{figure}

\begin{figure}[ht]
    \centering
    \begin{tabular}{ccc}
        \includegraphics[width=0.32\textwidth]{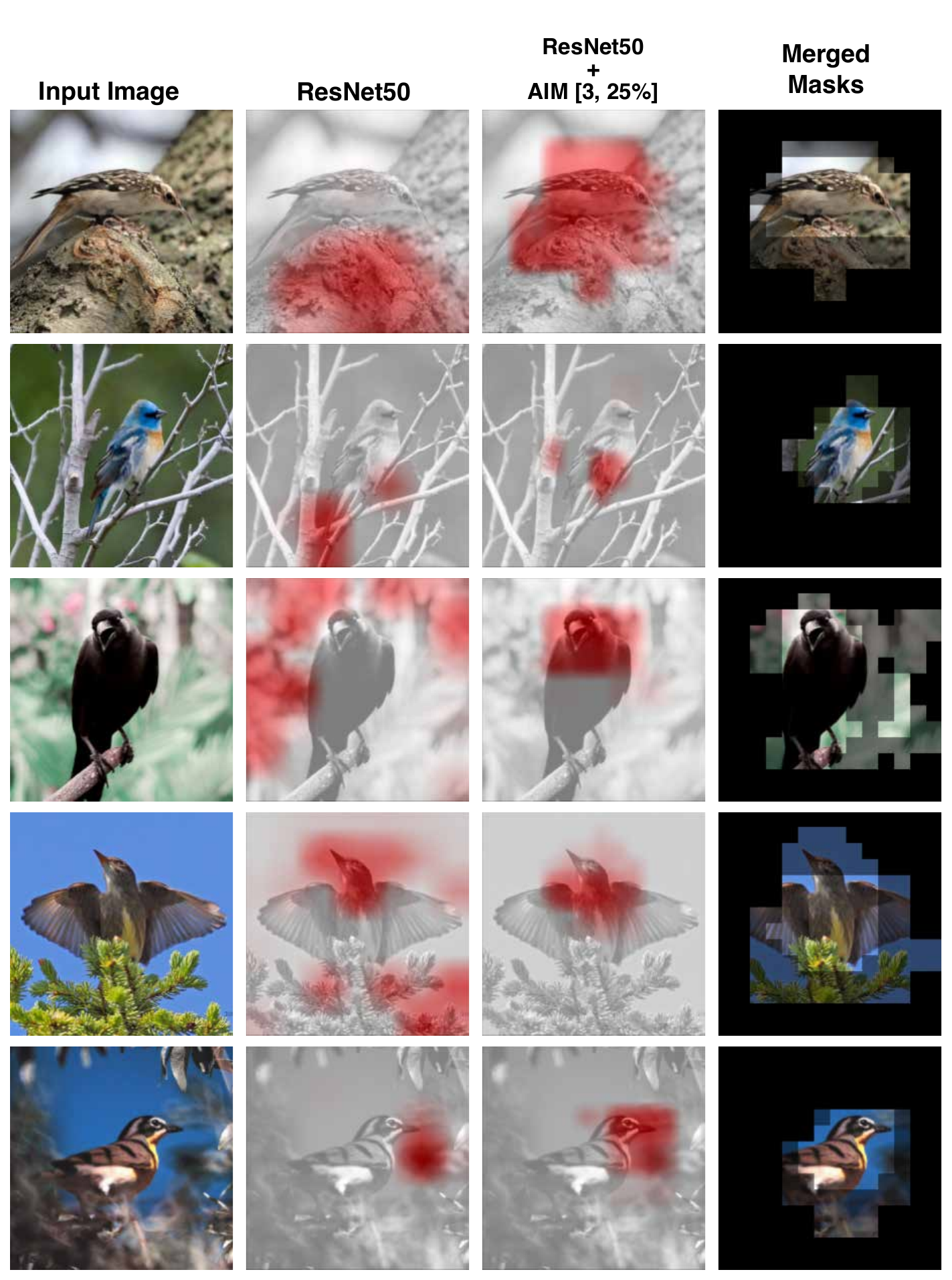} &
        \includegraphics[width=0.32\textwidth]{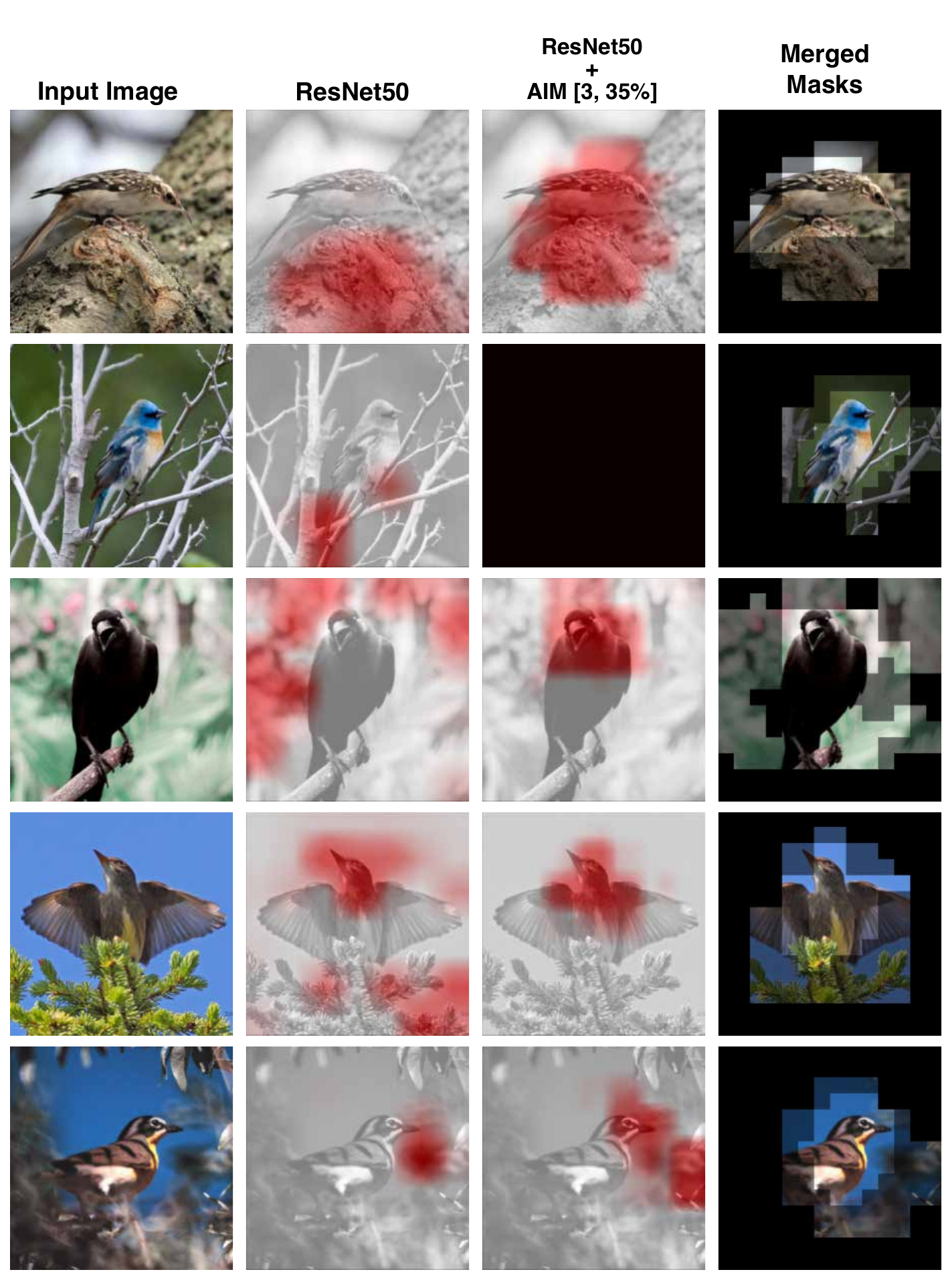} &
        \includegraphics[width=0.32\textwidth]{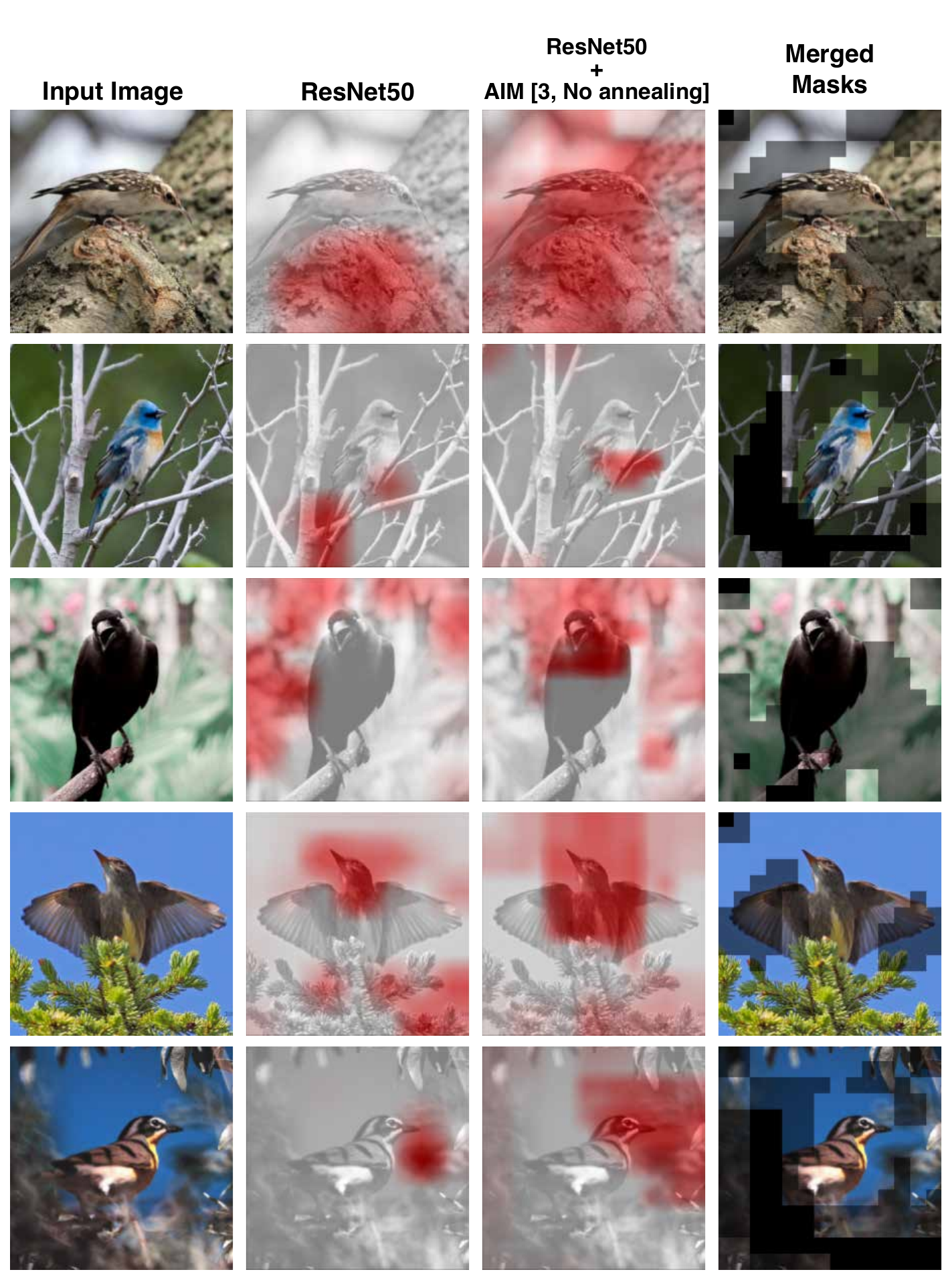}
    \end{tabular}
    \caption{\textbf{Qualitative results on the CUB-200 dataset.} Comparison of GradCAM attribution maps between ResNet50+AIM models with different mask active-area thresholds, along with the generated merged masks.}
    \label{fig:appendix-attribution_cub-ResNet50-b3}
\end{figure}

\begin{figure}[ht]
    \centering
    \begin{tabular}{ccc}
        \includegraphics[width=0.32\textwidth]{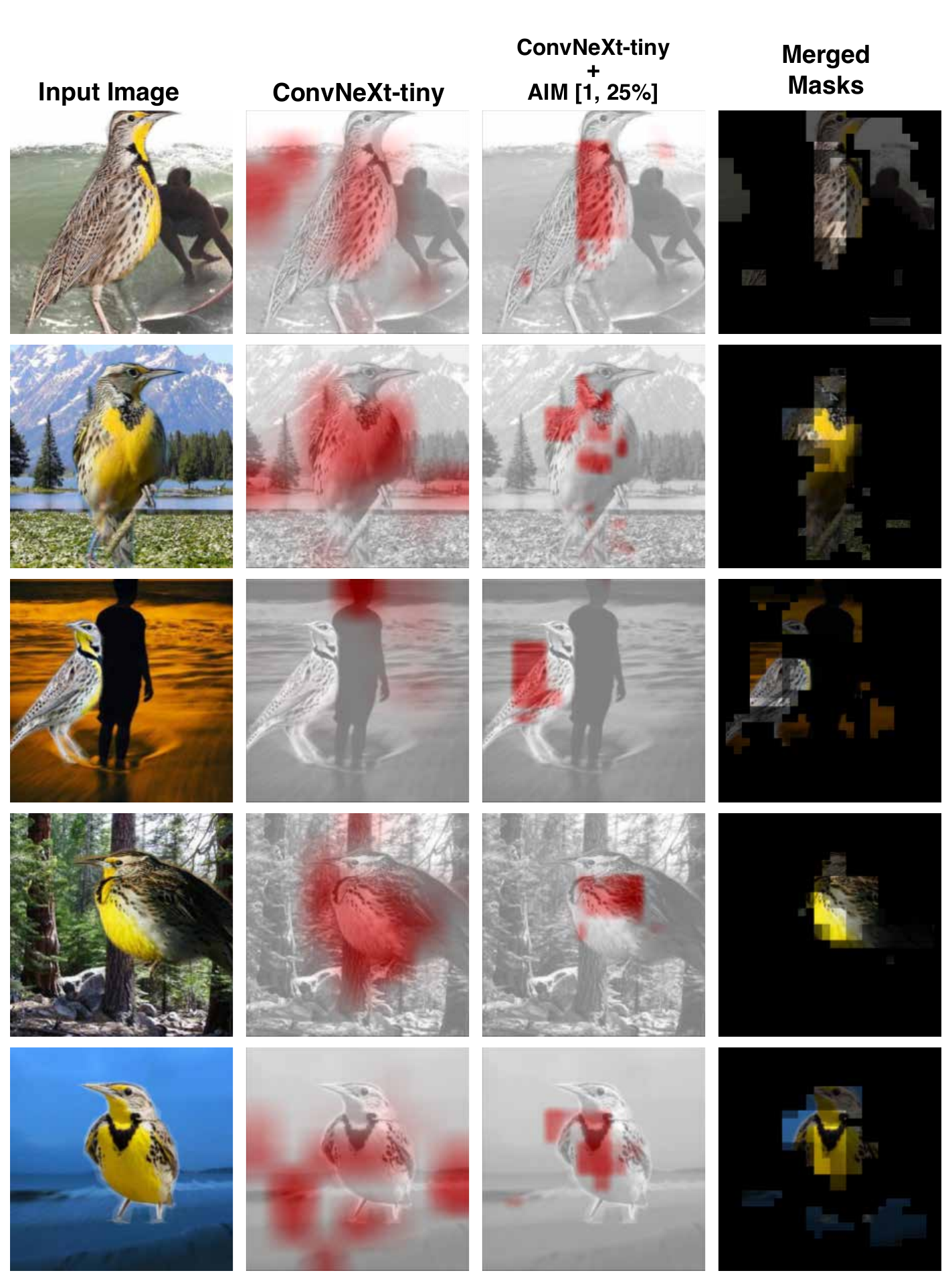} &
        \includegraphics[width=0.32\textwidth]{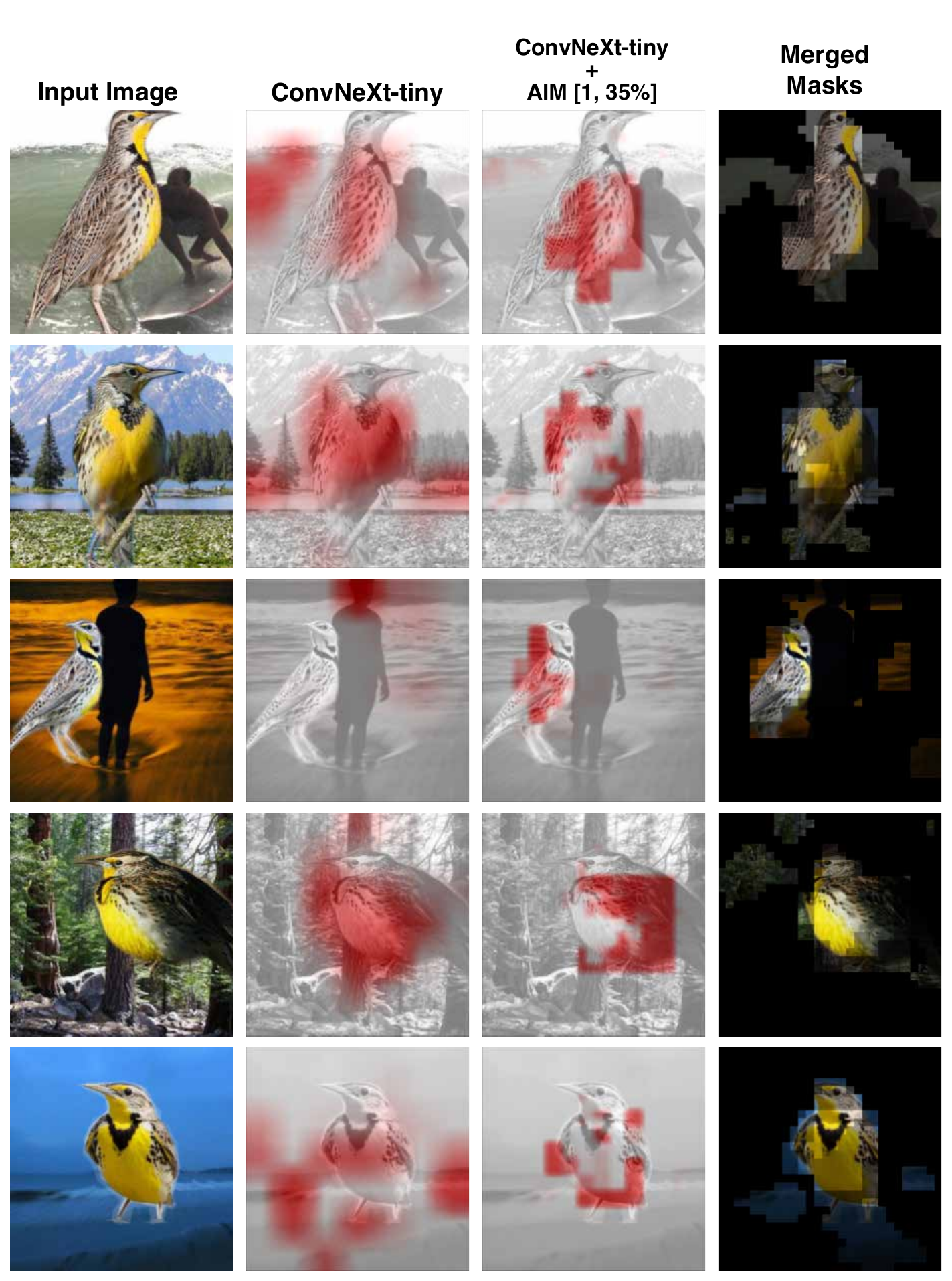} &
        \includegraphics[width=0.32\textwidth]{figures_pdfs/Appendix/attributions/cub/convnext_tiny/AIM-convnext_tiny1,_No_Annealing.pdf}
    \end{tabular}
    \caption{\textbf{Qualitative results on the Waterbirds-95\% dataset \cite{sagawa2019distributionally}.} Comparison of GradCAM attribution maps between ConvNeXt-tiny+AIM models with different mask active-area thresholds, along with the generated merged masks.}
    \label{fig:appendix-attribution_wb95-ConvNeXt-b1}
\end{figure}

\begin{figure}[ht]
    \centering
    \begin{tabular}{ccc}
        \includegraphics[width=0.32\textwidth]{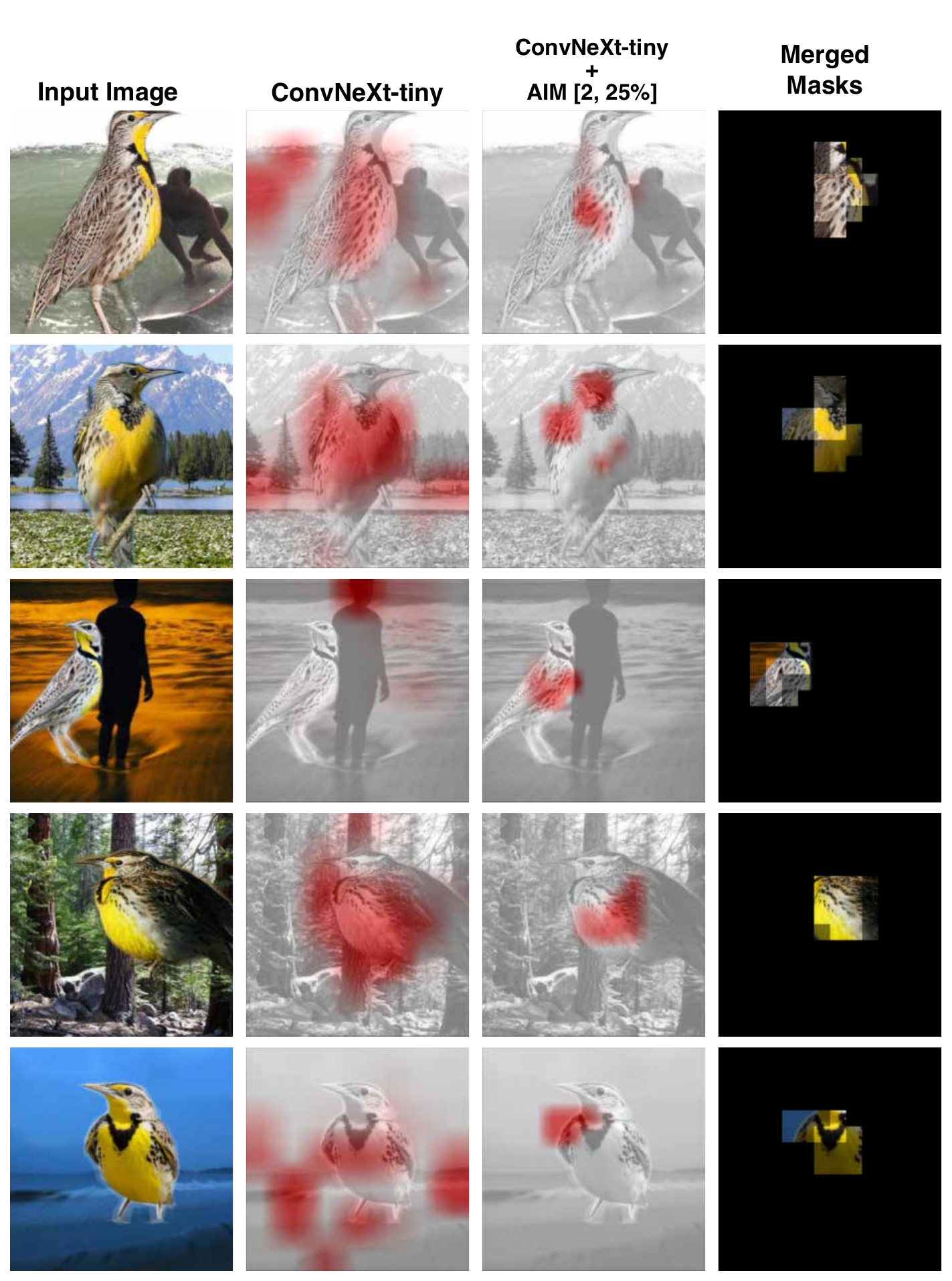} &
        \includegraphics[width=0.32\textwidth]{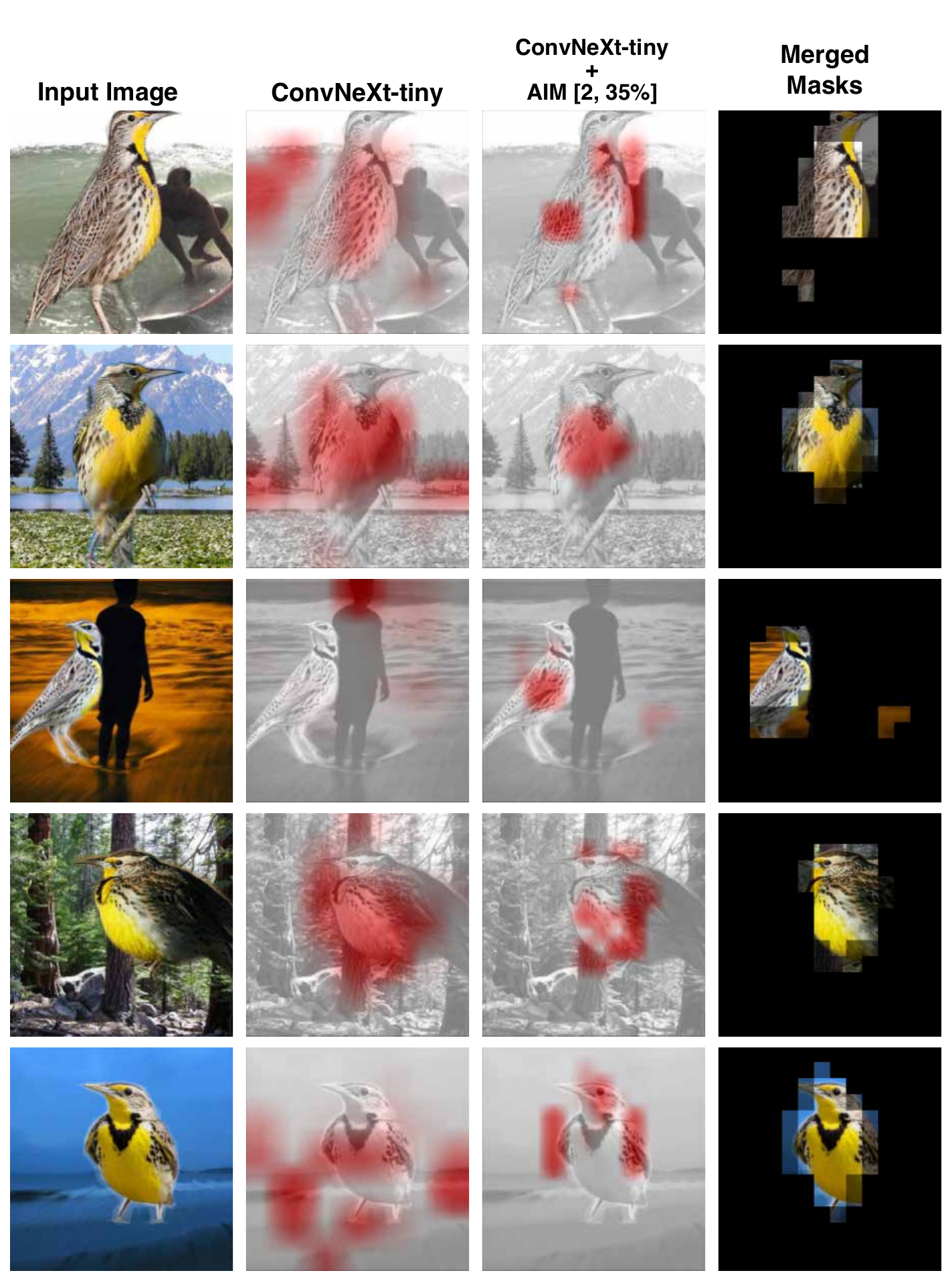} &
        \includegraphics[width=0.32\textwidth]{figures_pdfs/Appendix/attributions/cub/convnext_tiny/AIM-convnext_tiny2,_No_Annealing.pdf}
    \end{tabular}
    \caption{\textbf{Qualitative results on the Waterbirds-95\% dataset \cite{sagawa2019distributionally}.} Comparison of GradCAM attribution maps between ConvNeXt-tiny+AIM models with different mask active-area thresholds, along with the generated merged masks.}
    \label{fig:appendix-attribution_wb95-ConvNeXt-b2}
\end{figure}

\begin{figure}[ht]
    \centering
    \begin{tabular}{ccc}
        \includegraphics[width=0.32\textwidth]{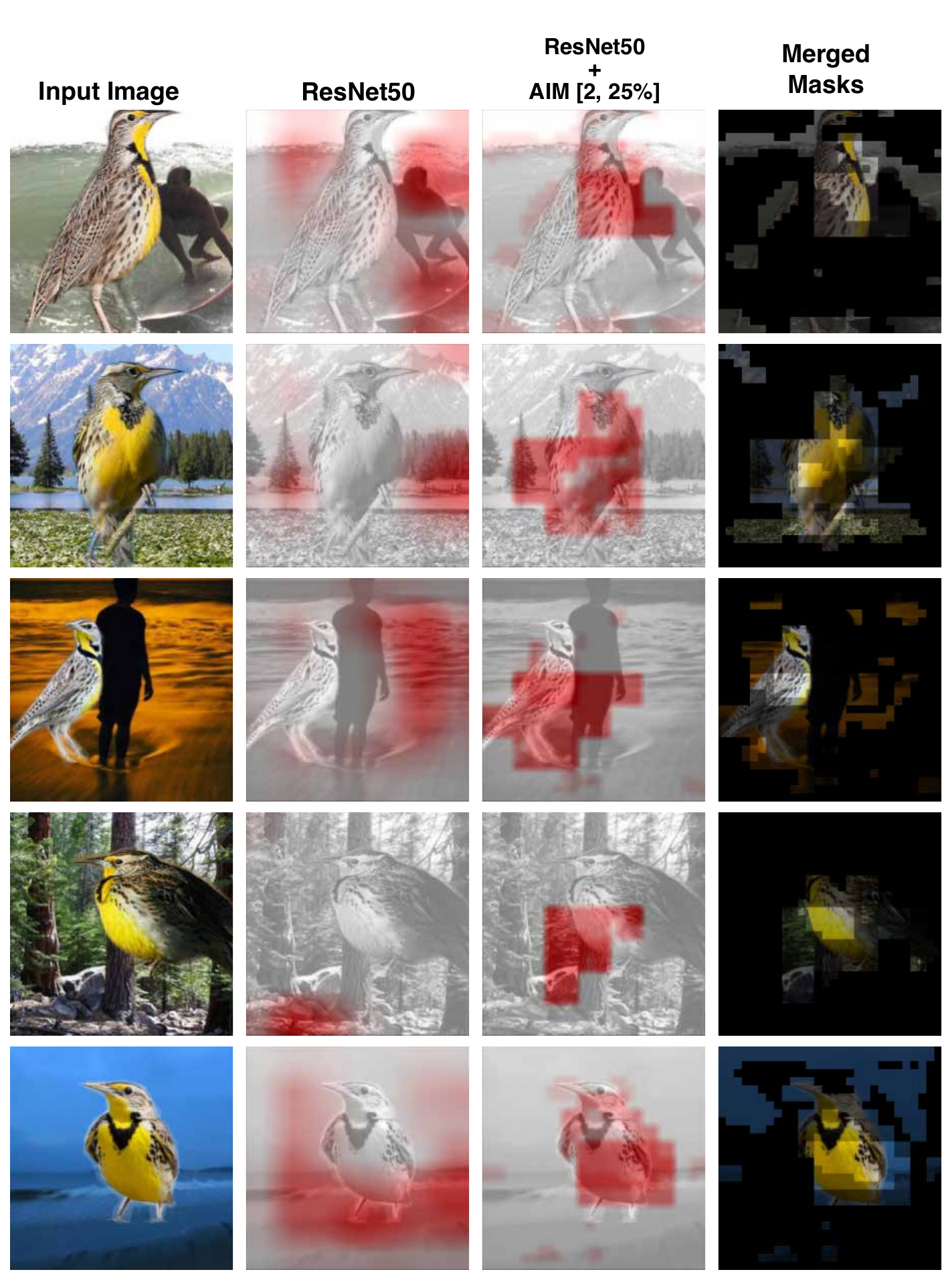} &
        \includegraphics[width=0.32\textwidth]{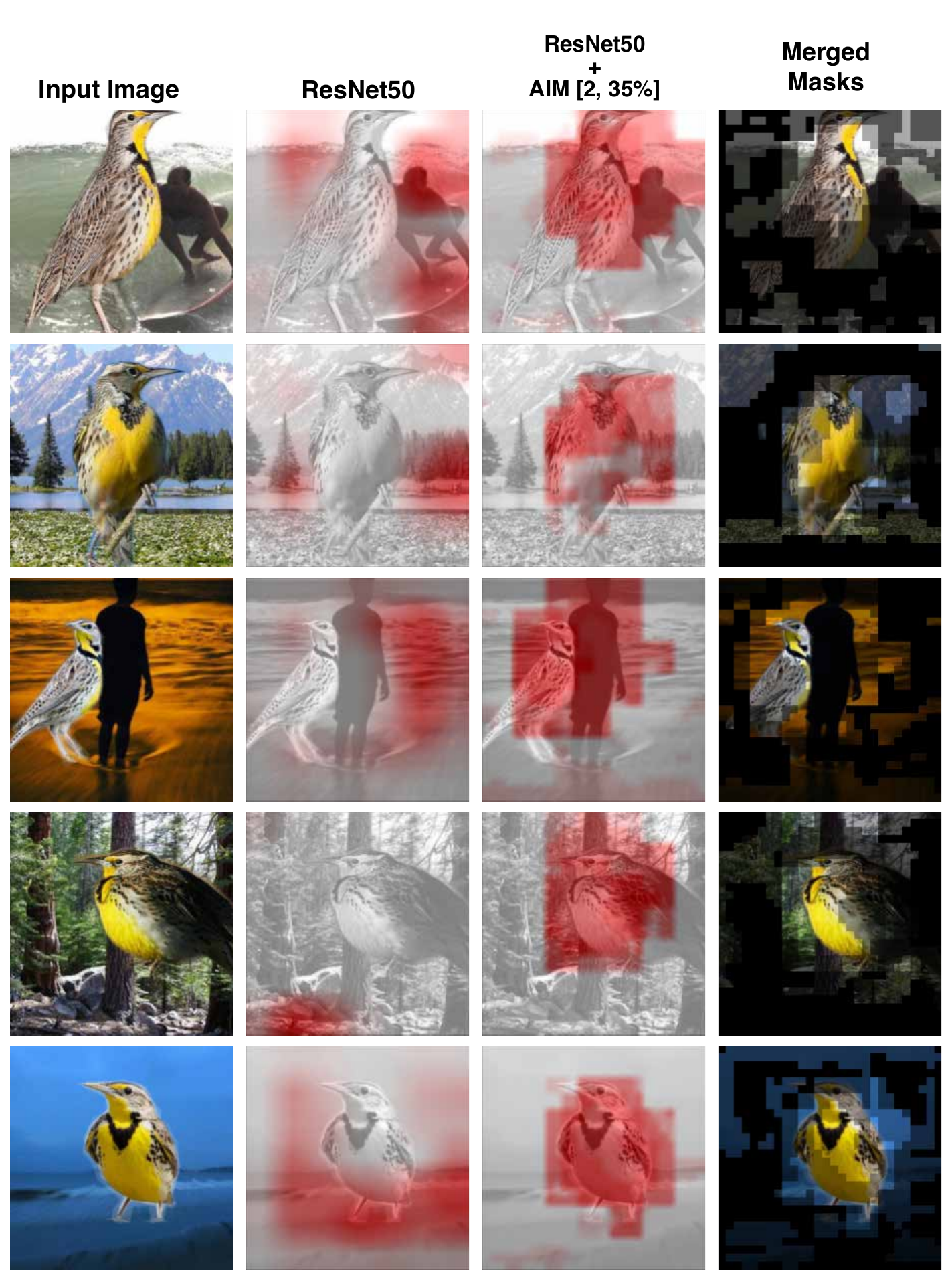} &
        \includegraphics[width=0.32\textwidth]{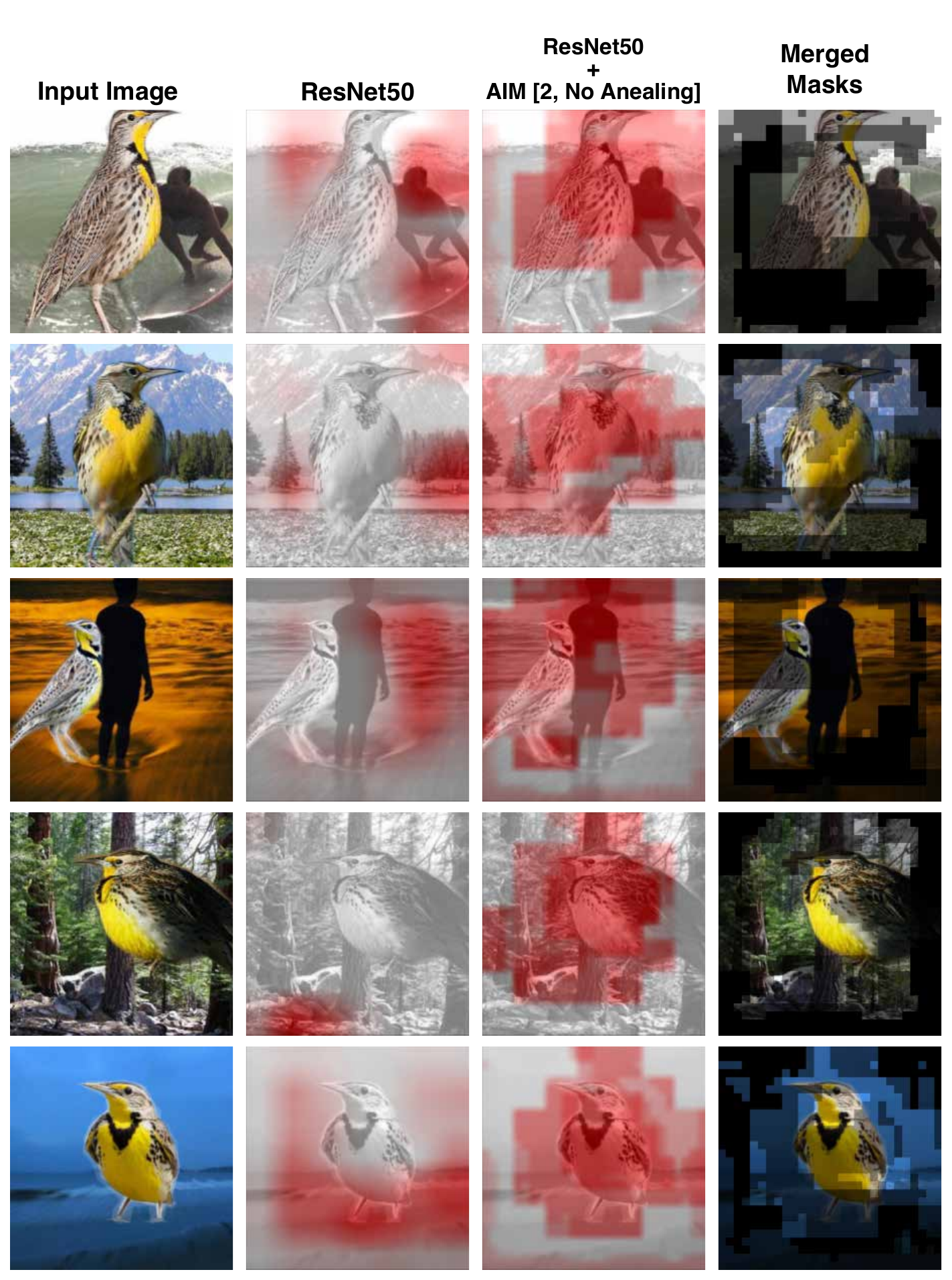}
    \end{tabular}
    \caption{\textbf{Qualitative results on the Waterbirds-95\% dataset \cite{sagawa2019distributionally}.} Comparison of GradCAM attribution maps between ResNet50+AIM models with different mask active-area thresholds, along with the generated merged masks.}
    \label{fig:appendix-attribution_wb95-ResNet50-b2}
\end{figure}

\begin{figure}[ht]
    \centering
    \begin{tabular}{ccc}
        \includegraphics[width=0.32\textwidth]{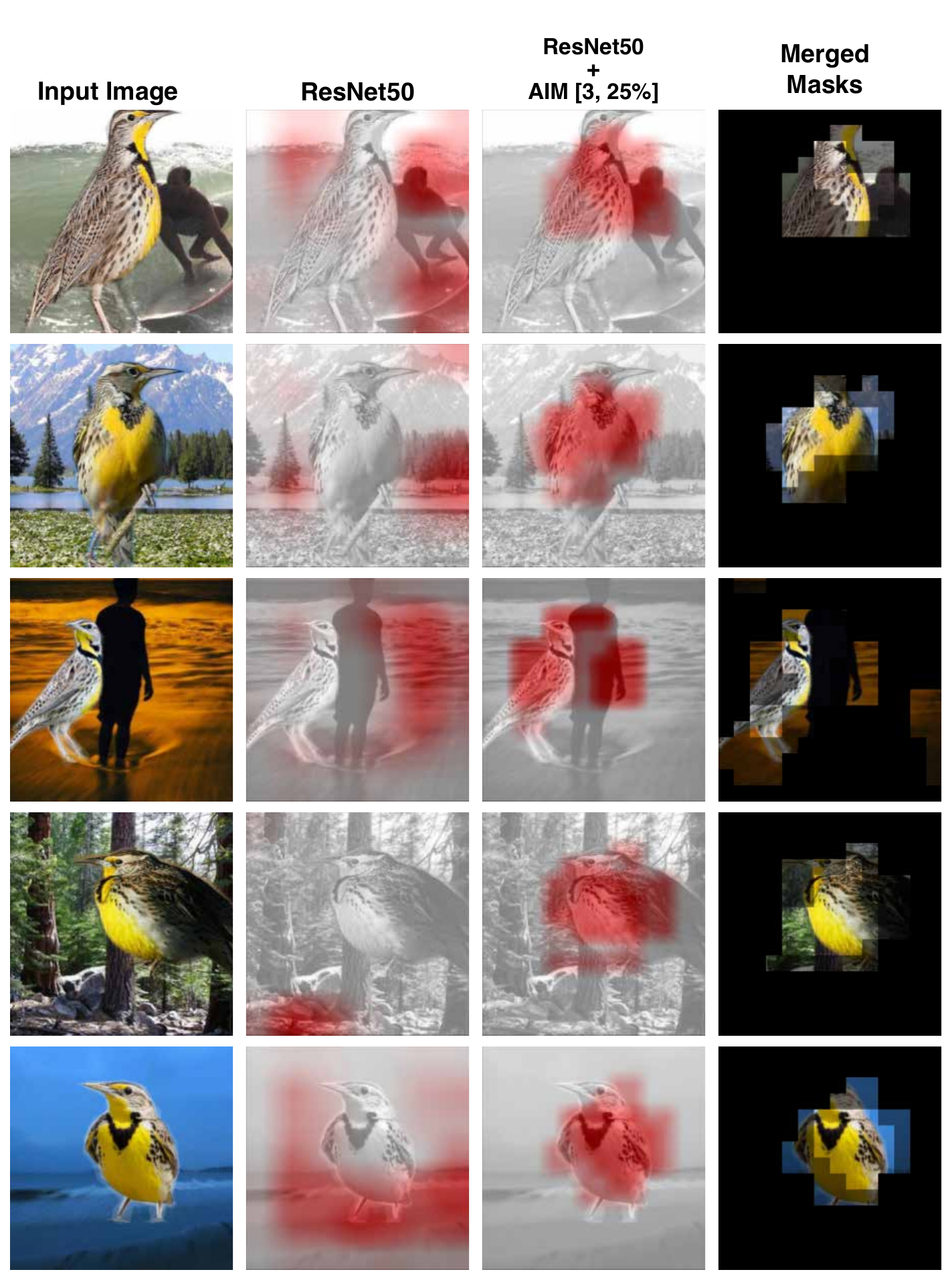} &
        \includegraphics[width=0.32\textwidth]{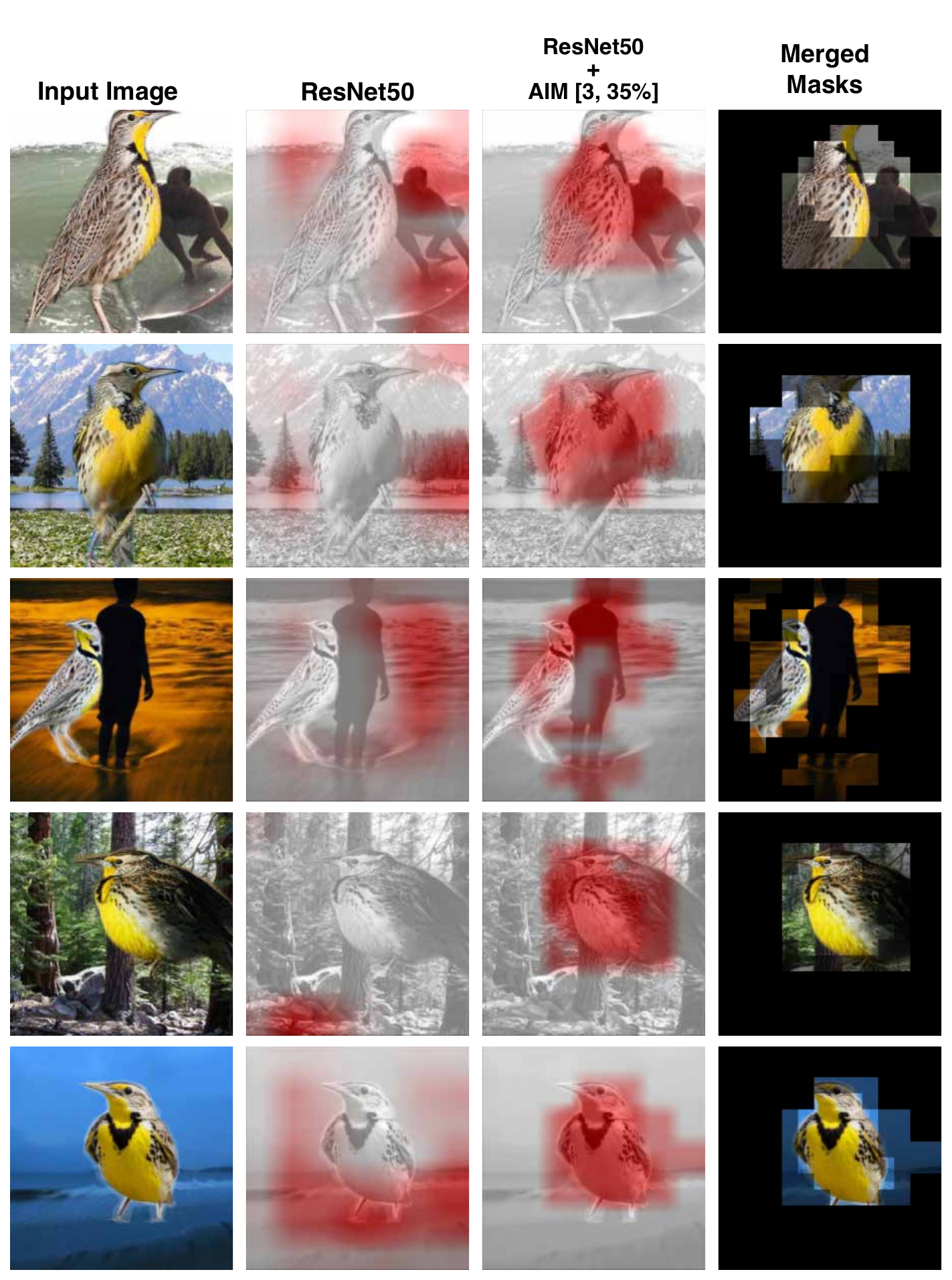} &
        \includegraphics[width=0.32\textwidth]{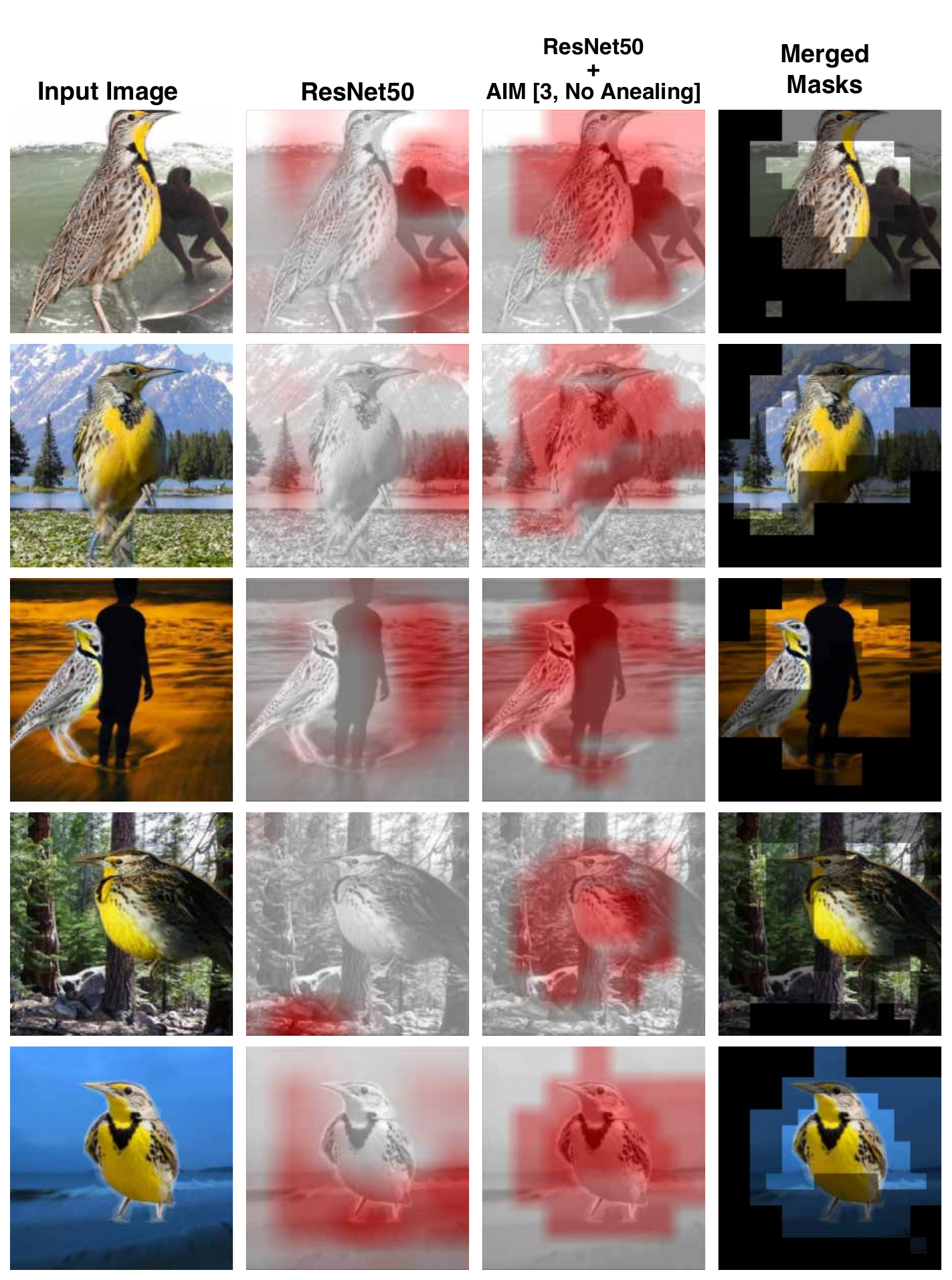}
    \end{tabular}
    \caption{\textbf{Qualitative results on the Waterbirds-95\% dataset \cite{sagawa2019distributionally}.} Comparison of GradCAM attribution maps between ResNet50+AIM models with different mask active-area thresholds, along with the generated merged masks.}
    \label{fig:appendix-attribution_wb95-ResNet50-b3}
\end{figure}

\begin{figure}[ht]
    \centering
    \begin{tabular}{ccc}
        \includegraphics[width=0.32\textwidth]{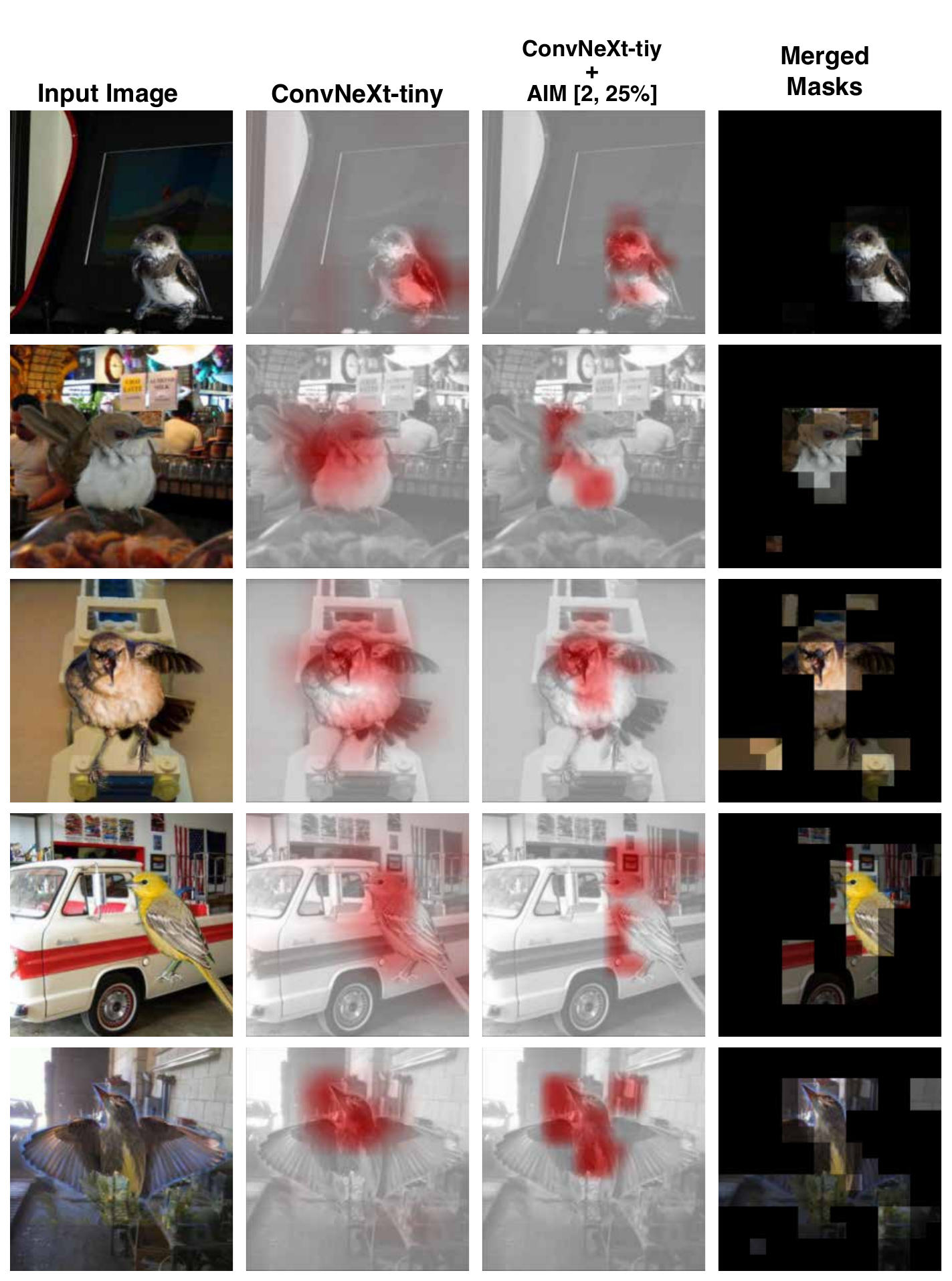} &
        \includegraphics[width=0.32\textwidth]{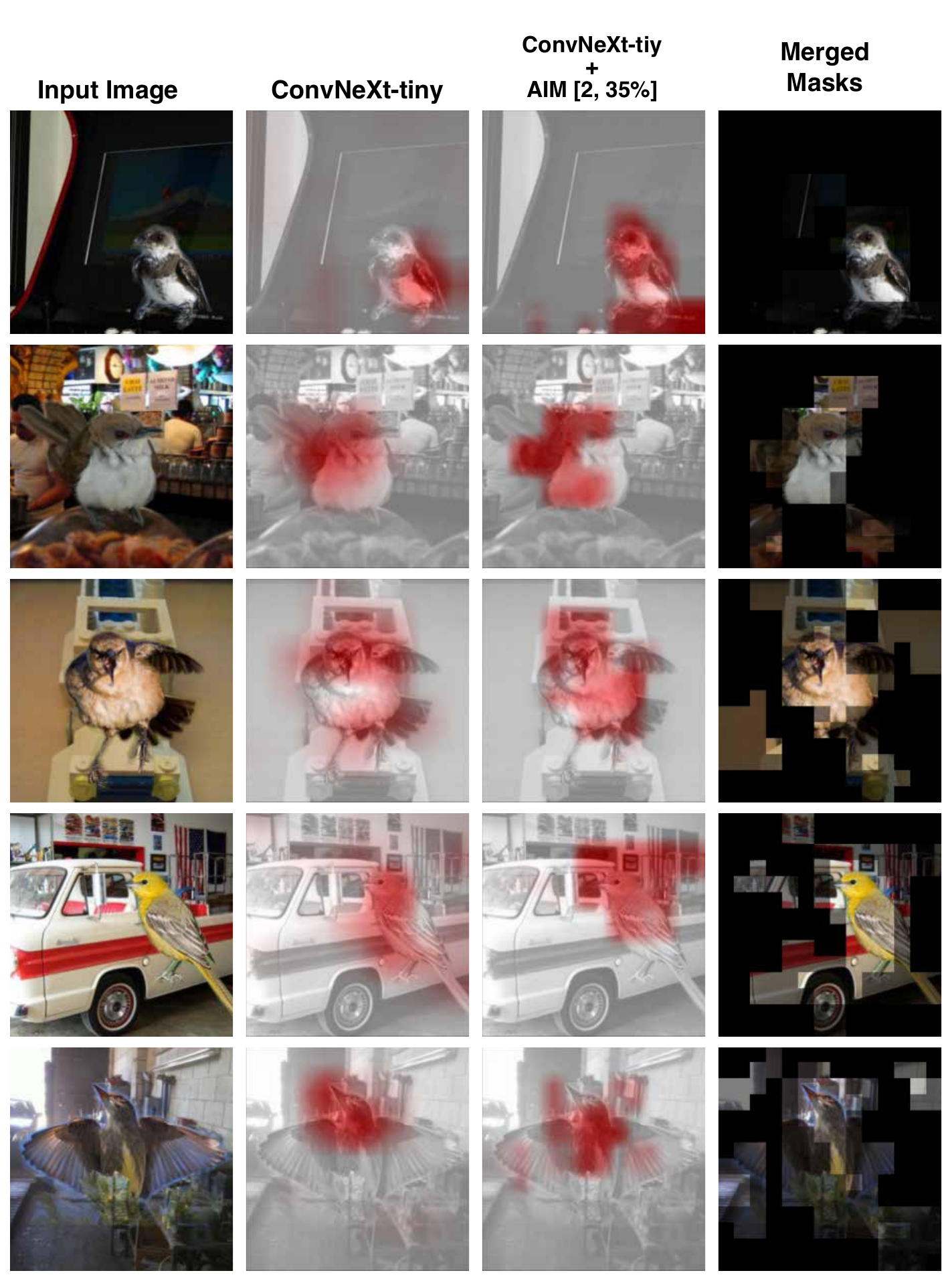} &
        \includegraphics[width=0.32\textwidth]{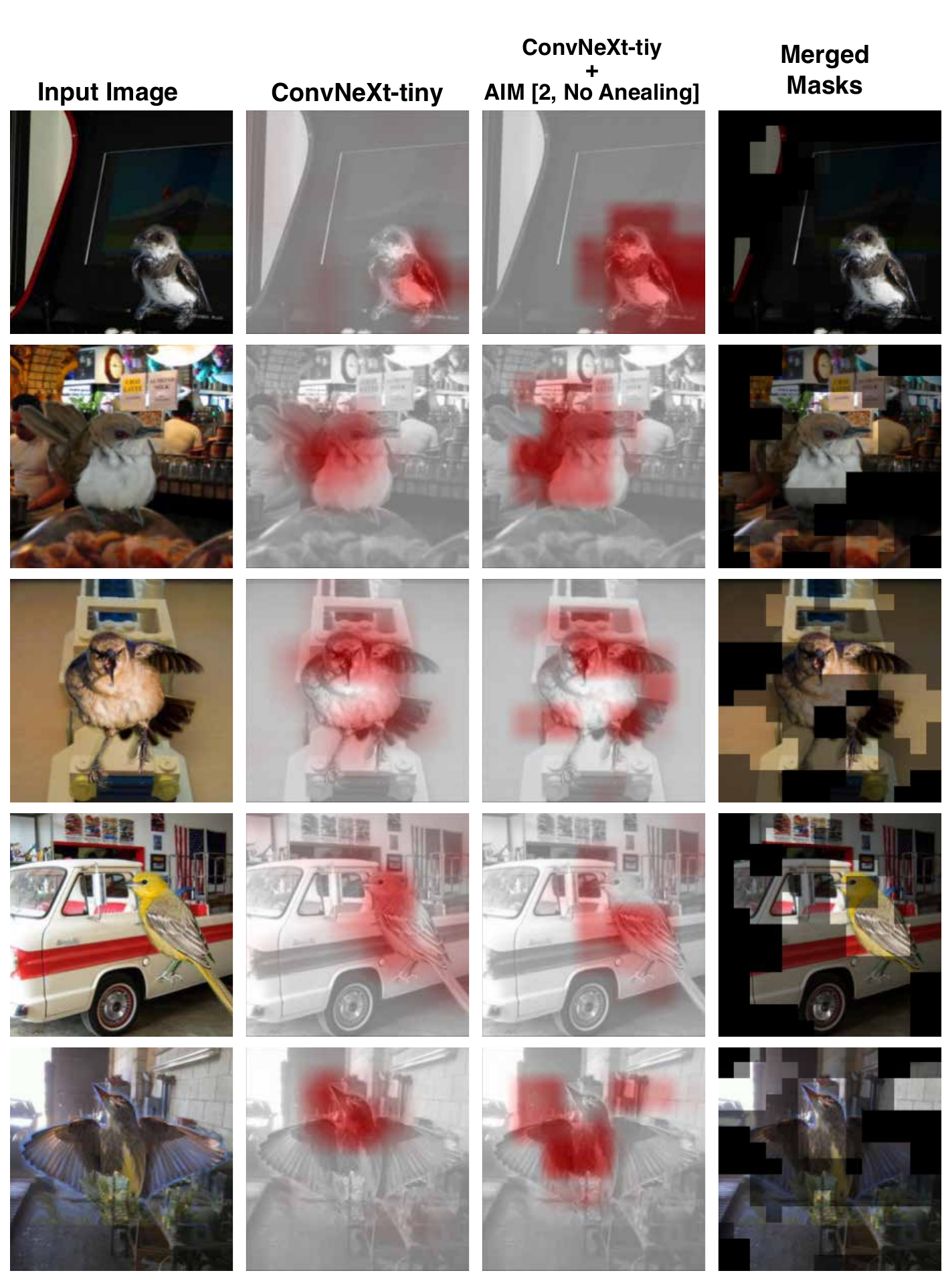}
    \end{tabular}
    \caption{\textbf{Qualitative results on the Travelingbirds dataset \cite{koh2020concept}.} Comparison of GradCAM attribution maps between ConvNeXt-tiny+AIM models with different mask active-area thresholds, along with the generated merged masks.}
    \label{fig:appendix-attribution_Travelingbirds-ResNet101}
\end{figure}

\begin{figure}[ht]
    \centering
    \begin{tabular}{ccc}
        \includegraphics[width=0.32\textwidth]{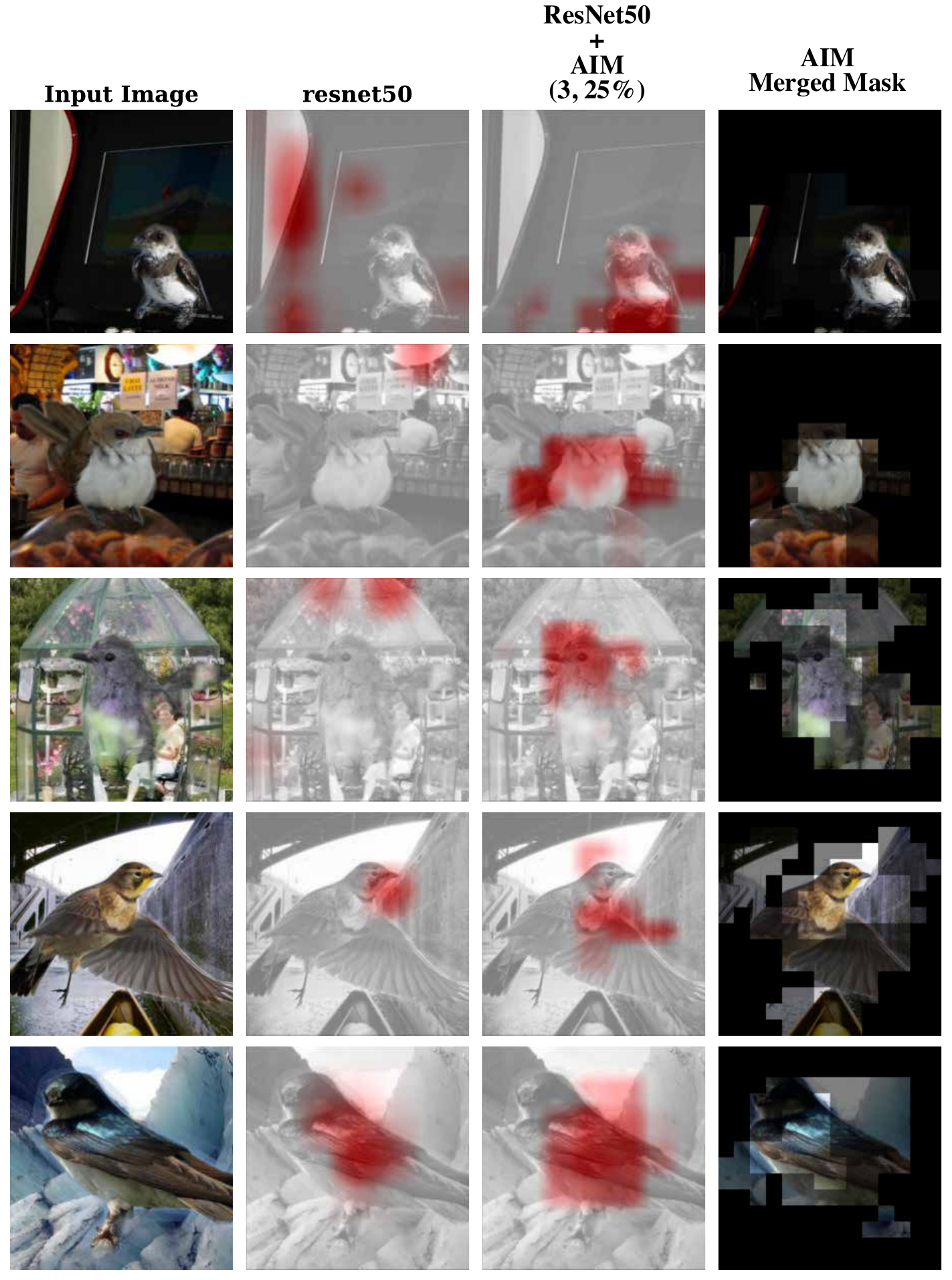} &
        \includegraphics[width=0.32\textwidth]{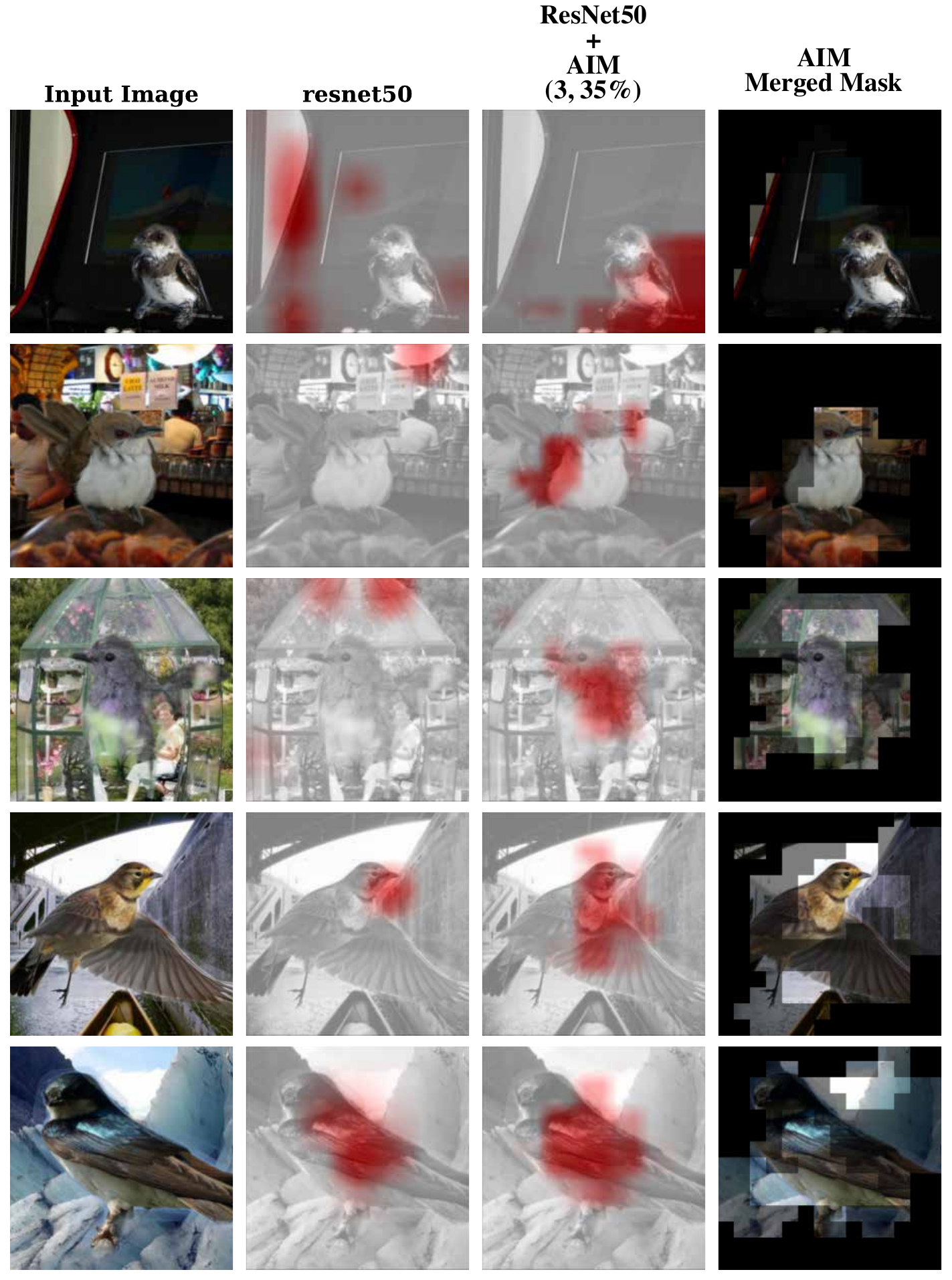} &
        \includegraphics[width=0.32\textwidth]{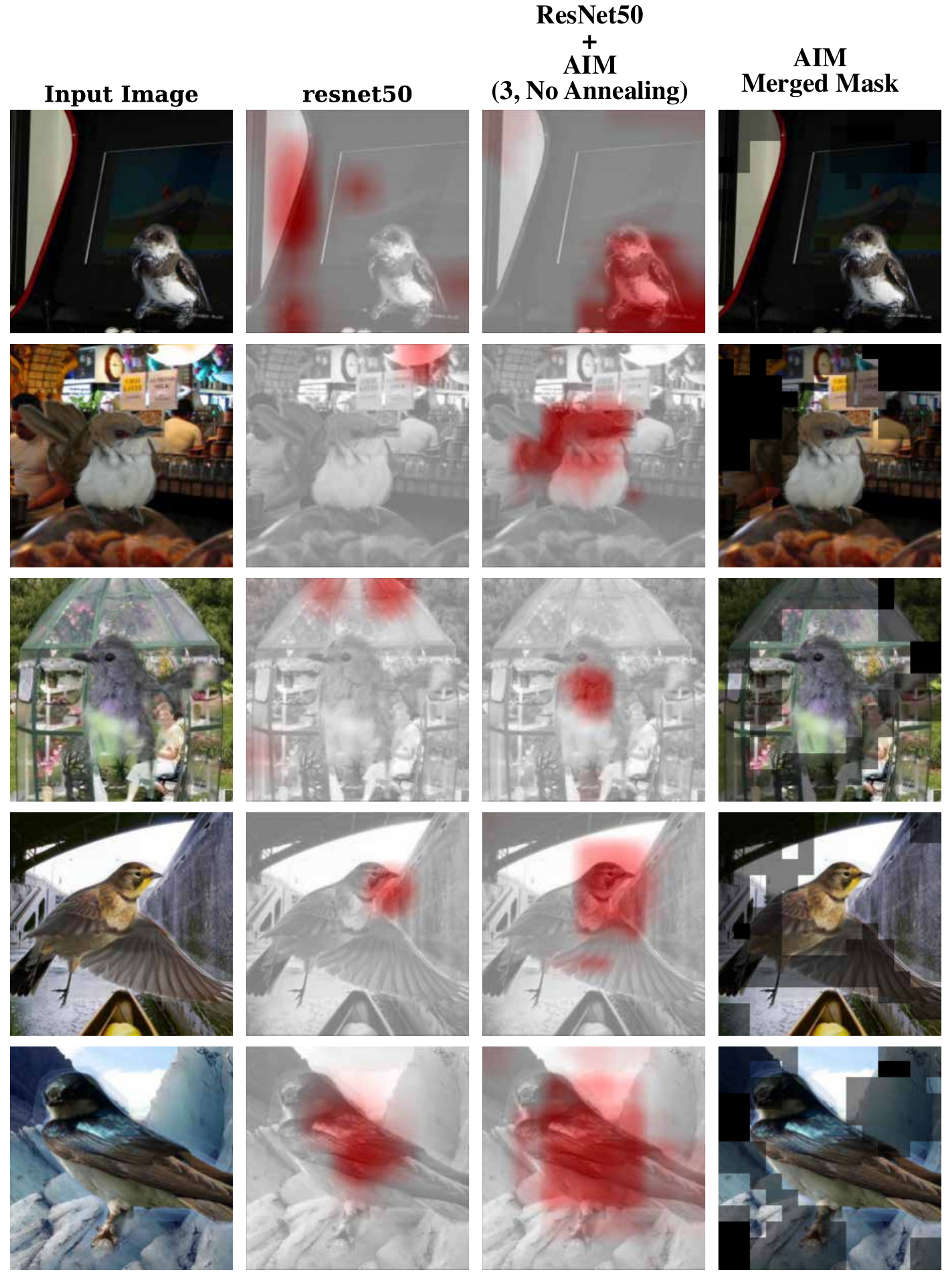}
    \end{tabular}
    \caption{\textbf{Qualitative results on the Travelingbirds dataset \cite{koh2020concept}.} Comparison of GradCAM attribution maps between ResNet50+AIM models with different mask active-area thresholds, along with the generated merged masks.}
    \label{fig:appendix-attribution_Travelingbirds-ResNet50}
\end{figure}

\begin{figure}[ht]
    \centering
    \begin{tabular}{ccc}
        \includegraphics[width=0.32\textwidth]{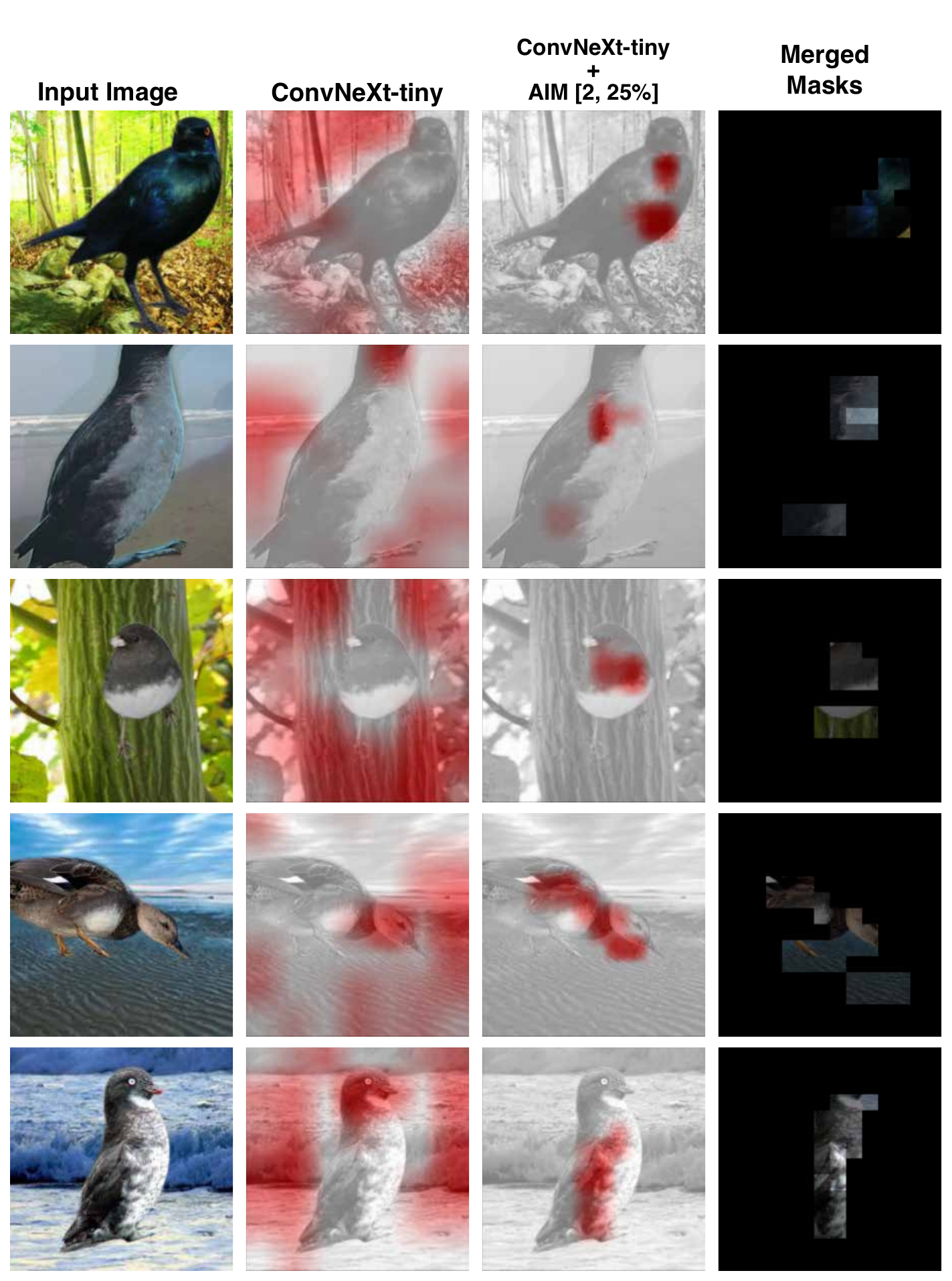} &
        \includegraphics[width=0.32\textwidth]{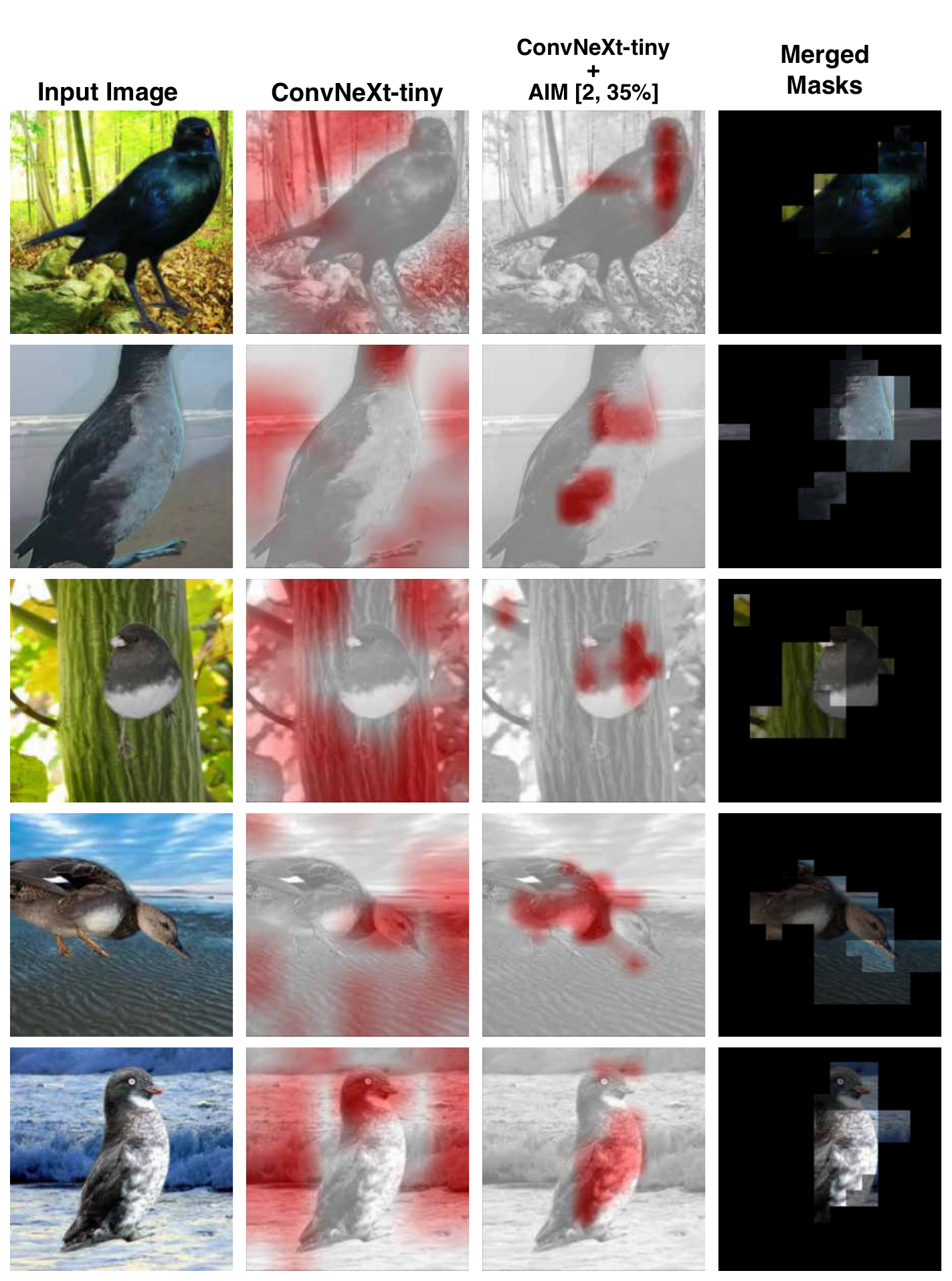} &
        \includegraphics[width=0.32\textwidth]{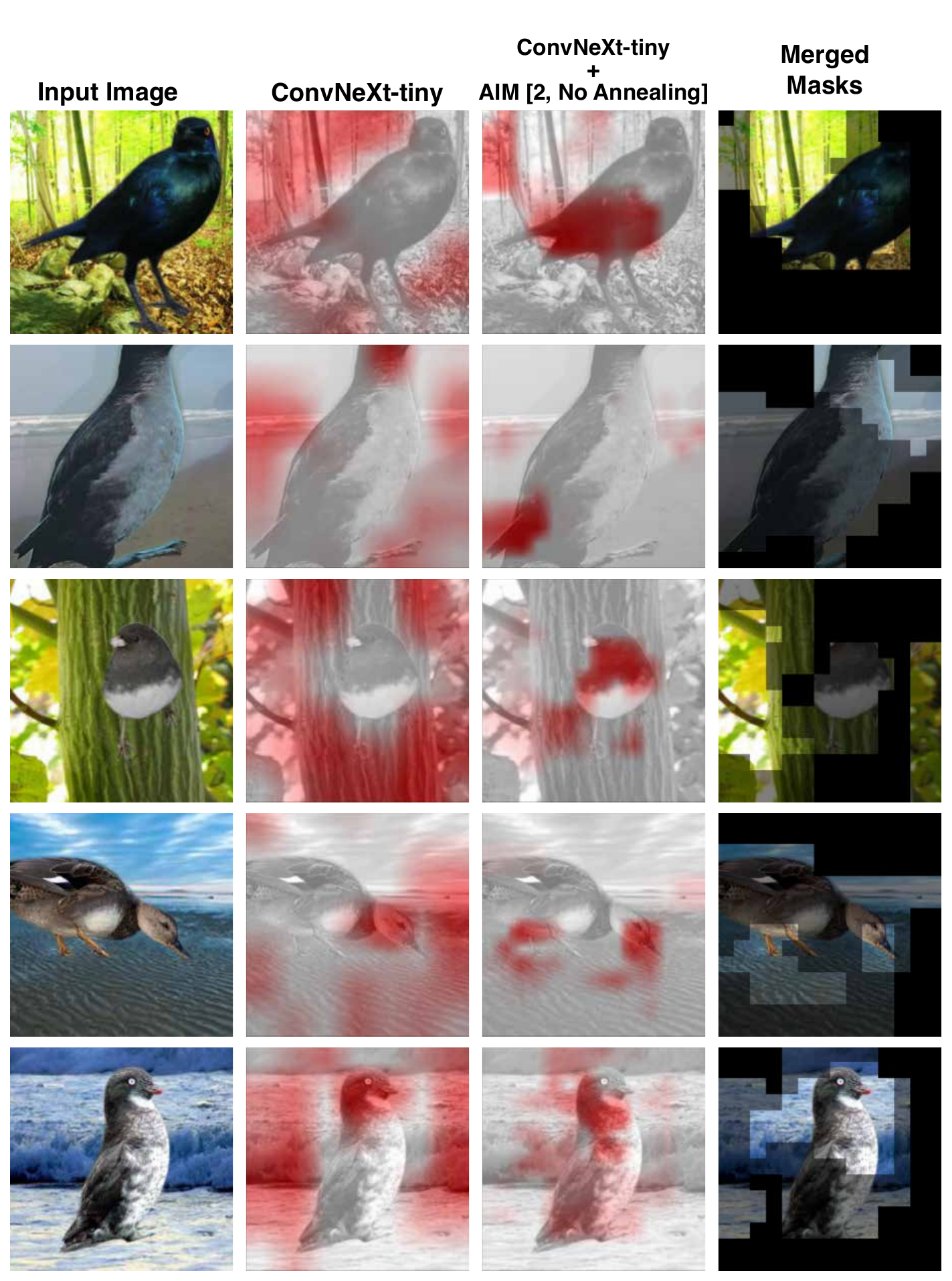}
    \end{tabular}
    \caption{\textbf{Qualitative results on the Waterbirds-100\% dataset \cite{petryk2022guiding}.} Comparison of GradCAM attribution maps between ConvNeXt-tiny+AIM models with different mask active-area thresholds, along with the generated merged masks.}
    \label{fig:appendix-attribution_wb100-convnext}
\end{figure}

\begin{figure}[ht]
    \centering
    \begin{tabular}{ccc}
        \includegraphics[width=0.32\textwidth]{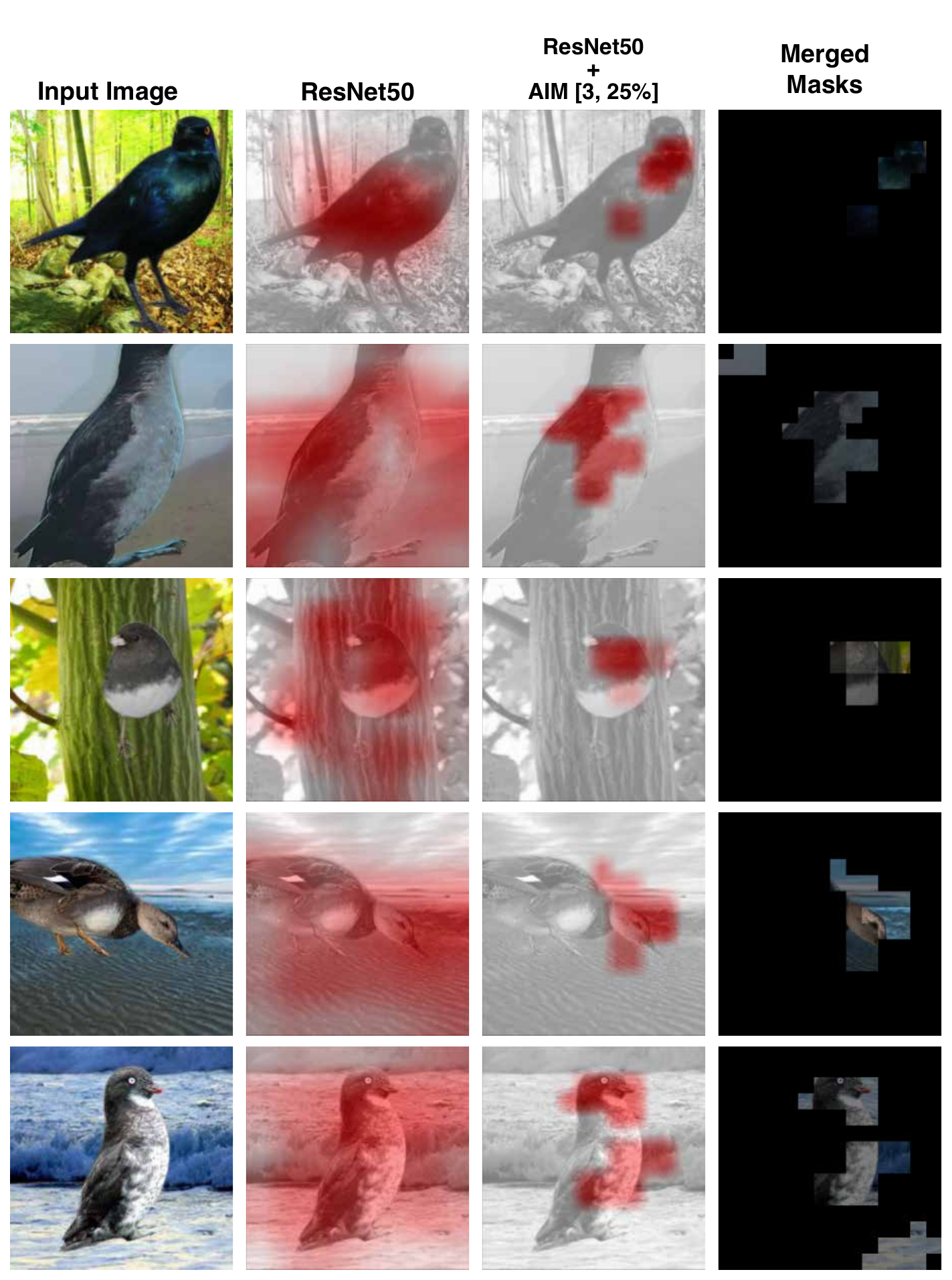} &
        \includegraphics[width=0.32\textwidth]{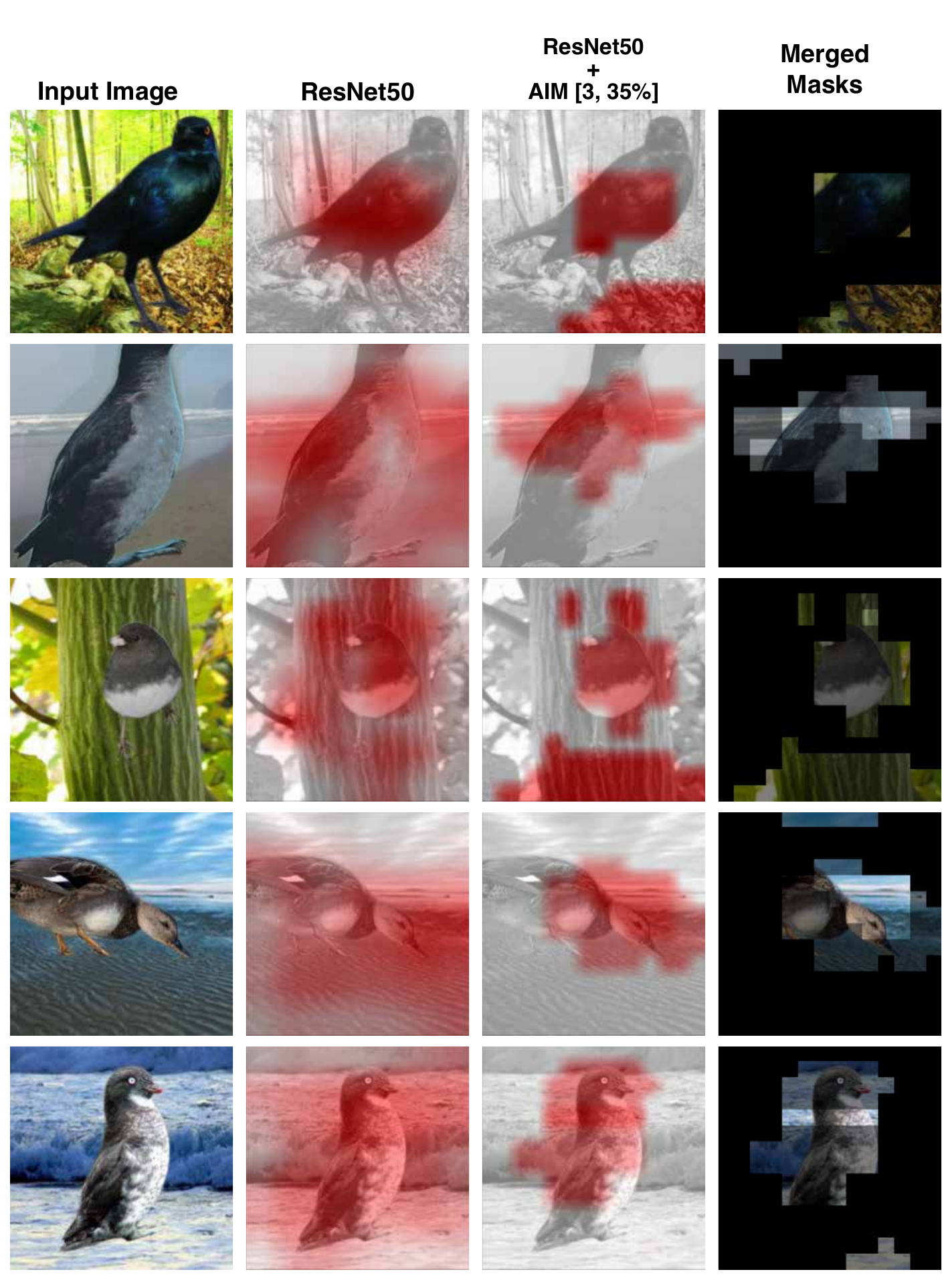} &
        \includegraphics[width=0.32\textwidth]{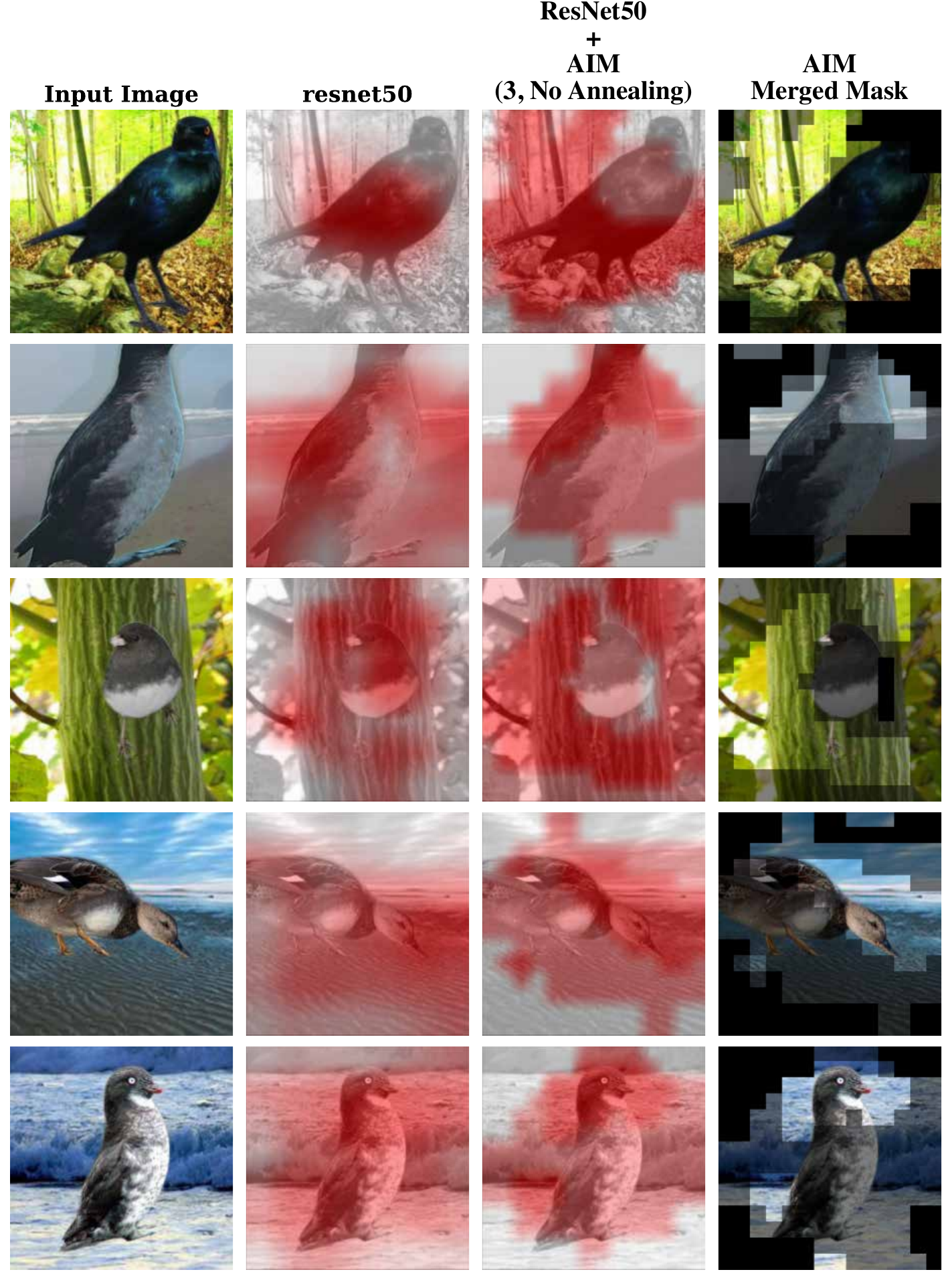}
    \end{tabular}
    \caption{\textbf{Qualitative results on the Waterbirds-100\% dataset \cite{petryk2022guiding}.} Comparison of GradCAM attribution maps between ResNet50+AIM models with different mask active-area thresholds, along with the generated merged masks.}
    \label{fig:appendix-attribution_wb100-resnet50}
\end{figure}

\begin{itemize}
    \item For the \textbf{CUB-200 dataset} we show the ConvNeXt-tiny+AIM ($T$=1) and ($T$=2) outputs in Figure~\ref{fig:appendix-attribution_cub-ConvNeXt-b1} and Figure~\ref{fig:appendix-attribution_cub-ConvNeXt-b2} respectively, while we illustrate the output of ResNet50+AIM ($T$=2) and ($T$=3) Figure~\ref{fig:appendix-attribution_cub-ResNet50-b2} and Figure~\ref{fig:appendix-attribution_cub-ResNet50-b3} respectively.

    \item For the \textbf{Waterbirds-95\%} dataset, we present the ConvNeXt-tiny+AIM outputs with mask active-area thresholds ($T$=1) and ($T$=2) in Figure~\ref{fig:appendix-attribution_wb95-ConvNeXt-b1} and Figure~\ref{fig:appendix-attribution_wb95-ConvNeXt-b2}, respectively. Additionally, we illustrate the outputs of ResNet50+AIM with ($T$=2) and ($T$=3) in Figure~\ref{fig:appendix-attribution_wb95-ResNet50-b2} and Figure~\ref{fig:appendix-attribution_wb95-ResNet50-b3}, respectively.

    \item For the \textbf{Waterbirds-100\%} dataset, we show the outputs of ConvNeXt-tiny+AIM and ResNet50+AIM in Figure~\ref{fig:appendix-attribution_wb100-convnext} and Figure~\ref{fig:appendix-attribution_wb100-resnet50}, respectively.

    \item For the \textbf{TravelingBirds} dataset, we present the outputs of ResNet50+AIM and ResNet101+AIM in Figure~\ref{fig:appendix-attribution_Travelingbirds-ResNet50} and Figure~\ref{fig:appendix-attribution_Travelingbirds-ResNet101}, respectively.
\end{itemize}

\section{Ablation results}
\label{sec:appendix:Ablation_results}
In this appendix section, we conduct an ablation study to explore the roles of individual components and parameters within our proposed method. By systematically adjusting or removing certain elements, we aim to gain a better understanding of their contributions to the model's performance.

\subsection{Active-Area Loss Annealing:}
\label{sec:appendix:Ablation_results-Active_Area_Loss_Annealing}
This section analyzes the performance variations in ConvNeXt-tiny+AIM models, focusing on two architectural variants: ConvNeXt-tiny+AIM ($T$=1) and ConvNeXt-tiny+AIM ($T$=2). The analysis examines changes in accuracy and energy pointing game (EPG) scores relative to the vanilla ConvNeXt-tiny model when the mask active-area threshold $\tau$ is adjusted on the Waterbrids-95\% dataset.\vspace{1em}

\begin{figure}[ht]
    \centering
    \includegraphics[width=1.0\linewidth]{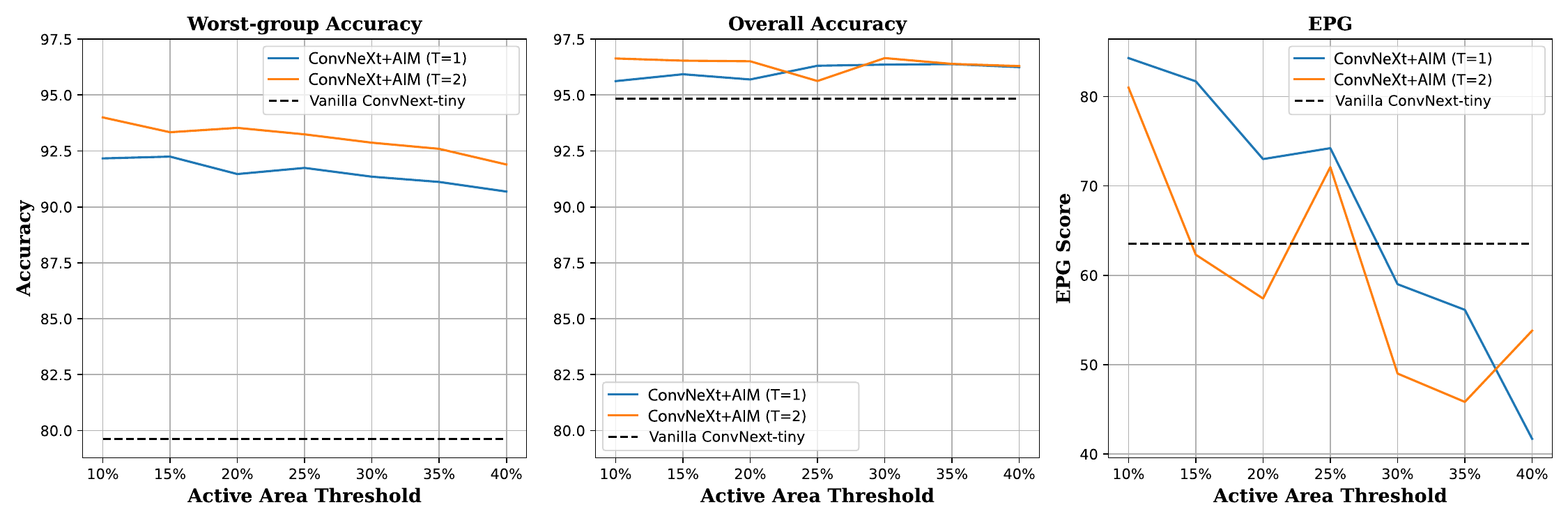}
    \caption{The first figure (on the right) illustrates how the worst-group accuracy of ConvNeXt-tiny+AIM ($T$=1) and ConvNeXt-tiny+AIM ($T$=2) changes with varying active-area thresholds. It shows that as the threshold increases, the worst-group accuracy decreases, while the change in overall accuracy is less pronounced. However, the EPG score also decreases with higher active-area thresholds. Despite this decline, the worst-group accuracy does not experience a significant drop, decreasing by approximately 3\%.}
    \label{fig:appendix:annealing_thrshold_annealing_graphs}
\end{figure}

The results in Table~\ref{table:appendix:annealing_thrshold_annealing} reveal that reducing the active-area threshold leads to improvements in both the worst group and overall accuracies; this can also be seen in Figure~\ref{fig:appendix:annealing_thrshold_annealing_graphs}. This finding suggests that by constraining the spatial areas upon which the models depend, they are compelled to identify and concentrate on regions pertinent to the task. This focus is evidenced by the higher EPG scores achieved, indicating effective task-related region identification.

\begin{table}[H]
    \centering
    \caption{This table presents the performance changes, in terms of accuracy and energy pointing game (EPG) scores on the Waterbrids-95\% dataset, for ConvNeXt-tiny+AIM models with two architectural variants: ConvNeXt-tiny+AIM ($T$=1) and ConvNeXt-tiny+AIM ($T$=2). These are compared against the vanilla ConvNeXt-tiny model when altering the mask active-area threshold $\tau$. The results indicate that a smaller active-area threshold leads to increases in both the worst group and overall accuracies. This suggests that limiting the spatial areas the models rely on encourages them to identify and focus on task-related regions, as reflected by high EPG scores. Another notable trend is that the ConvNeXt-tiny+AIM ($T$=1) variants achieve better EPG scores, implying that the masks in this variant, applied across all three stages of the top-down pathway, effectively concentrate on the regions of interest (ROI). This is also seen more clearly in Figure~\ref{fig:appendix:annealing_thrshold_annealing_graphs}.}
    \begin{tabular}{lcccc}
        \toprule
        \multirow{2}{*}{\textbf{Model}} & \multirow{2}{*}{\textbf{Annealing final threshold $\tau$}} &  \multicolumn{3}{c}{\textbf{Waterbirds-95\%}} \\
        \cline{3-5}
        & & \textbf{Worst-group} & \textbf{Overall} & \textbf{EPG}  \\
        \midrule
        Vanilla ConvNeXt-tiny & -- & 79.63 ($\pm$) & 93.755	($\pm$1.340) & 63.52 \\ \midrule
       AIM+ConvNeXt (1) & \multirow{2}{*}{10\%} & 92.16 ($\pm$) & 95.62 ($\pm$) & 84.3  \\
       AIM+ConvNeXt (2) &                       & 94 ($\pm$) & 96.63 ($\pm$) & 81 \\
        \midrule
       AIM+ConvNeXt (1) & \multirow{2}{*}{15\%} & 92.24 ($\pm$) & 95.93 ($\pm$) & 81.7  \\
       AIM+ConvNeXt (2) &                       & 93.33 ($\pm$) & 96.53 ($\pm$) & 62.3  \\
        \midrule
       AIM+ConvNeXt (1) & \multirow{2}{*}{20\%} & 91.46 ($\pm$) & 95.69 ($\pm$) & 73  \\
       AIM+ConvNeXt (2) &                       & 93.53 ($\pm$) & 96.51 ($\pm$) & 57.4  \\
        \midrule
       AIM+ConvNeXt (1) & \multirow{2}{*}{25\%} & 91.74 ($\pm$) & 96.31 ($\pm$) & 74.22  \\
       AIM+ConvNeXt (2) &                       & 93.24 ($\pm$) & 95.62 ($\pm$) & 72.1  \\
        \midrule
       AIM+ConvNeXt (1) & \multirow{2}{*}{30\%} & 91.35 ($\pm$) & 96.36 ($\pm$) & 59  \\
       AIM+ConvNeXt (2) &                       & 92.87 ($\pm$) & 96.65 ($\pm$) & 49  \\
        \midrule
       AIM+ConvNeXt (1) & \multirow{2}{*}{35\%} & 91.11 ($\pm$) & 96.38 ($\pm$) & 56.13  \\
       AIM+ConvNeXt (2) &                       & 92.59 ($\pm$) & 96.39 ($\pm$) & 45.83  \\
        \midrule
       AIM+ConvNeXt (1) & \multirow{2}{*}{40\%} & 90.68 ($\pm$) & 96.24 ($\pm$) & 41.7  \\
       AIM+ConvNeXt (2) &                       & 91.89 ($\pm$) & 96.29 ($\pm$) & 53.8  \\

        \bottomrule
        \end{tabular}
    
    \label{table:appendix:annealing_thrshold_annealing}
\end{table}

\subsection{Without The Auxiliary Losses:}
\label{sec:appendix:Ablation_results-Without_The_Auxiliary_Losses}
As discussed in Section~\ref{sec:method}, the AIM architecture incorporates a classification loss as an auxiliary loss at each stage in the top-down pathway, following the approach from \cite{FPN}. These losses help align the learned features at each stage with the final task. However, this raises an important question: How would the AIM network perform if these auxiliary losses were removed? What impact would this have on the network's overall performance?

Table \ref{tab:appendix:without_auxiliary_loss} examines the impact of removing auxiliary losses on the performance of ConvNeXt-tiny+AIM and ResNet50+AIM models across the CUB-200 and Waterbirds-95\% datasets. The results, measured in terms of accuracy and energy pointing game (EPG) scores, reveal notable performance fluctuations when auxiliary losses are omitted. On the Waterbirds-95\% dataset, some settings show performance gains, while others experience declines. Conversely, on the CUB-200 dataset, performance consistently deteriorates without the auxiliary losses. These observations highlight the critical role of auxiliary losses in maintaining stable and reliable performance across different models and datasets.

\begin{table}[H]
    \centering
    \caption{\textbf{AIM networks performance worsens without auxiliary losses.} This table presents the performance variations of ConvNeXt-tiny+AIM and ResNet50+AIM on the CUB-200 and Waterbirds-95\% datasets, measured in terms of accuracy (averaged over 4 different runs) and energy pointing game (EPG) scores. Red arrows indicate a decrease in the score compared to the corresponding architecture with the auxiliary losses, while green arrows represent an increase in scores. The results indicate that removing the auxiliary losses leads to fluctuations in performance across the datasets and models' variations. Specifically, on the Waterbirds-95\% dataset, performance increases for some settings while decreasing for others. On the CUB-200 dataset, performance consistently worsens compared to when auxiliary losses are used. These findings underscore the importance of auxiliary losses in achieving consistent performance across models and datasets.}
    \scalebox{0.8}{
    \begin{tabular}{l|ccccccc}
        \toprule
        \multirow{2}{*}{\textbf{Model}} & \multirow{2}{*}{\textbf{$\tau$}} & \multirow{2}{*}{\textbf{Auxiliary losses}} & \multicolumn{2}{c}{\textbf{CUB-200}}  &  \multicolumn{3}{c}{\textbf{Waterbirds-95\%}} \\
        \cline{6-8}\cline{4-5}
        & & & \textbf{Acc} & \textbf{EPG} & \textbf{Worst-group Acc} & \textbf{Overall} & \textbf{EPG}  \\
        \midrule

        \multirow{6}{*}{ConvNeXt-tiny+AIM (1, $\tau$)}  & 25\%           & no    & 88.29\textcolor{red}{$\downarrow$}	($\pm$0.16) & 59.4\textcolor{green}{$\uparrow$} & 92.24\textcolor{green}{$\uparrow$} ($\pm$0.73) & 96.84\textcolor{green}{$\uparrow$} ($\pm$0.086) & 73.82\textcolor{red}{$\downarrow$} \\
                                               & 25\%           & yes   & 88.63 ($\pm$0.13) &   58.54   & 91.31 ($\pm$1.1)  & 96.77 ($\pm$0.1) & 74.22 \\
                                               \cline{2-8}
                                               & 35\%           & no    & 88.51\textcolor{red}{$\downarrow$} ($\pm$0.075) & 50.12\textcolor{red}{$\downarrow$} & 91.7\textcolor{green}{$\uparrow$} ($\pm$0.76) & 96.70 \textcolor{green}{$\uparrow$} ($\pm$0.35) & 69.57\textcolor{green}{$\uparrow$} \\
                                               & 35\%           & yes	 & 88.82 ($\pm$0.21) &    57.49    & 90.1 ($\pm$1.44) & 96.63 $\pm$0.13) & 56.13 \\
                                               \cline{2-8}
                                               & no annealing   & no	 & 88.6\textcolor{red}{$\downarrow$}	($\pm$0.15) & 44.69\textcolor{red}{$\downarrow$}  & 92.23\textcolor{green}{$\uparrow$} ($\pm$2.88) & 94.89\textcolor{red}{$\downarrow$} ($\pm$2.13) & 48.17\textcolor{green}{$\uparrow$} \\
                                               & no annealing   & yes	 & 88.77 ($\pm$0.1) &    53.56    & 89.3 ($\pm$2.48)  & 96.07 ($\pm$0.16) & 42.57 \\ \midrule
        \multirow{6}{*}{ConvNeXt-tiny+AIM (2, $\tau$)}  & 25\%           & no    & 88.31\textcolor{red}{$\downarrow$} ($\pm$0.19) & 59.2\textcolor{red}{$\downarrow$}  & 93.18\textcolor{green}{$\uparrow$} ($\pm$1.1)  & 97.27\textcolor{green}{$\uparrow$} ($\pm$0.32) & 78.4\textcolor{green}{$\uparrow$} \\
                                               & 25\%           & yes   & 88.55 ($\pm$0.3)  &   60.27    & 90.1 ($\pm$1.46)  & 96.43 ($\pm$0.2) & 72.12 \\
                                               \cline{2-8}
                                               & 35\%           & no    & 88.47\textcolor{red}{$\downarrow$} ($\pm$0.17) & 54.19\textcolor{red}{$\downarrow$} & 92.01\textcolor{green}{$\uparrow$} ($\pm$0.51) & 96.85 \textcolor{green}{$\uparrow$} ($\pm$0.1) &  63.98\textcolor{green}{$\uparrow$} \\
                                               & 35\%           & yes	 & 88.67 ($\pm$0.25) & 56.82 & 91.41 ($\pm$0.22) & 96.40 ($\pm$0.23) & 45.83 \\
                                               \cline{2-8}
                                               & no annealing   & no	 & 88.68\textcolor{green}{$\uparrow$} ($\pm$0.15) & 41.49\textcolor{red}{$\downarrow$} & 91.78\textcolor{green}{$\uparrow$} ($\pm$0.74) & 95.94 \textcolor{red}{$\downarrow$} ($\pm$0.26) & 58.48\textcolor{red}{$\downarrow$} \\
                                               & no annealing   & yes	 & 88.62 ($\pm$0.21) & 55.54 & 88.7 ($\pm$1.4) & 96.39 ($\pm$0.104) & 65.84 \\ \midrule

        \multirow{6}{*}{ResNet50+AIM (2, $\tau$)}    & 25\%           & no    & 79.74\textcolor{red}{$\downarrow$} ($\pm$0.14) & 55.86\textcolor{red}{$\downarrow$} & 77.68\textcolor{red}{$\downarrow$}	($\pm$0.673) & 92.41 \textcolor{red}{$\downarrow$} ($\pm$0.184) & 61.89\textcolor{red}{$\downarrow$} \\
                                               & 25\%           & yes   & 80.9 ($\pm$0.345) &   57.56   & 91.31 ($\pm$1.1)  & 96.77 ($\pm$0.1) & 74.22 \\
                                               \cline{2-8}
                                               & 35\%           & no    & 79.85\textcolor{red}{$\downarrow$} ($\pm$0.07) & 54.54\textcolor{red}{$\downarrow$} & 77.41\textcolor{red}{$\downarrow$}	(0)	 & 92.35\textcolor{green}{$\uparrow$} ($\pm$0.1) & 51.66\textcolor{red}{$\downarrow$} \\
                                               & 35\%           & yes	 & 80.9	($\pm$0.12) &    57    & 90.1 ($\pm$1.44) & 92.34 ($\pm$0.25) & 56.13 \\
                                               \cline{2-8}
                                               & no annealing   & no	 & 79.27\textcolor{red}{$\downarrow$}	($\pm$0.23) & 47.49\textcolor{red}{$\downarrow$}  & 75.50\textcolor{red}{$\downarrow$} ($\pm$0.545) & 91.7 \textcolor{red}{$\downarrow$}($\pm$0.26) & 43.04\textcolor{green}{$\uparrow$} \\
                                               & no annealing   & yes	 & 80.84	($\pm$0.28) &    53.81    & 89.3 ($\pm$2.48)  & 91.9 ($\pm$0.11) & 42.57 \\ \midrule

        \multirow{6}{*}{ResNet50+AIM (3, $\tau$)}    & 25\%           & no    & 79.69\textcolor{red}{$\downarrow$} ($\pm$0.35) & 57.66\textcolor{green}{$\uparrow$} & 77.83\textcolor{red}{$\downarrow$} ($\pm$0.61) & 92.64 \textcolor{red}{$\downarrow$}($\pm$0.38) & 71.87\textcolor{green}{$\uparrow$} \\
                                               & 25\%           & yes   & 80.9 ($\pm$0.345) &   56.76   & 91.31 ($\pm$1.1)  & 96.77 ($\pm$0.1) & 61.89 \\
                                               \cline{2-8}
                                               & 35\%           & no    & 79.77\textcolor{red}{$\downarrow$} ($\pm$0.41) & 55.04\textcolor{green}{$\uparrow$} & 78.38\textcolor{red}{$\downarrow$} ($\pm$1.48) & 91.92\textcolor{red}{$\downarrow$} ($\pm$0.35) & 46.35\textcolor{red}{$\downarrow$} \\
                                               & 35\%           & yes	 & 80.9	($\pm$0.12)  & 54.80    & 90.1 ($\pm$1.44) & 92.42 ($\pm$0.46) & 52.14 \\
                                               \cline{2-8}
                                               & no annealing   & no	 & 79.55\textcolor{red}{$\downarrow$} ($\pm$0.21) & 48.73\textcolor{red}{$\downarrow$} & 74.87\textcolor{red}{$\downarrow$} ($\pm$2.0) & 92.32 \textcolor{red}{$\downarrow$}($\pm$0.32) & 36.79\textcolor{red}{$\downarrow$} \\
                                               & no annealing   & yes	 & 80.84	($\pm$0.28) & 52.71 & 89.3 ($\pm$2.48)  &  87.39 ($\pm$8.56)  & 41.78 \\

        \bottomrule
        \end{tabular}
        }
    
    \label{tab:appendix:without_auxiliary_loss}
\end{table}

\subsection{Do we need multiple mask estimators, one at each level?}
\label{sec:appendix:Ablation_results-single_level}
One of the building blocks of our proposed method is the use of mask estimators in the top-down pathway, where we employ a mask estimator module at each stage, which we denote here as the Full AIM model. Given that in convolutional neural networks the feature maps from the last level determine the final semantic features used in the classification layer, a question arises: \textit{can we rely solely on the mask generated at the first stage of the top-down pathway (this stage corresponds to the final convolutional layer in the backbone network used)?} \vspace{1em}

This question is motivated by the semantic richness of the feature maps at this level; thus, selecting areas by masking in these feature maps is supposed to truly represent the task-related semantics.

To investigate this question, we adjusted our network to use only the masks generated at the first stage of the top-down pathway. This is done by passing the first stage's masks to the subsequent stage after up-scaling them by a factor of 2, as the subsequent stage has higher spatial resolutions.

\begin{table}[H]
\centering
\caption{Test accuracies of different AIM models on the CUB-200 dataset with masks only adapted from the first stage of the top-down pathway (highlighted with \checkmark mark) versus the standard AIM network, which employs a mask estimator at each stage of the top-down pathway.}
\scalebox{0.9}{
    \begin{tabular}{@{}lcc@{}}
        \toprule
        \textbf{Model} & \textbf{One Mask Estimator} &  \textbf{CUB-200 (\%)} \\ \midrule

       AIM+ConvNeXt (1, 25\%)           & $\times$ & 88.635 ($\pm$0.13)  \\ 
       AIM+ConvNeXt (1, 35\%)           & $\times$ & \textbf{88.82 ($\pm$0.213)} \\ 
       AIM+ConvNeXt (1, No Annealing)   & $\times$ & \underline{88.775 ($\pm$0.044)}  \\ 
        \midrule
       AIM+ConvNeXt (2, 25\%)           & $\times$ & 88.557 ($\pm$0.296) \\ 
       AIM+ConvNeXt (2, 35\%)           & $\times$ & 88.677 ($\pm$0.25)   \\ 
       AIM+ConvNeXt (2, No Annealing)   & $\times$ & 88.622 ($\pm$0.211)  \\   
        \midrule \midrule

       AIM+ConvNeXt (1, 25\%) & \checkmark & 88.14 ($\pm$ 0.44) \\ 
       AIM+ConvNeXt (1, 35\%) & \checkmark & 88.22 ($\pm$ 0.26) \\ 
       AIM+ConvNeXt (1, No Annealing) & \checkmark & 88.25 ($\pm$ 0.33) \\ \midrule

       AIM+ConvNeXt (2, 25\%) & \checkmark & 88.52 ($\pm$ 0.22) \\ 
       AIM+ConvNeXt (2, 35\%) & \checkmark & 88.41 ($\pm$ 0.20) \\ 
       AIM+ConvNeXt (2, No Annealing) & \checkmark & 88.26 ($\pm$ 0.28) \\ 

        \bottomrule
    \end{tabular}
    }
    
    \label{table:appending:accuracy_scores_bernt}
\end{table}

\begin{figure}[h]
        \centering
        \includegraphics[width=1.0\textwidth]{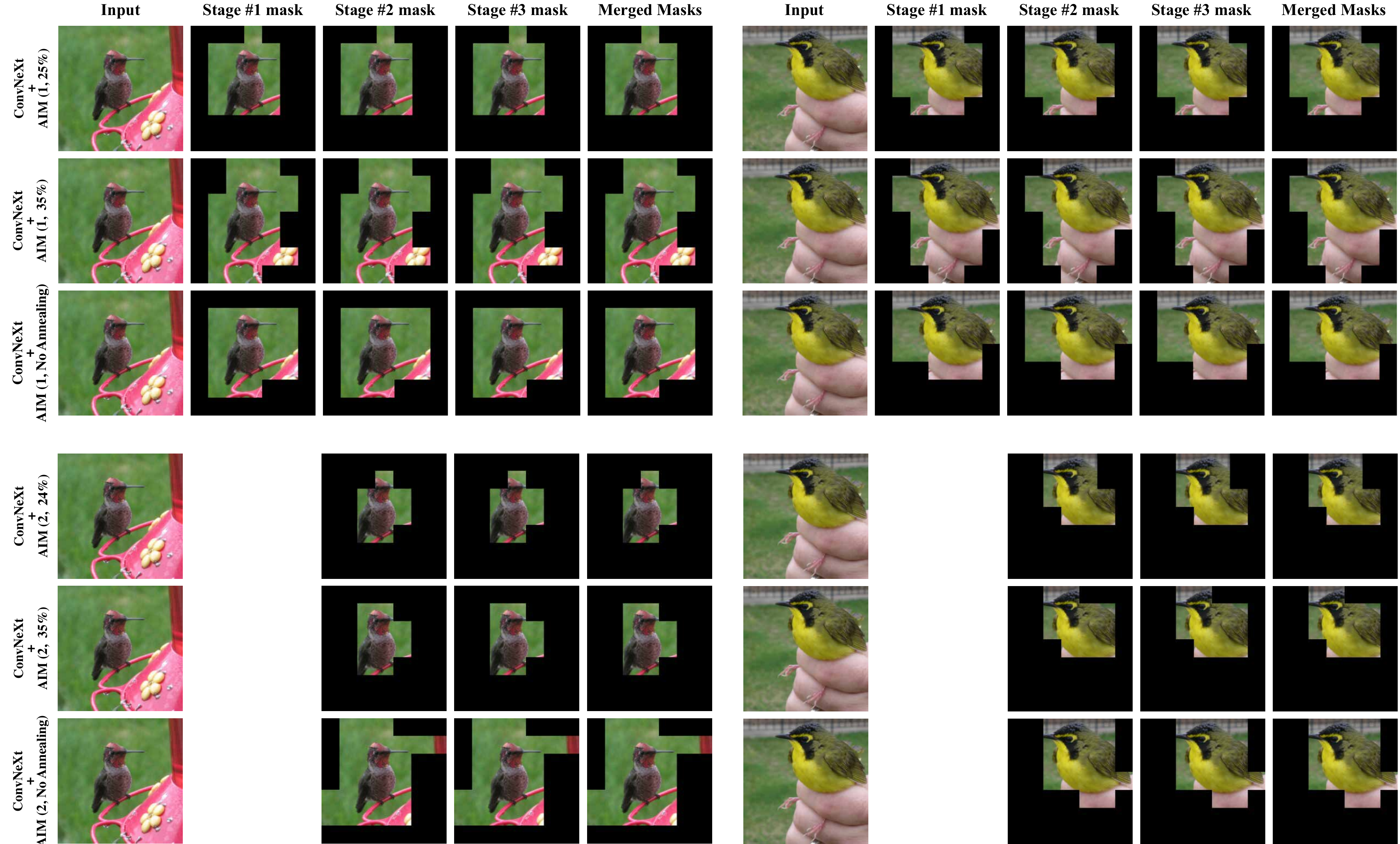}
        \caption{\textbf{Qualitative Comparison between the different architectural configurations of AIM when using masks from the last layer.This figure illustrates the application of masks generated at the last stage (stage \#3) and used in the subsequent stages after upsampling.}}
        \label{fig:appendix_bernt_idea}
\end{figure}

Table~\ref{table:appending:accuracy_scores_bernt} presents the test accuracies of different configurations of our AIM models on the CUB-200 dataset. In these experiments, we compare the standard AIM network, which employs a mask estimator at each stage of the top-down pathway, with a modified version that uses masks only from the first stage of the top-down pathway. The modified approach passes the first-stage masks to subsequent stages to be applied on the corresponding feature maps. These models are indicated with a \checkmark\ in the `One Mask Estimator' column in the Table \ref{table:appending:accuracy_scores_bernt}.

Analyzing the results, we observe that the models' results are very close to each other, with models using masks solely from the first stage generally achieving slightly lower test accuracies than those employing mask estimators at each stage. 

In convolutional neural networks, as input data moves through the network layers, the feature maps become richer in semantic information but lose spatial localization. We hypothesize that applying masks generated at the first stage of the top-down pathway, which is coarser and less localized (See Figure~\ref{fig:appendix_bernt_idea}) compared to those from later stages, to subsequent lower stages could inadvertently include non-task-related regions, potentially diminishing performance, especially on out-of-distribution datasets.

\subsection{The bottom-up bottlenecking approach:}
\label{sec:appendix:Ablation_results-Bottom_up_bottlenecking}
One of the most relevant works to ours is \cite{Verelst_2020}, which uses a bottom-up masking approach to focus on essential regions in the generated feature maps. This approach relies on employing mask units between the main blocks of a convolutional network to produce binary masks highlighting task-relevant regions for the network. These mask units utilize the Gumbel-softmax trick to generate binary masks, which are then used to spatially bottleneck feature maps produced after each main block.\vspace{1em}

\begin{figure}[h]
    \centering
    \includegraphics[width=0.8\linewidth]{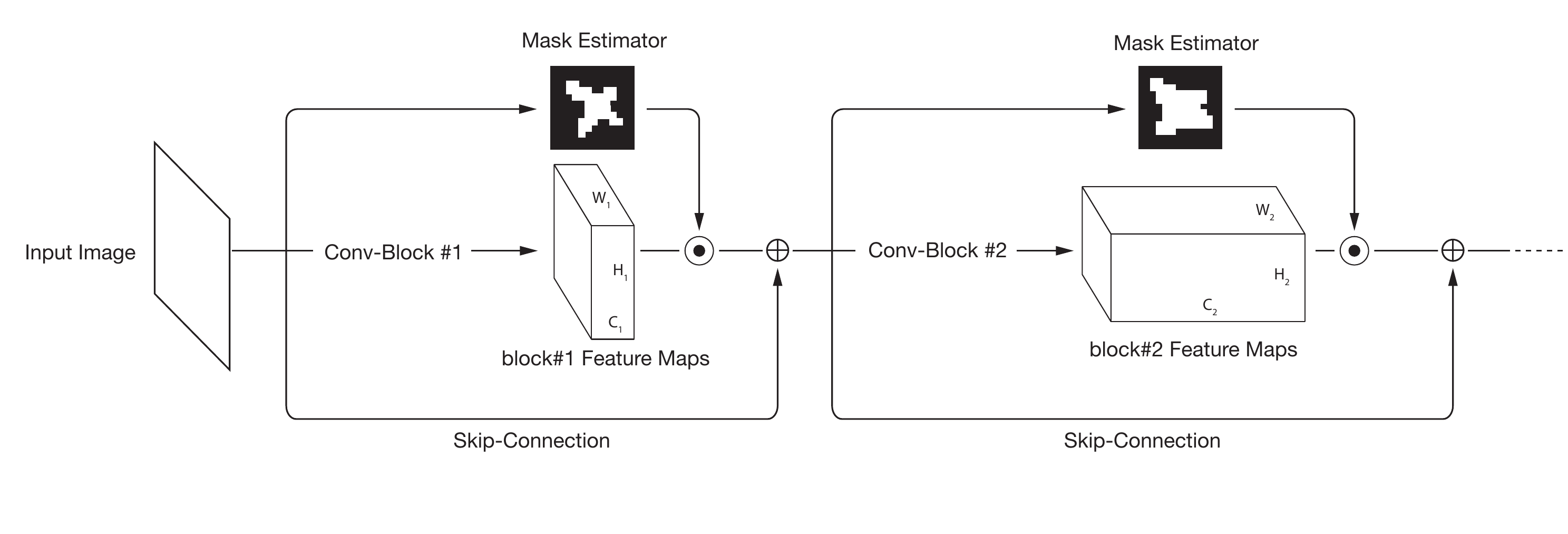}
    \caption{The illustration depicts the architecture of the bottom-up bottlenecking model, adapted from \cite{Verelst_2020}, highlighting the flow of data through the model, including skip connections and mask estimators. Each main convolutional block in the network consists of three branches: the first is a mask estimator that employs the Gumbel-softmax trick to predict a binary mask, the second is the original convolutional block, and the third is a skip connection. The generated masks are applied to the output of the convolutional block, resulting in spatially sparse feature maps. Subsequently, the skip connection performs an element-wise summation between these sparse feature maps and the block's input, transforming the sparse maps back into dense ones.}
    \label{fig:appendix-bigated_skip}
\end{figure}

However, as discussed in the related work Section~\ref{sec:related_work}, this bottom-up approach needs skip-connections to work (See Figure~\ref{fig:appendix-bigated_skip}) as the network needs more information to be flown from the shallow layer to the deepest layers to form an understanding of the scene, this compromised the inherent interpretability that comes naturally with the sparse feature maps, which with the skip-connections fall back to dense features maps.

\begin{figure}[h]
    \centering
    \includegraphics[width=0.8\linewidth]{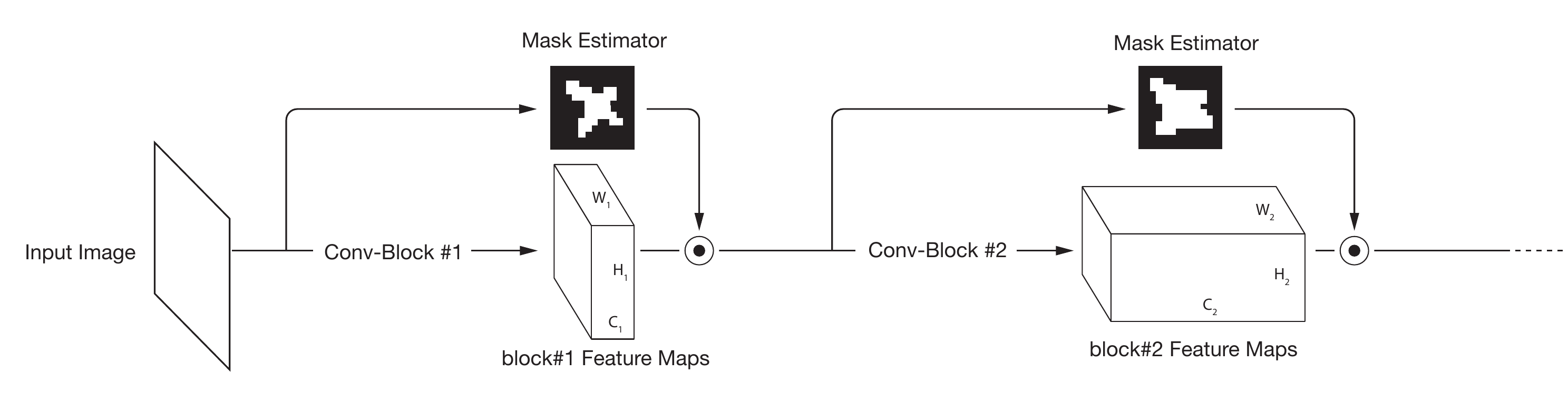}
    \caption{\textbf{\textit{Bottom-up} AIM Network } An illustration showing the same bottom-up bottlenecking model's architecture from Figrue~\ref{fig:appendix-bigated_skip} but without the skip-connection to maintain the sparsity of the feature maps.}
    \label{fig:appendix-bigated_no_skip}
\end{figure}

We closely follow the method described in \cite{Verelst_2020}, but with one key difference: \textsl{we utilized fully sparse feature maps without skip connections}. This decision aligns with our objective of creating an interpretable model by focusing on generating fully sparse feature maps, which makes the decision-making process transparent (See Figure~\ref{fig:appendix-bigated_no_skip}). For this purpose, we adhered to the same training setup we used for our approach, 200 epochs with active mask annealing applied for half the training period. We denote this architecture as \textit{bottom-up} AIM Network.

We tested three architectural variants that differed in the number of masked convolutional blocks and experimented with two annealing threshold settings.\vspace{1em}

\begin{table}[H]
    \centering
    \caption{This table presents the performance of Bi-gatedNet with ConvNeXt-tiny on the CUB-200 dataset when different blocks are masked and various mask active-area thresholds ($\tau$) are applied. Masking all three blocks resulted in the lowest accuracy with high variability, indicating unstable training. Masking only the last two blocks improved accuracy but yielded poorly localized masks. Masking only the last block achieved the highest accuracy, surpassing the unmasked ConvNeXt-tiny model, and produced highly focused and localized masks.}
            \begin{tabular}{@{}lcc@{}}
                \toprule
                \textbf{Model Name} & \textbf{Bottom-Up Blocks Masked} & \textbf{CUB-200 (\%)} \\ \midrule
                ConvNext-tiny                  & - & 86.827 ($\pm$0.761) \\ \midrule

                standard ConvNeXt-tiny+AIM ($T$=1, $\tau$=25\%)  & -  & \textbf{88.82 ($\pm$0.213)} \\ 
                standard ConvNeXt-tiny+AIM ($T$=2, $\tau$=25\%)  &  - & \underline{88.677 ($\pm$0.25)} \\ 
                
                \midrule
                
                \textit{Bottom-up} ConvNeXt-tiny+AIM ($\tau$=25\%)  & 1, 2, 3    & 72.79  ($\pm$8.51) \\ 
                \textit{Bottom-up} ConvNeXt-tiny+AIM ($\tau$=25\%)  & 2, 3       & 82.53 ($\pm$1.39) \\ 
                \textit{Bottom-up} ConvNeXt-tiny+AIM ($\tau$=25\%)  & 3          & 86.202 ($\pm$0.44) \\ \midrule
                
                \textit{Bottom-up} ConvNeXt-tiny+AIM ($\tau$=35\%)  & 1, 2, 3    & 67.45 ($\pm$8.52) \\
                \textit{Bottom-up} ConvNeXt-tiny+AIM ($\tau$=35\%)  & 2, 3       & 84.00 ($\pm$1.38) \\ 
                \textit{Bottom-up} ConvNeXt-tiny+AIM ($\tau$=35\%)  & 3          & 86.975 ($\pm$0.34) \\ \bottomrule
            \end{tabular}
        
        \label{table:bigated_results}    
\end{table}

\begin{figure}[h]
    \centering
    \includegraphics[width=1.0\linewidth]{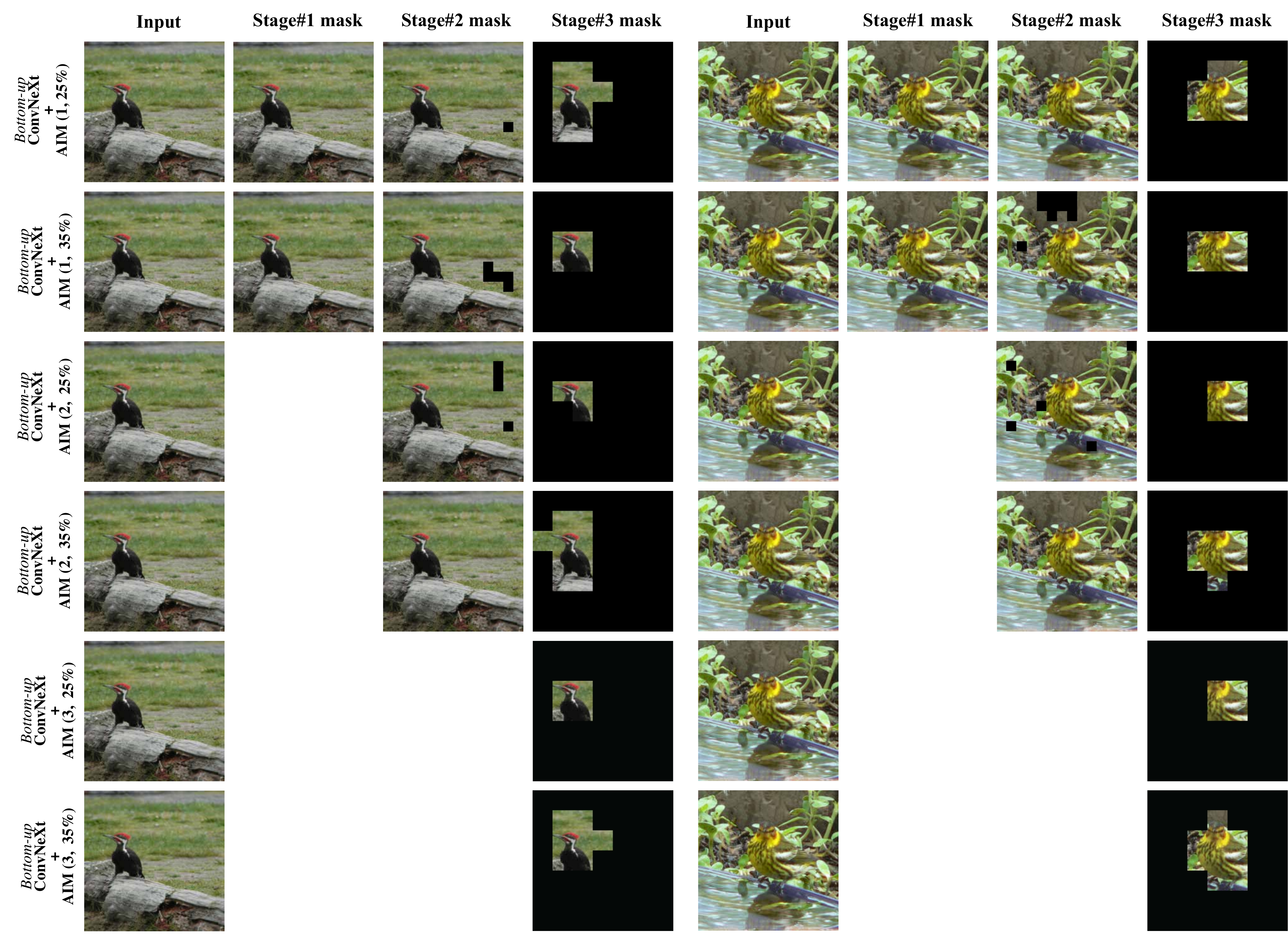}
    \caption{\textbf{\textit{Bottom-up} AIM Network Fails to Generate Focused Masks} This figure illustrates the masks generated at different stages of the ConvNeXt-tiny backbone within the Bottom-up AIM network. Even when we limit the active area region of the masks to 25\% and 35\% using mask active-area loss annealing, the masks from the earlier blocks remain fully activated, covering the entire image. Only the mask from the last block effectively focuses on the relevant region. This demonstrates that the Bottom-up AIM network does not produce localized, task-related masks in its earlier stages.}
    \label{fig:Appending_bigated_masks}
\end{figure}

Our main observation was that applying masks to all three blocks resulted in the lowest accuracy with high variability across trials (see Table \ref{table:bigated_results}), indicating unstable training and the model's inability to learn effective representations. Additionally, the masks generated in this configuration (see Figure~\ref{fig:Appending_bigated_masks}) showed almost no masking despite the application of the mask active-area loss with annealing.

When we adjusted the architecture to mask only the last two blocks, the accuracy improved sharply, by 10\% for the annealing threshold of 0.25 and 13\% for 0.35, reaching above 80\% accuracy (see Table \ref{table:bigated_results}). However, the quality of the masks remained poor: the masks at the lower levels were still almost fully active, and only the mask from the last block focused on the bird.

This is evident in Figure~\ref{fig:Appending_bigated_masks}, which shows that even when we limit the active area region to 25\% and 35\% using mask active-area loss annealing, the masks remain fully activated except for the one from the last block.

Furthermore, table \ref{table:bottm-up_vs_top_down} compares between our top-down and bottom-up approach in terms of sparsity and localization (EPG); the top-down method outperforms the bottom-up baseline on both metrics even without any sparsity loss (No annealing) on Waterbirds-95\% dataset.

\begin{table}[H]
    \centering
    \caption{Comparison of our top-down approach (ConvNext+AIM) against a bottom-up baseline on the Waterbirds-95\% dataset. Our method demonstrates superior performance across all metrics, highlighting its ability to improve localization (EPG) and sparsity even without an explicit sparsity objective (No annealing).}
    \begin{tabular}{@{}l@{}c@{\,\,}c@{\,\,}c@{}}
    \toprule
        \textbf{Model} &  \textbf{Worst group ACC ($\pm$ std) $\uparrow$} & \textbf{EPG ($\pm$ std) $\uparrow$} & \textbf{Sparsity score ($\pm$ std) $\downarrow$}\\ \toprule
        Bottom-up ([1, 2, 3], 25\%)    & 85.63 ($\pm$ 4.2)   & 29.28 ($\pm$ 21.4) & 100 ($\pm$ 0)\\ 
        ConvNext+AIM (1, 25\%)        &  \textbf{92.21 ($\pm$ 1.2)}  & \textbf{77.1 ($\pm$ 0.06)} & 17.7 ($\pm$ 0.6)\\ \midrule
        Bottom-up ([1, 2, 3], No annealing)    & 87.41 ($\pm$ 4.2)   & 24.66 ($\pm$ 9.4) & 100 ($\pm$ 0)\\ 
        ConvNext+AIM (1, No annealing) &  \textbf{89.5 ($\pm$ 4.13))}  & \textbf{43.43 ($\pm$ 5.13)} &  71.48 ($\pm$ 0) \\
        \bottomrule
    
    \end{tabular}
    
    \label{table:bottm-up_vs_top_down}

\end{table}

In summary, training with the bottom-up approach indicates that masking fewer blocks leads to better performance and more robust masks.

\subsection{Emphasizing peripheral regions in mask estimator initialization}
\label{sec:appendix:Ablation_results-initialization}
\begin{figure}[ht]
    \centering
    \includegraphics[width=0.3\linewidth]{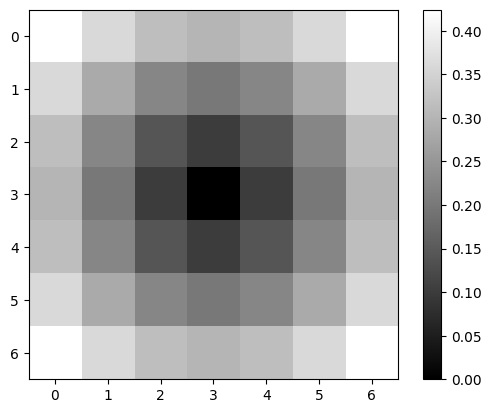}
    \caption{An illustration showing the weight matrix of a convolutional layer with one (7 $\times$ 7) filter, initialized using the proposed idea of weighting the convolutional filter weights according to their spatial distance from the center of the filter kernel. Specifically, we assign larger initial values to the weights corresponding to the edges of the filters than to those closer to the center, effectively biasing the filters to be more responsive to features in the outer areas of the input.}
    \label{fig:mask_estimator_initialization}
\end{figure}

During training, the layers of the mask estimators are typically initialized randomly. To investigate the effect of the initialization scheme on the performance of the mask estimators, we propose an alternative initialization method that emphasizes the peripheral regions over the central regions. This is achieved by weighting the convolutional filter weights according to their spatial distance from the center of the filter kernel. Specifically, we assign larger initial values to the weights corresponding to the edges of the filters than to those closer to the center, effectively biasing the filters to be more responsive to features in the outer areas of the input (see Figure~\ref{fig:mask_estimator_initialization} for an illustration).

However, our experiments on the CUB-200 dataset reveal that this specialized initialization does not have a significant impact on performance. As shown in Table~\ref{tab:appendix:mask_estimator_initialization}, the model's accuracy remains essentially unchanged compared to when using standard random initialization. This suggests that the mask estimators are robust to the initial weight distribution and that the network is capable of learning effective representations regardless of this specific initialization strategy.

\begin{center}
\begin{table}[H]
    \centering
    \caption{\textbf{Emphasizing Peripheral Regions in Mask Initialization Does Not Impact AIM Network Performance.} This figure compares the AIM network's performance on the CUB-200 dataset using two different initialization strategies for the mask estimators: standard random initialization and an initialization that emphasizes peripheral regions by assigning larger weights to filter edges. The results indicate that the model's accuracy remains essentially unchanged between the two approaches. This suggests that the mask estimators are robust to the initial weight distribution, and the network can effectively learn meaningful representations regardless of this specific initialization strategy.}
    \begin{tabular}{l c}
        \hline
        \textbf{Model} & \textbf{CUB-200 (\%)} \\
        \hline
        \textit{Edge-emphasized initialization} ConvNeXt-tiny+AIM (2, 25\%)      & \textbf{88.76 (±0.171)} \\
        \textit{Standard random initialization} ConvNeXt-tiny+AIM (2, 25\%)                 & 88.678 (±0.251) \\
        \hline
    \end{tabular}
    
    \label{tab:appendix:mask_estimator_initialization}
\end{table}
\end{center}

\subsection{User Perception Study Details}
\label{sec:appendix:user_study}

\begin{figure}[h]
    \centering
    \includegraphics[width=0.6\textwidth]{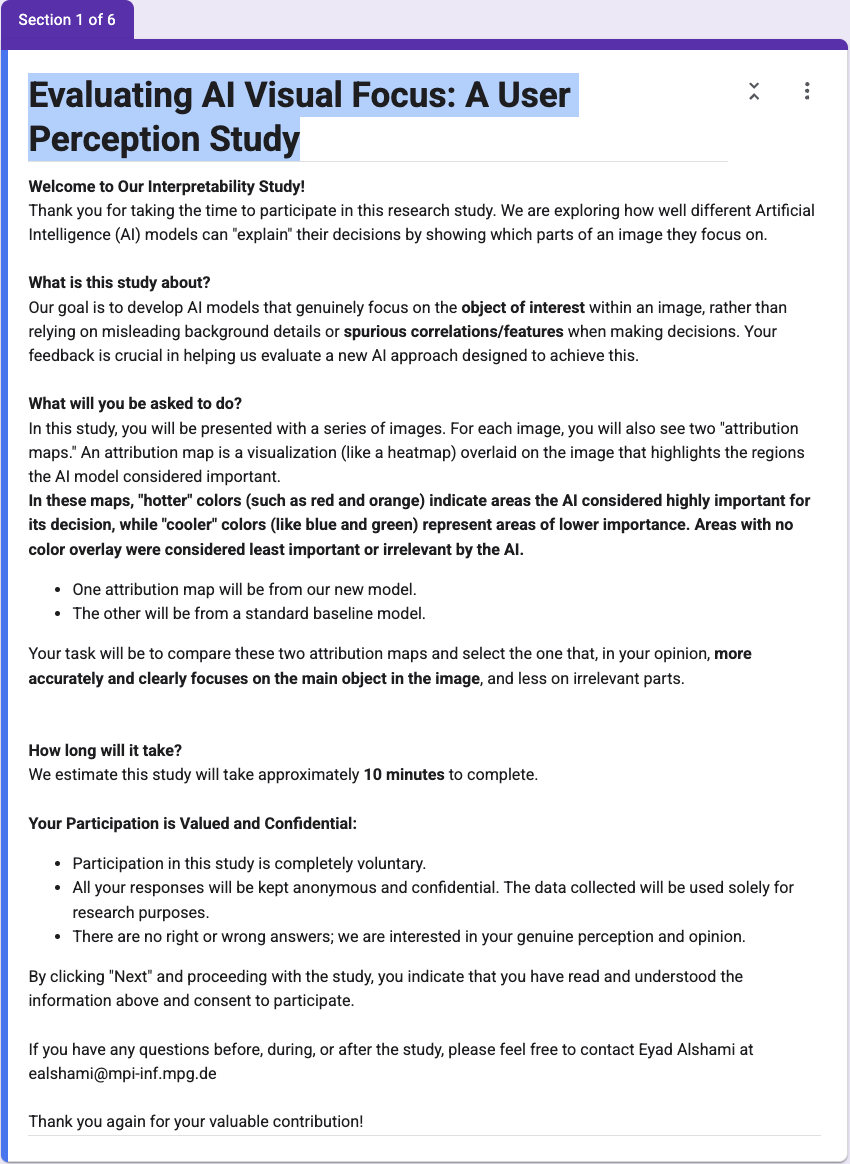}
    \caption{The user study welcome page, which provided participants with initial instructions.}
    \label{fig:appendix_welcome_page}
\end{figure}

To quantitatively assess if our model's improved focus is perceptually meaningful, we conducted an online user study. Participants accessing the study were first directed to a welcome page that outlined the research goals, the task procedure, and our data privacy policy (Fig.~\ref{fig:appendix_welcome_page}). The introduction defined attribution maps as heatmaps used to visualize an AI's focus and assured participants that all responses would be anonymous and confidential.

The study was designed to be completed in approximately 10 minutes. To achieve this while still covering a broad set of 100 images from the Waterbirds-100\% dataset, the images were randomly partitioned into four disjoint subsets of 25. Each of the 107 participants was randomly assigned to evaluate exactly one of these four subsets. Following this assignment, participants proceeded to the main evaluation task, which consisted of 25 forced-choice comparison trials.

\begin{figure}[h]
    \centering
    \includegraphics[width=0.8\textwidth]{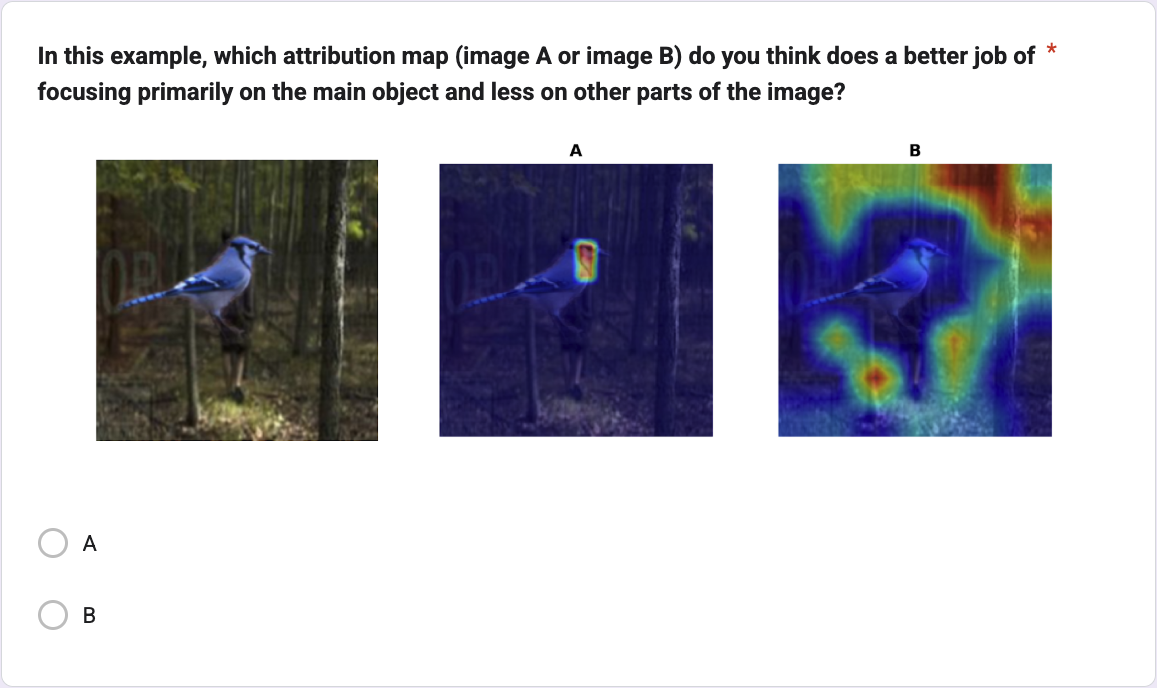}
    \caption{An example trial from our user perception study. Participants were shown the original image (left) and two GradCAM attribution maps from the baseline ConvNeXt (right) and our \methodName-equipped model (center). In this case, the baseline model is distracted by the spurious forest background, while our method correctly localizes the waterbird. This clear distinction in focus led users to significantly prefer our model's explanations.}
    \label{fig:user_study_example}
\end{figure}

In each trial, participants were shown an image from their assigned subset, displayed alongside two GradCAM attribution maps: one from the baseline vanilla ConvNeXt and one from our \methodName-equipped model. For each pair, they were asked to select the map that, in their opinion, \textit{``more accurately and clearly focuses on the main object in the image, and less on irrelevant parts''.} An example of this trial layout is shown in Fig.~\ref{fig:user_study_example}. To mitigate positional bias, the on-screen placement of the two maps was randomized for every trial.

A total of 107 participants completed the study, yielding $107 \times 25 = 2675$ individual evaluations. The results show a strong and significant user preference for our model. \methodName's attribution maps were chosen in \textbf{70.7\%} of the evaluations. A two-sided binomial test confirms that this outcome is highly significant ($p < 0.00001$), providing strong evidence that the explanations generated by \methodName are more aligned with human intuition regarding object-centric focus compared to the baseline.

\end{document}